\documentclass{article}

\usepackage{amsmath}
\usepackage{amssymb}
\usepackage{marvosym}
\usepackage{minted}
\usepackage[preprint]{neurips_2026}

\makeatletter
\def\@noticestring{%
  Preprint.\par
  (\Letter)Corresponding authors: \texttt{jgwu@pku.edu.cn} (Jiageng Wu);
  \texttt{smy@pku.edu.cn} (Mingyang Sun).%
}
\makeatother

\usepackage[utf8]{inputenc} 
\usepackage[T1]{fontenc}    
\usepackage{hyperref}       
\usepackage{url}            
\usepackage{booktabs}       
\usepackage{amsfonts}       
\usepackage{nicefrac}       
\usepackage{microtype}      
\usepackage[table]{xcolor} 
\usepackage{array}
\usepackage{multirow}
\usepackage{tabularx}
\usepackage[most]{tcolorbox}
\usepackage{fancyvrb}
\usepackage{enumitem}
\usepackage{longtable}

\usepackage{blindtext}

\usepackage{amsthm}
\usepackage{amsmath}

\usepackage{algorithm}
\usepackage{algpseudocode}

\usepackage{tikz}
\usetikzlibrary{positioning,calc,arrows.meta,fit}

\theoremstyle{definition} 

\newcommand{\heatcell}[2]{\cellcolor{orange!#1}\textcolor{black}{#2}}
\newcommand{\metricdelta}[1]{\textcolor{blue}{\scriptsize(#1)}}
\newcommand{\regdelta}[1]{\textcolor{red}{\scriptsize(#1)}}

\newcolumntype{L}[1]{>{\raggedright\arraybackslash}p{#1}}

\newtcolorbox{prompttemplate}[1]{
  title={#1},
  breakable,
  boxrule=0.5mm,
  colback=white,
  colframe=black,
  fonttitle=\bfseries,
  listing options={
    upquote=true,
    basicstyle=\ttfamily,
    columns=fullflexible
  }
}

\title{OptArgus: A Multi-Agent System to Detect Hallucinations in LLM-based Optimization Modeling}

\author{%
  Zhong Li$^{\clubsuit}$ \quad
  Zihan Guo$^{\blacktriangle}$ \quad
  Xiaohan Lu$^{\blacktriangle}$ \quad
  Juntao Wang$^{\heartsuit, \spadesuit}$ \\
  \textbf{Jie Song}$^{\blacktriangle}$ \quad
  \textbf{Chao Shen}$^{\bigstar}$ \quad
  \textbf{Jiageng Wu}$^{\blacktriangle}$\textsuperscript{(\Letter 1)} \quad
  \textbf{Mingyang Sun}$^{\blacktriangle}$\textsuperscript{(\Letter 2)} \\
  $^{\clubsuit}$Great Bay University,
  $^{\blacktriangle}$Peking University,
  $^{\heartsuit}$Jilin University,\\
  $^{\bigstar}$Zhejiang University,
  $^{\spadesuit}$Shenzhen Loop Area Institute \\
}

\begin{document}

\maketitle

\begin{abstract}
 
Large language models (LLMs) are increasingly used to translate natural-language optimization problems into mathematical formulations and solver code, but matching the reference objective value is not a reliable test of correctness: an artifact may agree numerically while still changing the underlying optimization semantics. We formulate this issue as \emph{optimization-modeling hallucination detection}, namely structural consistency auditing over the problem description, symbolic model, and solver implementation. We develop, to our knowledge, the first fine-grained hallucination taxonomy specifically for optimization modeling, spanning objective, variable, constraint, and implementation failures. We use this taxonomy to design OptArgus, a multi-agent detector with conductor routing, specialist auditors, and evidence consolidation. To evaluate this setting, we introduce a three-part benchmark suite with $484$ clean artifacts, $1266$ controlled injected artifacts, and $6292$ natural LLM-generated artifacts. Against a matched single-agent baseline, OptArgus produces fewer false alarms on clean artifacts, more accurate top-ranked localization on controlled single-error cases, and stronger detection on natural model outputs. Together, these contributions turn optimization-modeling hallucination detection into a concrete empirical problem and suggest that modular, taxonomy-grounded auditing is a practical route to more reliable optimization modeling.

\end{abstract}

\section{Introduction}
\label{sec:introduction}

Optimization modeling \citep{bisschop2006aimms,hart2017pyomo} concerns the specification of a decision problem as a mathematical program, typically by defining decision variables, constraints that characterize feasible decisions, and an objective function that captures the desired notion of optimality. In practice, this process often begins with a \textit{stakeholder} describing the task, operational context, and business rules in natural language; \textit{domain specialists} then interpret the application semantics, \textit{operations-research (OR) experts} formulate the corresponding mathematical model, and technically trained \textit{modelers} implement that formulation in solver code. This workflow underlies decision support in logistics \citep{bartolacci2012optimization}, supply chains \citep{garcia2015supply},  transportation \citep{gkiotsalitis2022public}, energy systems \citep{dincer2017optimization}, finance \citep{cornuejols2018optimization}, healthcare \citep{rais2011operations}, etc. Despite its importance, conventional optimization modeling remains constrained by three persistent barriers \citep{fan2025artificial,xiao2025survey}: it relies on scarce cross-disciplinary expertise, it incurs substantial time and financial costs for formulation, validation, and implementation, and it is often insufficiently agile because even modest changes in requirements, assumptions, or data interfaces can force the model and code to be revised. These limitations make \textbf{automatic optimization modeling} an attractive research direction.

\begin{figure}[h]
  \centering
  \includegraphics[width=0.98\textwidth]{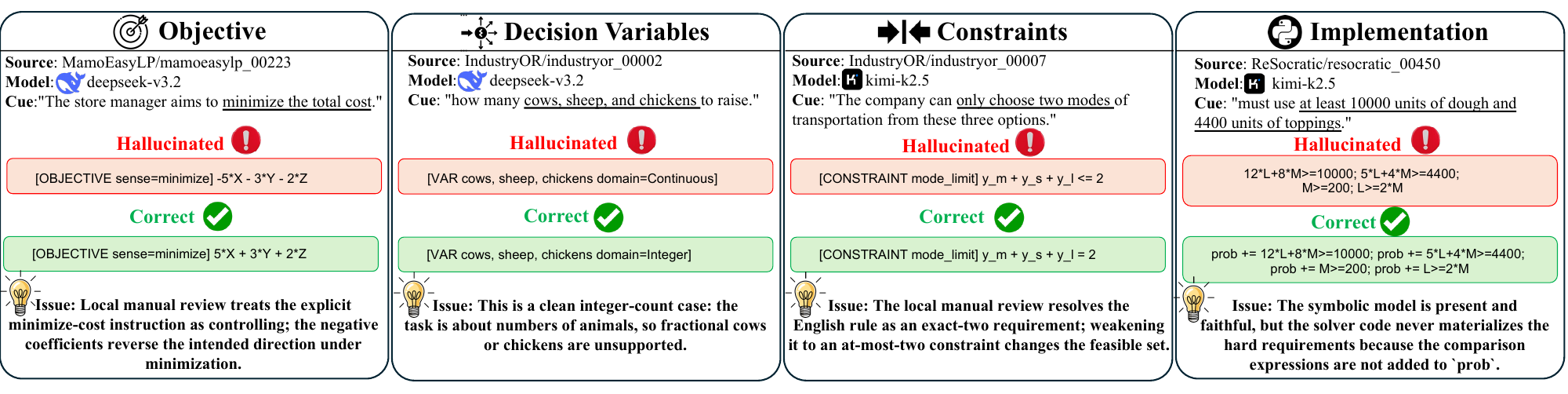}
  \caption{Representative optimization-modeling hallucinations. Each benchmark-derived panel contrasts a hallucinated fragment with its corrected counterpart and illustrates one taxonomy family: objective, variable, constraint, or implementation.}
  \label{fig:intro_hallucination_examples}
\end{figure}

Recent studies~\citep{xiao2024chain,ahmaditeshnizi2024optimus,huang2025orlm,lu2025optmath,chen2025solver} have explored the use of LLMs for automatic optimization modeling, where a model is prompted with a natural-language problem description  and asked to produce the corresponding mathematical formulation and solver-ready code. These systems have reported promising results on many benchmarks~\citep{li2026constructing}, suggesting that end-to-end LLM-based optimization modeling is promising. A common limitation of these studies, however, is that success is often judged primarily by whether the generated artifacts produce objective values consistent with the reference optimal value. For optimization modeling, this criterion is too weak: even when the optimal value matches, the mathematical formulation and solver code may still fail to faithfully encode the intended problem. In this paper, we refer to such structurally wrong but superficially plausible artifacts as \textbf{optimization-modeling hallucinations}. Figure~\ref{fig:intro_hallucination_examples} illustrates how this gap arises: the \textit{objective panel} reverses optimization intent through sign errors, the \textit{variable panel} relaxes an integer-count decision into a continuous one, the \textit{constraint panel} weakens an exact rule into an at-most rule, and the \textit{implementation panel} preserves the symbolic requirement in appearance while failing to materialize it in solver code. These artifacts can therefore look plausible, executable, and even numerically correct on a benchmark instance while still altering the underlying optimization problem. Empirically, Table~\ref{tab:appendix_v484_generator_snapshot} (see the last column) shows that even among optimal-value matches, hallucinations remain at rates of up to $31.1\%$ for some evaluated models. In high-stakes settings, failing to detect such hallucinations can lead to systematically wrong allocations, infeasible or noncompliant plans, and unwarranted confidence in decision-support systems precisely where reliability matters most. 

This motivates a different view of LLM-based automatic optimization modeling. Rather than treating optimal-value agreement as the primary success criterion, we ask whether the generated symbolic model $M$ and solver implementation $S$ faithfully realize the optimization structure specified by the given natural-language problem description $P$. This perspective connects our setting to broader work on hallucination detection~\citep{huang2025survey,zhang2025siren,li-etal-2025-fg,liu2025enhancing}, while emphasizing a distinctive feature of optimization modeling: correctness depends on explicit mathematical objects whose meaning is distributed across symbolic and executable representations. Accordingly, we formalize  hallucination detection of optimization modeling as a structural consistency problem over $(P,M,S)$ and organize the error space through a fine-grained OR-oriented taxonomy developed through substantial OR-expert effort and iterative refinement. The taxonomy partitions failures into objective, variable, constraint, and implementation families and is used throughout the paper. Figure~\ref{fig:taxonomy-fishbone} gives the compact view, while Table~\ref{tab:hallucination-taxonomy-summary} and App.~\ref{app:taxonomy} provide the complete subtype inventory, together with precise definitions and illustrative examples for all $18$ objective types, $18$ variable types, $31$ constraint types, and $16$ implementation types.

Building on this formulation and taxonomy, we conduct a systematic quantitative study of hallucinations in current LLM-based optimization modeling systems. We construct a benchmark-and-evaluation suite under a unified expert-driven protocol, consisting of a \texttt{clean} benchmark of $484$ OR-expert-validated reference artifacts for measuring false alarms and restraint on correct cases, an \texttt{injected} benchmark of $1266$ controlled artifacts spanning the defensibly constructible specific hallucination types in the active seed universe, and a \texttt{natural} benchmark of $6292$ real LLM-generated artifacts fully labeled by OR experts for testing performance on naturally occurring hallucinations. We pair these benchmarks with task-specific auditing metrics that emphasize clean-case restraint, controlled localization, and natural hallucination detection rather than objective-value matching alone. On this testbed, we propose and compare two detectors for auditing completed optimization artifacts: a Single-Agent Detector that audits the full $(P,M,S)$ tuple in one pass, and OptArgus, a multi-agent detector that routes each case across specialized objective, variable, constraint, and implementation experts before consolidating their findings. Across the \texttt{clean}, \texttt{injected}, and \texttt{natural} benchmarks, OptArgus is more restrained on correct artifacts, localizes controlled errors more accurately, and detects natural hallucinations more effectively than the Single-Agent Detector while producing shorter reports. Together, these results show that optimization-modeling hallucinations can be studied rigorously and that modular, expert-aligned auditing is a practical path toward more reliable LLM-based optimization modeling.

In summary, our contributions are fivefold: (1) we formulate optimization-modeling hallucination detection as structural consistency analysis over problem descriptions, symbolic models, and executable solver artifacts (see Sec.~\ref{sec:problem}); (2) we develop a fine-grained OR-expert-built taxonomy that organizes objective, variable, constraint, and implementation failures (see Sec.~\ref{sec:problem} and App.~\ref{app:taxonomy}); (3) we introduce OptArgus, a taxonomy-grounded multi-agent detector for auditing completed optimization artifacts, together with a matched Single-Agent Detector baseline for direct comparison (see Sec.~\ref{sec:framework}); (4) we build a three-part OR-expert benchmark suite---\texttt{clean}, \texttt{injected}, and \texttt{natural}---under a unified protocol for measuring clean-case restraint, controlled localization, and natural hallucination detection (see Sec.~\ref{sec:benchmark} and App.~\ref{app:data_construction}); and (5) we provide a systematic empirical study showing that multi-agent auditing improves restraint on correct artifacts, localization on controlled errors, and detection on naturally occurring hallucinations (see Sec.~\ref{sec:experiments}).

\section{Related Work}
\label{sec:related_work}


\textbf{LLMs for Optimization Modeling.} Recent advances in LLMs have led to growing interest in natural-language-to-optimization modeling, where a verbal problem specification is translated into a mathematical model and executable solver code~\citep{xiao2025survey}. Representative systems include CoE~\citep{xiao2024chain}, OptiMUS~\citep{ahmaditeshnizi2024optimus}, ORLM~\citep{huang2025orlm}, OptMath~\citep{lu2025optmath}, LLMOPT~\citep{jiang2025llmopt}, and SIRL~\citep{chen2025solver}. These methods span prompt-based, agentic, and fine-tuned settings. CoE and OptiMUS are especially relevant because they use multi-agent to translate a natural-language problem into a symbolic model and solver code. The key difference is that their specialization serves \emph{artifact generation}, whereas OptArgus uses specialization for \emph{auditing} completed optimization artifacts. In addition, success in prior work is primarily judged by objective-value agreement, lacking explicit hallucination diagnosis.

\textbf{Existing Natural-Language-to-Optimization Benchmarks and Evaluation.}
Progress on these systems has been closely tied to benchmarks such as NL4Opt~\citep{Ramamonjison2023NL4Opt}, IndustryOR~\citep{huang2025orlm}, MAMO~\citep{huang2025mamo}, OptiBench~\citep{wang2024optibench}, ReSocratic~\citep{yang2024optibench}, OptMath~\citep{lu2025optmath}, Bench4Opt~\citep{wang2025orgeval},  ProOPF~\citep{shen2026proopf}, and MIPLIB-NL~\citep{li2026constructing}. These resources have been valuable for generation research, but they are built around problem statements, reference answers, or reference formulations rather than hallucination auditing. Their evaluations typically emphasize objective gap, compilation success, textual similarity, or coarse formulation matching. \citet{refai2025peering} move beyond such coarse metrics with a component-level evaluation framework for LLM-generated optimization formulations, including precision and recall for decision variables and constraints, objective and constraint RMSE, optimality gap, and efficiency measures, and they use it to compare models and prompting strategies. However, their framework remains formulation-centric: it measures deviation from ground-truth formulations rather than defining hallucination detection as a taxonomy-grounded auditing task over the full $(P,M,S)$ tuple, and it does not introduce OR-expert-labeled hallucination benchmarks, explicit symbolic-versus-implementation divergence labels, or detectors for localizing failure modes across optimization-modeling modules.

\textbf{Hallucination Detection and Mathematical Reasoning.}
Hallucination in general LLM systems has been studied through factuality, faithfulness, retrieval, uncertainty estimation, self-evaluation, and evidence-based verification \citep{huang2025survey,zhang2025siren,sahoo2024comprehensive}. In mathematical reasoning, recent work emphasizes process-level supervision and fine-grained error diagnosis \citep{li-etal-2025-fg,liu2025enhancing,wang2026survey,lightman2024lets,luo2024improve}. Optimization modeling differs in a crucial way: errors are tied to explicit mathematical objects whose meaning depends on variable definitions, index sets, feasibility structure, and solver implementation, making the problem less about factual verification of free-form text and more about structural consistency across coupled symbolic and executable artifacts. To the best of our knowledge, prior work does not provide either of the hallucination detectors studied here: both the Single-Agent Detector and OptArgus are introduced in this paper.

{\color{black}
\section{Problem Formulation and Hallucination Taxonomy}
\label{sec:problem}

We formalize optimization-modeling hallucination detection as structural consistency auditing over three coupled artifacts: a natural-language problem description $P$, a symbolic model $M$, and a solver implementation $S$. The key point is that correctness is structural rather than merely behavioral: a model can match a reference objective value on one instance while still omitting a nonbinding constraint, or relaxing an integer variable. Semantically equivalent reformulations are not hallucinations, even when they differ syntactically from a reference formulation.

\noindent\textbf{Artifacts and consistency.} Let $M=(O,V,C,A)$, with $O$ the objective, $V$ the decision variables, $C$ the constraints, and $A$ auxiliary objects such as parameters, sets, and indices. We audit the tuple $G=(P,M,S)$. Let $\mathcal{R}_{\mathrm{sym}}(P,M)\in\{0,1\}$ indicate whether $M$ faithfully encodes the optimization semantics of $P$, and let $\mathcal{R}_{\mathrm{imp}}(M,S)\in\{0,1\}$ indicate whether $S$ faithfully realizes $M$. Overall structural consistency is
$
\mathcal{R}(P,M,S)=
\mathcal{R}_{\mathrm{sym}}(P,M)\land
\mathcal{R}_{\mathrm{imp}}(M,S).
$
Objective-value agreement is only instance-level evidence: matching the certified value does not imply that $(P,M,S)$ is structurally correct.
This decomposition separates symbolic failures, such as translating an ``exactly two'' rule into ``at most two,'' from implementation failures, such as omitting a constraint family in code.

\noindent\textbf{Hallucination taxonomy.} Because prior optimization-modeling benchmarks~\citep{huang2025mamo,li2026constructing} mainly evaluate generation quality via objective-value agreement, they do not provide a systematic label space for structural hallucinations over $(P,M,S)$. We therefore construct a hallucination taxonomy for optimization modeling through OR-expert manual analysis and iterative refinement. The resulting taxonomy partitions hallucinations into objectives ($\mathcal{T}^{O}$), decision variables ($\mathcal{T}^{V}$), constraints ($\mathcal{T}^{C}$), and implementation ($\mathcal{T}^{I}$) families.  We treat these four families as disjoint label spaces and write
$
\mathcal{T}=
\mathcal{T}^{O}\sqcup
\mathcal{T}^{V}\sqcup
\mathcal{T}^{C}\sqcup
\mathcal{T}^{I}.
$ Figure~\ref{fig:taxonomy-fishbone} gives the compact fishbone view, while the full subtype inventory, definitions, and examples appear in Table~\ref{tab:hallucination-taxonomy-summary} and App.~\ref{app:taxonomy}.

\begin{figure}[h]
\centering
\includegraphics[width=0.88\linewidth]{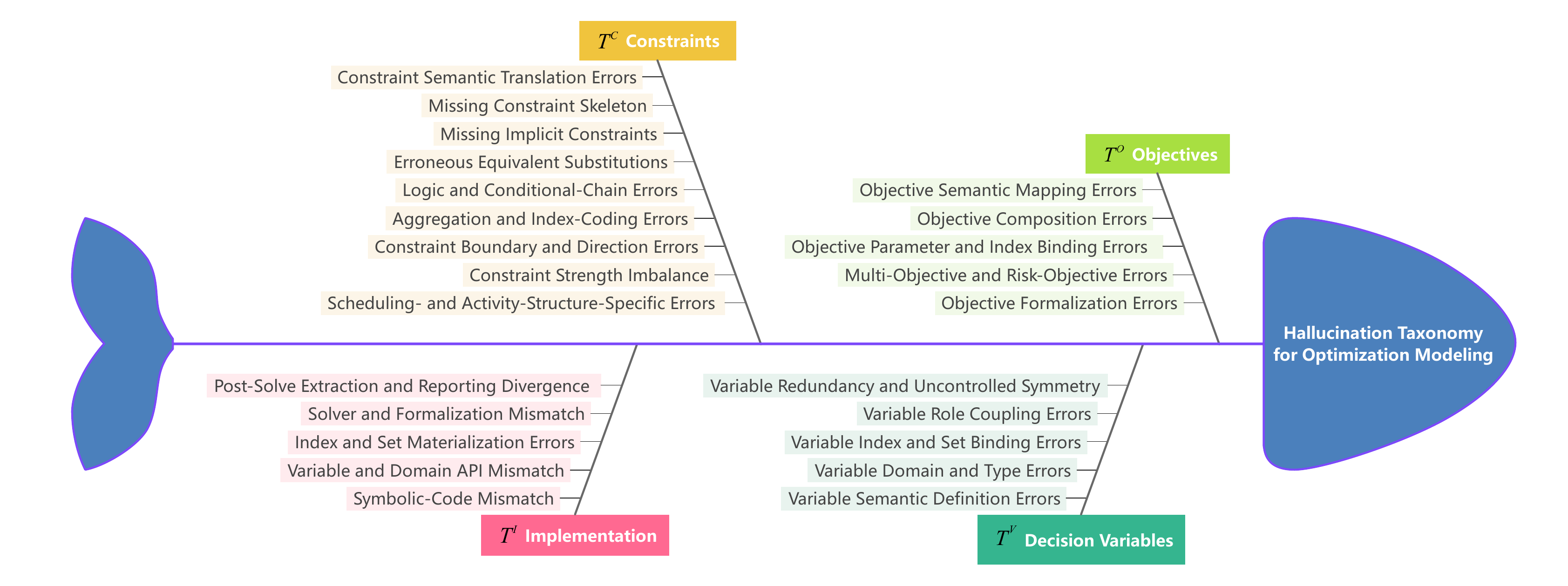}
\caption{Compact fishbone view of the OR-expert-built optimization-modeling hallucination taxonomy. The complete subtype inventory, definitions, and examples are given in Table~\ref{tab:hallucination-taxonomy-summary} and App.~\ref{app:taxonomy}.}
\label{fig:taxonomy-fishbone}
\end{figure}



\noindent\textbf{Taxonomy-grounded detection task.} Let $E^{O}$, $E^{V}$, and $E^{C}$ denote the objective, variable, and constraint elements identified during auditing. For $k\in\{O,V,C\}$, $E^{k}$ includes both elements explicitly present in the generated symbolic model $M$ and expected elements inferred from $P$ whose absence may constitute an omission. Let $E^{I}$ denote implementation elements identified in $S$, together with expected solver-side counterparts aligned to symbolic elements in $M$. We treat
$
E
=
E^{O}
\sqcup
E^{V}
\sqcup
E^{C}
\sqcup
E^{I}
$
as a module-indexed disjoint union.
For each element $e\in E$, let $m(e)\in\{O,V,C,I\}$ denote its family; for example, $m(e)=V$ if $e$ is a variable declaration. The notation $\mathcal{T}^{m(e)}$ then means the subset of taxonomy labels that is valid for that family, such as variable hallucination types when $m(e)=V$. A candidate hallucination is a pair $(e,t)$, where $t\in\mathcal{T}^{m(e)}$ is the specific taxonomy type assigned to element $e$. Finally, let $y(e,t)\in\{0,1\}$ be the gold indicator of whether element $e$ exhibits hallucination type $t$, including omission-type failures for expected elements. The gold hallucination set is
$
\mathcal{H}(G)=\{(e,t):e\in E,\ t\in\mathcal{T}^{m(e)},\ y(e,t)=1\}.
$
The tuple is hallucinated iff $\mathcal{H}(G)\neq\varnothing$; equivalently, structural consistency corresponds to $\mathcal{H}(G)=\varnothing$. The detector predicts a possibly empty set
$
\widehat{\mathcal{H}}(G)\subseteq
\mathcal{H}(G),
$
thereby deciding both whether a hallucination exists and where it occurs. This makes hallucination detection a structured prediction problem over taxonomy-grounded element labels, not a single binary judgment on the final solution value.
}

\section{Methodology: Taxonomy-Grounded Multi-Agent Detection Framework}
\label{sec:framework}

Optimization-modeling hallucination detection is not a uniform checking task: the system must interpret the business semantics in $P$, inspect the symbolic structure of $M$, and verify whether the solver code $S$ faithfully realizes that structure. Because these checks rely on different evidence and often interact across modules, we adopt a multi-agent design rather than a single monolithic prompt. Here, a \emph{multi-agent} system means a coordinated LLM system in which role-specialized agents cooperate through routing and shared intermediate state~\citep{guo2024large}. OptArgus instantiates this idea with a conductor, four specialists aligned with objective, variable, constraint, and implementation failures, and a visualization agent. For comparison, we also build a monolithic Single-Agent Detector, documented in App.~\ref{app:single_agent_prompt}; Sec.~\ref{sec:experiments} shows that OptArgus substantially improves clean-case restraint, controlled localization, report conciseness, and natural-benchmark robustness relative to this non-multi-agent baseline. The organization of the rest of this section follows the logic of the detector itself. Sec.~\ref{subsec:ma_formulation} first defines the shared execution framework, namely how agents interact and what state they share. Sec.~\ref{subsec:specialists} then explains what each specialist branch actually checks and what kind of local findings it produces. Sec.~\ref{subsec:aggregation} finally explains how those branch-local findings are coordinated, consolidated, reranked, and turned into the final diagnosis. In other words, the order moves from system skeleton, to local branch reasoning, to global decision.

\begin{figure}[h]
    \centering
    \includegraphics[width=1\linewidth]{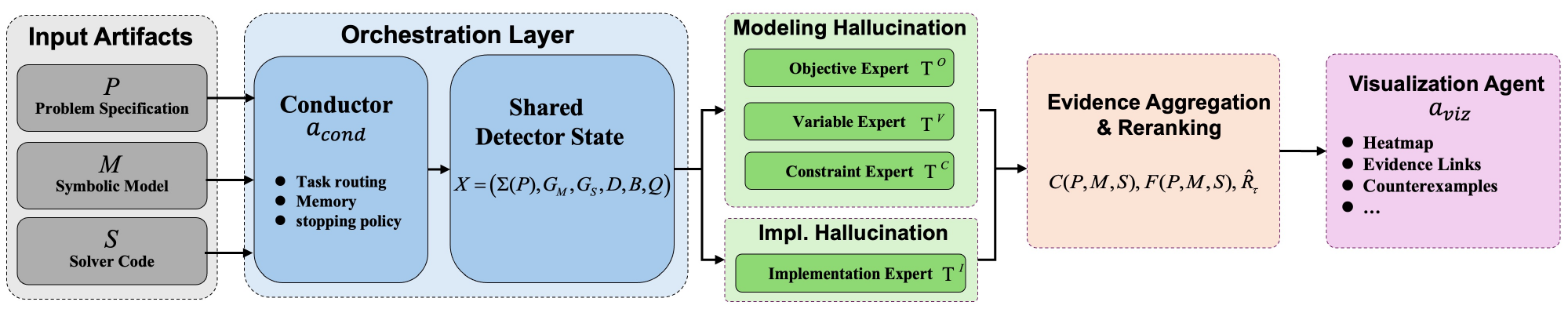}
    \caption{OptArgus workflow. A \textit{conductor} routes the audit tuple to four specialized experts---objective, variable, constraint, and implementation---which audit the artifact from their respective perspectives and exchange optional cross-agent critiques over shared dependencies. Their structured findings are then consolidated and reranked before a visualization agent emits the final auditable report. A more detailed system diagram appears in App.~\ref{app:engineering}.}
    \label{fig:OptArgus}
\end{figure}

\subsection{Shared Execution Framework}
\label{subsec:ma_formulation}

We describe OptArgus at three levels: the end-to-end workflow, the shared state that carries information across agents, and the execution rule that updates that state.

\textbf{Workflow.} Figure~\ref{fig:OptArgus} summarizes OptArgus (App.~\ref{app:engineering} gives the full diagram). In brief, the \textit{conductor} first performs case triage on the audited tuple $(P,M,S)$ and selects the most relevant branches. The \textit{selected specialists} then inspect the artifact from their own perspectives and write findings into shared state; when a local diagnosis depends on another branch, the conductor requests targeted follow-up review. Finally, the accumulated findings are consolidated into a short diagnosis list and an auditable report. This coordination avoids unnecessary fan-out on easy cases while still enabling cross-module verification on difficult ones. App.~\ref{app:routing} details the routing policy.

\textbf{Shared-State Abstraction.} We represent OptArgus as a directed state graph over six agents with shared state $X^{(k)}=\left(\Sigma(P),G_M,G_S,D,\mathcal{B}^{(k)},Q^{(k)}\right)$ at iteration $k$. The shared state stores the semantic schema of the problem ($\Sigma(P)$), symbolic and implementation graphs ($G_M$ and $G_S$), cross-artifact dependencies ($D$), accumulated evidence ($\mathcal{B}^{(k)}$), and the current task queue ($Q^{(k)}$). App.~\ref{app:canonical_ir} gives the concrete representations behind these objects.

\textbf{Execution Rule.} At step $k$, the \textit{conductor} selects an active set $\Pi^{(k)}$, the \textit{selected agents} emit branch-specific updates, and the \textit{shared state} is merged into $X^{(k+1)}$ until convergence or a fixed interaction budget is reached. Our prototype is implemented in LangGraph, but the formulation itself is framework-agnostic; Apps.~\ref{app:canonical_ir} and~\ref{app:detector_algorithm} give the runtime view and algorithmic summary.

\subsection{Specialist Branches and Local Outputs}
\label{subsec:specialists}

We now specify components that populate this shared framework and the local findings they return.

\textbf{Expert Roles and Outputs.} OptArgus decomposes diagnosis into four specialist branches that mirror the four major categories of the taxonomy. This alignment lets each branch focus on the evidence most natural for its own failure mode while still producing outputs in a shared structured format. Each branch returns a small set of candidates consisting of an implicated element, a proposed taxonomy label, a bounded support score, and an evidence bundle. The first three branches primarily audit symbolic modeling choices against the problem description, whereas the implementation branch audits symbolic-to-code fidelity. Concretely, the four branches play the following roles:
\begin{itemize}[leftmargin=*,itemsep=2pt,topsep=2pt,parsep=0pt,partopsep=0pt]
    \item \textbf{Objective expert.} This branch checks whether the optimization target in the artifact matches the intended goal in the problem description, such as objective direction, term composition, and code-level realization. The aligned subtype inventory is given in App.~\ref{app:subsec:obj_mod_hal}.
    \item \textbf{Variable expert.} This branch checks whether the artifact defines the right decision objects and domains, with particular attention to dimensional structure and discrete-versus-continuous interpretation. The aligned subtype inventory is given in App.~\ref{app:subsec:dec_var_hal}.
    \item \textbf{Constraint expert.} This branch checks whether the feasible-region logic preserves the intended business rules, including but not limited to missing skeletons, weakened rules, aggregation mistakes, and boundary conditions. The aligned subtype inventory is given in App.~\ref{app:subsec:con_mod_hal}.
    \item \textbf{Implementation expert.} This branch checks whether the solver code faithfully realizes the symbolic model, targeting issues such as dropped indices, wrong objective sense, omitted materialization, incompatible solver settings, and misreported outputs while suppressing semantically equivalent code patterns. The aligned subtype inventory is given in App.~\ref{app:subsec:imp_hal}.
\end{itemize}

The runtime details and exact prompts of each expert are deferred to App.~\ref{app:nodes} and App.~\ref{app:multi_agent_prompts}.

\subsection{Coordination, Consolidation, and Final Decision}
\label{subsec:aggregation}

We now explain how OptArgus converts branch-local judgments into a system-level diagnosis through conductor coordination, candidate consolidation, deterministic reranking, and open-set abstention.

\textbf{Coordination.} OptArgus does not simply call all experts once and average their outputs. Instead, the \textit{conductor} uses the dependency structure to decide which branch should act next and when a second opinion is needed. This dynamic routing is essential because many hallucinations span more than one module. App.~\ref{app:routing} gives the detailed routing and cross-agent review policy.

\textbf{Candidate Consolidation.} The selected branches first produce a high-recall pool of structured candidates grounded in problem text, symbolic fragments, solver-code anchors, and cross-agent notes. OptArgus then applies deterministic normalization to resolve aliases, suppress duplicates, and remove derivative warnings whose evidence is already subsumed by a more specific root-cause candidate. App.~\ref{app:parsing_tips} gives the candidate schema and consolidation rules.

\textbf{Deterministic Reranking and Abstention.} We next apply a frozen deterministic reranker that turns the consolidated candidate pool into a short ranked diagnosis list by favoring stronger support, higher severity, better route consistency, and more specific root-cause explanations. This stage is fixed post-processing rather than a learned model; App.~\ref{app:parsing_tips} summarizes the qualitative calibration policy. The final output is a short ranked list $F(P,M,S)$; if no candidate survives, OptArgus abstains. This open-set behavior is central to the clean-case restraint measured in Sec.~\ref{subsec:controlled_results}.

\textbf{Final Report.} The \textit{visualization} agent converts the surviving findings and evidence bundles into an auditable report for benchmarking, diagnosis, and human repair. App.~\ref{app:detector_algorithm} gives a compact algorithmic summary, and again App.~\ref{app:engineering} shows the full system-level realization.

\section{Experiment Setup}
\label{sec:benchmark}

This section specifies the empirical questions, evaluation suite, metrics, and compared methods. Because existing optimization-modeling benchmarks \citep{li2026constructing} emphasize generation or answer matching rather than auditing completed $(P,M,S)$ tuples, we introduce both a three-benchmark evaluation suite and a matched single-agent baseline tailored to hallucination detection. The code, data, and benchmark artifacts are not publicly released at this stage; we will make them public upon acceptance.

\noindent\textbf{Research Questions.} The empirical study is organized around four research questions: \textbf{RQ1.} Can OptArgus avoid false alarms on correct artifacts while still pinpointing the injected error when one is present? \textbf{RQ2.} Can OptArgus effectively detect naturally occurring hallucinations in real LLM-generated optimization artifacts? \textbf{RQ3.} Does multi-agent diagnosis provide consistent gains over a comparable single-agent baseline? and \textbf{RQ4.} Which design choices account for OptArgus's gains, and do the gains persist under an alternative backbone LLM?

\noindent\textbf{Benchmark Suite Construction and Metrics.} The benchmark suite is designed to separate three questions that are often mixed together in optimization-modeling evaluation: whether a detector stays quiet on correct artifacts, whether it can localize a known injected error, and whether it still works on naturally occurring hallucinations in real model outputs. Starting from a common seed corpus, we first certify a clean reference pool, then build a controlled injected benchmark from it, and finally expand the same seeds into a natural benchmark of real LLM-generated artifacts. Full construction protocols appear in App.~\ref{app:clean_construction}, App.~\ref{app:injected_construction}, and App.~\ref{app:natural_construction}, and the formal metric definitions appear in App.~\ref{app:clean_metrics}, App.~\ref{app:injected_metrics}, and App.~\ref{app:natural_metrics}. In the main text, we keep only the benchmark role, scale, and the intuitive meaning of the reported metrics.

\begin{itemize}[leftmargin=*,itemsep=2pt,topsep=3pt,parsep=0pt,partopsep=0pt]
\item \textbf{Clean benchmark.} We first curate $484$ OR-expert-validated reference artifacts from five benchmark families. This benchmark isolates restraint on correct artifacts: can a detector stay quiet when nothing is wrong? We therefore report \textit{EmptyReportRate} and \textit{MeanFindings$_{\mathrm{clean}}$}. \textit{EmptyReportRate}\,$\in[0,1]$ asks how often the detector returns no finding at all on a clean artifact, so higher is better. \textit{MeanFindings$_{\mathrm{clean}}$}\,$\in[0,\infty)$ counts the average number of findings on clean cases, so lower is better. The two metrics are complementary: one measures abstention frequency, the other false-alarm verbosity once abstention fails.

\item \textbf{Injected benchmark.} Starting from the same clean references, we build a taxonomy-complete controlled benchmark, \texttt{injected}, by instantiating one specific hallucination type at a time from the inventory in App.~\ref{app:taxonomy}. The evaluated version contains $1266$ controlled artifacts, and each case has an exact gold label at all three taxonomy levels in Table~\ref{tab:hallucination-taxonomy-summary}: major category, subcategory, and specific hallucination type, together with artifact-level hallucination presence. This benchmark isolates precise localization under single-error supervision. We report \textit{Top-1 MajorCategoryHit}, \textit{Top-1 SubcategoryHit}, \textit{Top-1 SpecificTypeHit}, and \textit{MeanFindings}. The three Top-1 metrics lie in $[0,1]$ and ask whether the detector's first returned diagnosis already matches the gold major category, subcategory, or specific hallucination type; higher is better. \textit{MeanFindings}\,$\in[0,\infty)$ measures report length, so lower is better if localization quality is preserved.

\item \textbf{Natural benchmark.} Controlled injection gives exact supervision but not the open-ended error patterns of real model outputs. We therefore expand the same $484$ seeds across a finalized thirteen-model panel that includes both frontier general-purpose models and fine-tuned models for optimization modeling, and manually annotate the resulting $6292$ artifacts with OR-expert artifact-level and major-category labels. This \texttt{natural} benchmark tests detection quality under realistic, overlapping, model-dependent failures. We evaluate it with seven F1 scores in $[0,1]$ (higher is better): artifact-level \textit{Halluc-F1}; four major-category F1 scores for objective, variable, constraint, and implementation failures; and the aggregate summaries \textit{MajorCategoryMacro-F1} and \textit{MajorCategoryMicro-F1}, which average the four major-category scores uniformly and by the empirical label distribution, respectively.
\end{itemize}

\noindent\textbf{Baselines and Compared Methods.} The comparison setup is deliberately matched. Existing systems are designed either to generate optimization artifacts from natural-language inputs \citep{ahmaditeshnizi2024optimus} or to measure discrepancies between generated and reference formulations \citep{refai2025peering}, rather than to diagnose hallucinations in a completed $(P,M,S)$ artifact. As a result, there is no off-the-shelf detector that can be evaluated directly under our taxonomy and output protocol. We therefore compare OptArgus against the strongest directly comparable non-multi-agent baseline we can construct under the same inputs, labels, and scoring rules.

\begin{itemize}[leftmargin=*,itemsep=2pt,topsep=3pt,parsep=0pt,partopsep=0pt]

\item \textbf{Multi-Agent Detector (OptArgus).} Our main method is the taxonomy-grounded multi-agent detector introduced in Sec.~\ref{sec:framework}. A conductor routes each case to specialized objective, variable, constraint, and implementation experts, and a consolidation layer turns their findings into a short ranked diagnosis list. The full prompt stack appears in App.~\ref{app:multi_agent_prompts}.

\item \textbf{Single-Agent Detector (Baseline).} The baseline audits the full $(P,M,S)$ tuple in a single LLM call. It uses the same taxonomy, artifact triple, abstention-oriented output schema, and scoring protocol as OptArgus. The only intended change is architectural: branch specialization, cross-module reconciliation, and final labeling must all be handled in one monolithic pass. The exact prompt appears in App.~\ref{app:single_agent_prompt}.

\item \textbf{Diagnostic Variants for RQ4.} To attribute the gains of OptArgus, we additionally evaluate component ablations that remove or alter specialist calibration, routing rescue, all-expert fan-out, and final reranking (see App.~\ref{app:component_ablations} for more details), plus a same-protocol backbone-sensitivity comparison under an alternative LLM (see App.~\ref{app:backbone_sensitivity} for more details).

\end{itemize}

\section{Experimental Results, Analysis, and Discussion}
\label{sec:experiments}

We now answer the four research questions posed in Sec.~\ref{sec:benchmark}. The clean and controlled benchmarks give the clearest view of calibration and exact localization; the natural benchmark then tests whether the same advantage survives on real LLM-generated artifacts; and the ablation/backbone analyses examine which design choices drive the effect and whether it is backbone-specific.
\subsection{Clean and Controlled Results}
\label{subsec:controlled_results}

As shown in Table~\ref{tab:v484_rq1_main}, we first address RQ1 and the controlled side of RQ3. Panel A shows a large calibration gain on certified-correct artifacts: OptArgus raises EmptyReportRate from $0.483$ to $0.853$ while reducing mean clean-case findings from $0.556$ to $0.159$. This matters because a detector that fabricates plausible errors on correct artifacts quickly becomes unusable in debugging or repair loops.

Panel B asks the other half of RQ1: once an error is present, can the detector identify it correctly? On the 1266-case \texttt{injected} benchmark, OptArgus improves all three Top-1 localization metrics while also returning shorter reports. Top-1 MajorCategoryHit rises from $0.723$ to $0.767$, Top-1 SubcategoryHit from $0.431$ to $0.473$, and Top-1 SpecificTypeHit from $0.339$ to $0.403$, while MeanFindings drops from $2.151$ to $1.224$. This combination matters: the system finds the right error more often without padding the report with extra guesses.

Taken together, the clean and controlled results answer RQ1 positively and provide a clear affirmative result for RQ3: multi-agent decomposition improves both restraint and localization.

\begin{table*}[h]
\centering
\caption{Results on the clean and controlled benchmarks; blue means better under the metric direction.}
\label{tab:v484_rq1_main}
\small
\resizebox{\textwidth}{!}{%
\begin{tabular}{lL{2.45cm}L{2.75cm}|L{2.75cm}L{2.6cm}L{2.75cm}L{2.35cm}}
\toprule
& \multicolumn{2}{c|}{\textbf{Panel A. Clean Benchmark (484 cases)}} & \multicolumn{4}{c}{\textbf{Panel B. Controlled Injected Benchmark (1266 cases)}} \\
\cmidrule(lr){2-3}\cmidrule(lr){4-7}
\multicolumn{1}{c}{Method} & \multicolumn{1}{c}{EmptyReportRate $\uparrow$} & \multicolumn{1}{c|}{MeanFindings$_{\mathrm{clean}}$ $\downarrow$} & \multicolumn{1}{c}{Top-1 MajorCategoryHit $\uparrow$} & \multicolumn{1}{c}{Top-1 SubcategoryHit $\uparrow$} & \multicolumn{1}{c}{Top-1 SpecificTypeHit $\uparrow$} & \multicolumn{1}{c}{MeanFindings $\downarrow$} \\
\midrule
Single-Agent Detector  & 0.483 & 0.556 & 0.723 & 0.431 & 0.339 & 2.151 \\
OptArgus & \textbf{0.853} \metricdelta{+0.370} & \textbf{0.159} \metricdelta{-0.397} & \textbf{0.767} \metricdelta{+0.044} & \textbf{0.473} \metricdelta{+0.042} & \textbf{0.403} \metricdelta{+0.064} & \textbf{1.224} \metricdelta{-0.927} \\
\bottomrule
\end{tabular}%
}
\end{table*}

\subsection{Natural Results and Analysis}
\label{subsec:natural_results}

As shown in Table~\ref{tab:natural_v484_full_overlap}, we next address RQ2 and the natural side of RQ3. The main pattern is simple: OptArgus improves every reported natural-benchmark metric. Halluc-F1 rises from $0.521$ to $0.617$; Objective-F1, Variable-F1, Constraint-F1, and Implementation-F1 all improve; and both summary metrics rise as well, with MajorCategoryMacro-F1 increasing from $0.382$ to $0.512$ and MajorCategoryMicro-F1 from $0.379$ to $0.486$.

\begin{table}[h]
\centering
\caption{Natural-benchmark results on $6292$ OR-expert-labeled artifacts. Supports: objective $92$, variable $1126$, constraint $571$, implementation $675$. Blue means better under the metric direction.}
\label{tab:natural_v484_full_overlap}
\small
\resizebox{\linewidth}{!}{%
\begin{tabular}{llllllll}
\toprule
\multicolumn{1}{c}{\multirow{2}{*}{Method}} & \multicolumn{1}{c}{\multirow{2}{*}{Halluc-F1 $\uparrow$}} & \multicolumn{1}{c}{Objective-F1 $\uparrow$} & \multicolumn{1}{c}{Variable-F1 $\uparrow$} & \multicolumn{1}{c}{Constraint-F1 $\uparrow$} & \multicolumn{1}{c}{Implementation-F1 $\uparrow$} & \multicolumn{1}{c}{MajorCategory} & \multicolumn{1}{c}{MajorCategory} \\
& & ($n{=}92$) & ($n{=}1126$) & ($n{=}571$) & ($n{=}675$) & \multicolumn{1}{c}{Macro-F1 $\uparrow$} & \multicolumn{1}{c}{Micro-F1 $\uparrow$} \\
\midrule
Single-Agent Detector & 0.521 & 0.404 & 0.328 & 0.504 & 0.294 & 0.382 & 0.379 \\
OptArgus & \textbf{0.617} \metricdelta{+0.096} & \textbf{0.541} \metricdelta{+0.137} & \textbf{0.495} \metricdelta{+0.167} & \textbf{0.579} \metricdelta{+0.075} & \textbf{0.431} \metricdelta{+0.137} & \textbf{0.512} \metricdelta{+0.130} & \textbf{0.486} \metricdelta{+0.107} \\
\bottomrule
\end{tabular}%
}
\end{table}

Three points are worth emphasizing: (1) \textit{the gains carry over from controlled single-error cases to real model outputs}. Natural artifacts can be correct, can contain one or more errors, and they may mix symbolic and implementation problems in the same case. OptArgus still improves every metric, which suggests that the method is not tied only to the cleaner single-error setting; (2) \textit{the largest major-category gains appear on variable, objective, and implementation errors}. Variable-F1 rises from $0.328$ to $0.495$, Objective-F1 from $0.404$ to $0.541$, and Implementation-F1 from $0.294$ to $0.431$, while Constraint-F1 also rises from $0.504$ to $0.579$. This likely reflects OptArgus's specialist design: variable errors depend on variable-definition checks, objective errors depend on semantic alignment between the stated goal and the formal objective, and implementation errors depend on symbolic-to-code consistency checks; and (3) \textit{the improvement is not driven only by the most common categories}. The natural label distribution is highly imbalanced, with only $92$ objective positives but $1126$ variable positives. If the gains came only from common labels, Micro-F1 could rise without much change in Macro-F1. Instead, both summary metrics improve substantially, showing that the detector is stronger overall and also more balanced across major categories.

These observations answer RQ2 positively and strengthen the answer to RQ3. The benefit of routed multi-agent diagnosis is not limited to clean laboratory-style perturbations; it also appears on realistic LLM outputs with uneven, overlapping error patterns. App.~\ref{app:natural_construction} gives  per-model analysis.

\subsection{\texorpdfstring{Ablations and Backbone Sensitivity}{Ablations and Backbone Sensitivity}}
\label{subsec:ablation_backbone_results}

We finally address RQ4 through component ablations and a backbone-sensitivity analysis, with detailed setups in App.~\ref{app:component_ablations} and App.~\ref{app:backbone_sensitivity}. The component ablations in Table~\ref{tab:ablation_full} show that specialist calibration, routing rescue, and final deterministic reranking all contribute to the reported gains. The OptArgus with All Experts variant slightly improves some  metrics, but Table~\ref{tab:ablation_efficiency} shows the cost of exhaustive fan-out: routed OptArgus uses $53.6\%$ fewer specialist calls and $33.7\%$ fewer total tokens.

All main experiments and component ablations use DeepSeek-V3.2 as the shared backbone for both the Single-Agent Detector and OptArgus. To test whether the conclusion depends on this choice, Tables~\ref{tab:backbone_sensitivity_clean_injected} and~\ref{tab:backbone_sensitivity_natural} repeat the matched comparison with Qwen3-Max-Preview as the backbone for both methods. The same-backbone comparison preserves the main conclusion: OptArgus remains stronger across the tracked clean, injected, and natural metrics, although the controlled Top-1 SpecificTypeHit margin is small and should be interpreted conservatively.

\textbf{Discussion and Takeaways.} Several broader lessons follow from the full evaluation: (1) optimization-modeling hallucination detection cannot be reduced to answer matching alone. Across the \texttt{clean}, \texttt{injected}, and \texttt{natural} benchmarks, the central reliability question is whether the full tuple $(P,M,S)$ remains structurally aligned, not whether the final objective value happens to agree on one instance; (2) the error space is genuinely structured. The controlled \texttt{injected} benchmark shows that localization quality matters at multiple taxonomy levels, while the \texttt{natural} benchmark shows that variable, constraint, and implementation failures dominate real outputs far more than simple objective mistakes; (3) multi-agent decomposition helps most when it is paired with explicit consolidation rather than treated as a loose collection of opinions. The results do not show a tradeoff in which OptArgus becomes more accurate only by becoming more verbose. Instead, it abstains more on the \texttt{clean} benchmark, localizes better with fewer findings on the controlled \texttt{injected} benchmark, and improves both aggregate and balanced major-category F1 on the \texttt{natural} benchmark. The most plausible interpretation is that specialist analysis produces sharper local evidence, while consolidation and reranking keep that extra evidence from turning into noisy reports; and  (4) the practical implication is straightforward. Future optimization-modeling systems should not be evaluated only by execution success or objective agreement, and future detectors should not be judged only by aggregate artifact-level accuracy. A reliable protocol should test at least three axes: clean-case restraint, controlled localization, and natural-distribution detection.

\section{Conclusion}

We present a taxonomy-grounded view of hallucination in LLM-based optimization modeling and a corresponding multi-agent detection framework that audits the full tuple of problem description, symbolic formulation, and solver implementation. To evaluate this setting, we also introduce a three-part benchmark suite that separates clean-case calibration, controlled taxonomy-level localization, and natural-distribution detection under OR-expert annotation. Across this protocol, OptArgus consistently improves over the matched single-agent baseline: it is more restrained on certified-correct artifacts, more precise and more concise on controlled injected cases, and stronger on natural hallucination detection at both the aggregate and major-category levels. More broadly, the results support two conclusions: optimization-modeling reliability cannot be judged adequately by execution success or objective agreement alone, and modular specialist auditing is most effective when paired with explicit consolidation and abstention control. This combination of formulation, benchmark suite, and detector design provides a stronger foundation for trustworthy optimization-modeling systems and a practical template for future work on verification, diagnosis, and repair. Limitations and broader impacts are discussed in App.~\ref{app:limitations_impacts}.

\bibliographystyle{ACM-Reference-Format}
\bibliography{References}

\clearpage
{\large{\textbf{Appendix for \textit{OptArgus: A Multi-Agent System to Detect Hallucinations in LLM-based Optimization Modeling}}}}

\appendix
\section*{\large Table of Contents}
{\footnotesize
\begin{itemize}
    \item[A] \hyperref[app:engineering]{Detector Engineering Details and Prompt Materials}
    \begin{itemize}
        \item[A.1] \hyperref[app:deployment]{Roadmap and Design Principles}
        \item[A.2] \hyperref[app:canonical_ir]{Representation and Shared State}
        \item[A.3] \hyperref[app:nodes]{Agent Roles and Responsibilities}
        \item[A.4] \hyperref[app:routing]{Routing and Cross-Agent Coordination}
        \item[A.5] \hyperref[app:parsing_tips]{Evidence Aggregation, Calibration, and Abstention}
        \item[A.6] \hyperref[app:detector_algorithm]{End-to-End Algorithm}
        \item[A.7] \hyperref[app:multi_agent_prompts]{OptArgus Prompt Stack}
        \item[A.8] \hyperref[app:single_agent_prompt]{Single-Agent Detector}
    \end{itemize}
    \item[B] \hyperref[app:data_construction]{Benchmark Construction, OR-Expert Annotation, and Metrics}
    \begin{itemize}
        \item[B.1] \hyperref[app:data_obj_match]{Why Objective Agreement Is Not Enough}
        \item[B.2] \hyperref[app:clean_benchmark]{Clean Benchmark}
        \begin{itemize}
            \item[B.2.1] \hyperref[app:clean_construction]{Construction and OR-Expert Validation}
            \item[B.2.2] \hyperref[app:clean_metrics]{Metrics}
        \end{itemize}
        \item[B.3] \hyperref[app:injected_benchmark]{Injected Benchmark}
        \begin{itemize}
            \item[B.3.1] \hyperref[app:injected_construction]{Construction and OR-Expert Error Injection}
            \item[B.3.2] \hyperref[app:injected_metrics]{Metrics}
        \end{itemize}
        \item[B.4] \hyperref[app:natural_benchmark]{Natural Benchmark}
        \begin{itemize}
            \item[B.4.1] \hyperref[app:natural_construction]{Model Panel and OR-Expert Annotation}
            \item[B.4.2] \hyperref[app:natural_metrics]{Metrics}
        \end{itemize}
    \end{itemize}
    \item[C] \hyperref[app:ablation_backbone]{Ablation and Backbone Sensitivity Details}
    \begin{itemize}
        \item[C.1] \hyperref[app:component_ablations]{Component Ablations}
        \item[C.2] \hyperref[app:backbone_sensitivity]{Backbone Sensitivity}
    \end{itemize}
    \item[D] \hyperref[app:taxonomy]{Details on Hallucination Taxonomy}
    \begin{itemize}
        \item[D.1] \hyperref[app:subsec:obj_mod_hal]{Objective-Function Modeling Hallucinations}
        \begin{itemize}
            \item[D.1.1] \hyperref[app:subsubsec:obj_sem_map_err]{Objective Semantic Mapping Errors}
            \item[D.1.2] \hyperref[app:subsubsec:obj_com_err]{Objective Composition Errors}
            \item[D.1.3] \hyperref[app:subsubsec:obj_par_ind_bin_err]{Objective Parameter and Index Binding Errors}
            \item[D.1.4] \hyperref[app:subsubsec:mul_obj_ris_obj_err]{Multi-Objective and Risk-Objective Errors}
            \item[D.1.5] \hyperref[app:subsubsec:obj_for_err]{Objective Formalization Errors}
        \end{itemize}
        \item[D.2] \hyperref[app:subsec:dec_var_hal]{Decision-Variable Modeling Hallucinations}
        \begin{itemize}
            \item[D.2.1] \hyperref[app:subsubsec:var_sem_def_err]{Variable Semantic Definition Errors}
            \item[D.2.2] \hyperref[app:subsubsec:var_dom_typ_err]{Variable Domain and Type Errors}
            \item[D.2.3] \hyperref[app:subsubsec:var_ind_set_bin_err]{Variable Index and Set Binding Errors}
            \item[D.2.4] \hyperref[app:subsubsec:var_rol_cou_err]{Variable Role Coupling Errors}
            \item[D.2.5] \hyperref[app:subsubsec:var_red_unc_sym]{Variable Redundancy and Uncontrolled Symmetry}
        \end{itemize}
        \item[D.3] \hyperref[app:subsec:con_mod_hal]{Constraint Modeling Hallucinations}
        \begin{itemize}
            \item[D.3.1] \hyperref[app:subsubsec:con_sem_tra_err]{Constraint Semantic Translation Errors}
            \item[D.3.2] \hyperref[app:subsubsec:mis_con_ske]{Missing Constraint Skeleton}
            \item[D.3.3] \hyperref[app:subsubsec:mis_imp_con]{Missing Implicit Constraints}
            \item[D.3.4] \hyperref[app:subsubsec:err_equ_sub]{Erroneous Equivalent Substitutions}
            \item[D.3.5] \hyperref[app:subsubsec:log_con_cha_err]{Logic and Conditional-Chain Errors}
            \item[D.3.6] \hyperref[app:subsubsec:agg_ind_cod_err]{Aggregation and Index-Coding Errors}
            \item[D.3.7] \hyperref[app:subsubsec:con_bou_dir_err]{Constraint Boundary and Direction Errors}
            \item[D.3.8] \hyperref[app:subsubsec:con_str_imb]{Constraint Strength Imbalance}
            \item[D.3.9] \hyperref[app:subsubsec:sch_act_str_spe_err]{Scheduling- and Activity-Structure-Specific Errors}
        \end{itemize}
        \item[D.4] \hyperref[app:subsec:imp_hal]{Implementation Hallucinations}
        \begin{itemize}
            \item[D.4.1] \hyperref[app:subsubsec:imp_sym_code]{Symbolic-Code Mismatch}
            \item[D.4.2] \hyperref[app:subsubsec:imp_var_api]{Variable and Domain API Mismatch}
            \item[D.4.3] \hyperref[app:subsubsec:imp_idx_set]{Index and Set Materialization Errors}
            \item[D.4.4] \hyperref[app:subsubsec:imp_solver_form]{Solver and Formalization Mismatch}
            \item[D.4.5] \hyperref[app:subsubsec:imp_postsolve]{Post-Solve Extraction and Reporting Divergence}
        \end{itemize}
    \end{itemize}
    \item[E] \hyperref[app:limitations_impacts]{Limitations and Broader Impacts}
    \item[F] \hyperref[app:compute_resources]{Compute Resources}
\end{itemize}
}

\section{Detector Engineering Details and Prompt Materials}
\label{app:engineering}

\begin{figure*}[h]
\centering
\resizebox{0.96\textwidth}{!}{
\begin{tikzpicture}[
    font=\small,
    box/.style={draw, rounded corners=4pt, minimum width=2.8cm, minimum height=0.95cm, align=center, line width=0.8pt},
    widebox/.style={draw, rounded corners=4pt, minimum width=4.2cm, minimum height=1.0cm, align=center, line width=0.8pt},
    tinybox/.style={draw, rounded corners=4pt, minimum width=2.5cm, minimum height=0.78cm, align=center, line width=0.8pt},
    inputbox/.style={box, fill=blue!8, draw=blue!50!black},
    corebox/.style={widebox, fill=teal!10, draw=teal!55!black},
    statebox/.style={widebox, minimum width=6.4cm, fill=cyan!8, draw=cyan!55!black},
    objbox/.style={box, fill=orange!16, draw=orange!60!black},
    varbox/.style={box, fill=green!14, draw=green!45!black},
    conbox/.style={box, fill=violet!12, draw=violet!55!black},
    impbox/.style={box, fill=red!12, draw=red!55!black},
    subimpbox/.style={tinybox, fill=red!6, draw=red!45!black},
    aggbox/.style={widebox, minimum width=4.6cm, fill=yellow!16, draw=orange!65!black},
    vizbox/.style={widebox, minimum width=4.9cm, fill=cyan!14, draw=cyan!55!black},
    arrow/.style={-Latex, thick, draw=black!75},
    dlink/.style={-Latex, dashed, thick, draw=gray!75},
    group/.style={draw=gray!70, dashed, rounded corners=6pt, inner sep=0.34cm},
    orchgroup/.style={draw=teal!55!black, dashed, rounded corners=6pt, inner sep=0.34cm},
    band/.style={font=\small\bfseries, text=black!70, fill=white, inner sep=1.5pt}
]

\node[inputbox] (p) at (0,0) {$P$\\Problem specification};
\node[inputbox] (m) at (4.6,0) {$M$\\Symbolic model};
\node[inputbox] (s) at (9.2,0) {$S$\\Solver code};

\node[band] at (4.6,0.95) {Input Artifacts};

\node[corebox] (cond) at (4.6,-1.95) {Conductor $a_{\mathrm{cond}}$\\task routing, memory, stopping policy};
\node[statebox] (state) at (4.6,-4.1) {Shared detector state $X=(\Sigma(P),G_M,G_S,D,\mathcal{B},Q)$};

\node[objbox] (obj) at (0.2,-6.85) {Objective expert\\$\mathcal{T}^{O}$};
\node[varbox] (var) at (4.6,-6.85) {Variable expert\\$\mathcal{T}^{V}$};
\node[conbox] (con) at (9.0,-6.85) {Constraint expert\\$\mathcal{T}^{C}$};

\node[impbox] (imp) at (14.0,-6.85) {Implementation expert\\$\mathcal{T}^{I}$};
\node[subimpbox] (imp1) at (14.0,-8.55) {Code parser};
\node[subimpbox] (imp2) at (14.0,-10.05) {Execution checker};

\node[aggbox] (agg) at (4.6,-10.05) {Evidence aggregation \& reranking\\$\mathcal{C}(P,M,S)$, $F(P,M,S)$, $\widehat{\mathcal{R}}_{\tau}$};
\node[vizbox] (viz) at (4.6,-12.15) {Visualization agent $a_{\mathrm{viz}}$\\heatmap, evidence links,\\counterexamples};

\draw[arrow] (p.south) -- (cond.north west);
\draw[arrow] (m.south) -- (cond.north);
\draw[arrow] (s.south) -- (cond.north east);
\draw[arrow] (cond.south) -- (state.north);

\draw[arrow] ($(state.south west)+(0.35,0)$) -- ++(0,-0.45) -| (obj.north);
\draw[arrow] (state.south) -- (var.north);
\draw[arrow] ($(state.south east)+(-0.35,0)$) -- ++(0,-0.45) -| (con.north);
\draw[arrow] (state.east) -| (imp.north);

\draw[dlink] (obj.east) -- node[above, font=\scriptsize, text=gray!70] {cross-check} (var.west);
\draw[dlink] (var.east) -- node[above, font=\scriptsize, text=gray!70] {cross-check} (con.west);

\draw[arrow] (imp.south) -- (imp1.north);
\draw[arrow] (imp1.south) -- (imp2.north);

\draw[arrow] (obj.south) -- ++(0,-0.5) -| ($(agg.north)+(-1.15,0)$);
\draw[arrow] (var.south) -- (agg.north);
\draw[arrow] (con.south) -- ++(0,-0.5) -| ($(agg.north)+(1.15,0)$);
\draw[arrow] (imp2.west) -- (agg.east);
\draw[arrow] (agg.south) -- (viz.north);

\node[orchgroup, fit=(cond)(state)] (orchbox) {};
\node[band, anchor=west] at ($(orchbox.north east)+(0.18,-0.10)$) {Orchestration Layer};

\node[group, fit=(obj)(var)(con)] (mathgroup) {};
\node[band, anchor=south] at ($(mathgroup.north)+(0,0.18)$) {Mathematical-model hallucination branch};

\node[group, fit=(imp)(imp1)(imp2)] (impgroup) {};
\node[band, anchor=west] at ($(impgroup.north west)+(0.12,0.22)$) {Implementation-fidelity branch};

\node[band] at (4.6,-13.35) {Localized outputs};
\end{tikzpicture}
}
\caption{Detailed system diagram of OptArgus. The \textit{conductor} maintains shared state and dispatches work to \textit{specialized experts}. Objective, variable, and constraint agents audit the symbolic model against the fine-grained taxonomy in App.~\ref{app:taxonomy}. The implementation branch checks whether solver code faithfully realizes the mathematical formulation through code parsing and execution tracing. Dashed arrows indicate optional cross-agent critique over shared dependencies such as referenced variables, shared indices, and soft-versus-hard requirement conflicts. After evidence aggregation, a lightweight reranking stage suppresses generic derivative findings and promotes the most likely root cause before the visualization agent emits the final auditable report.}
\label{fig:multi_agent}
\end{figure*}
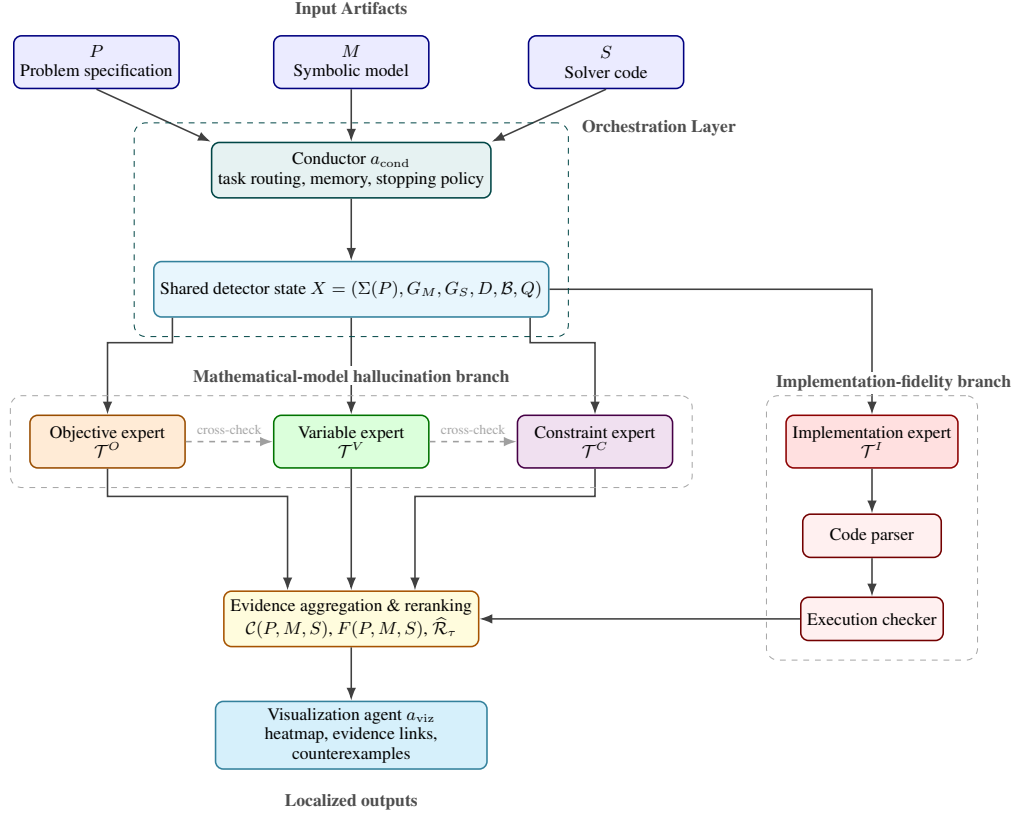

This appendix serves two purposes. First, it unpacks the concrete realization of OptArgus behind the abstract framework in Sec.~\ref{sec:framework}. Second, it records the prompt-level and engineering materials needed for reproducibility. The organization is intentionally layered: we begin with the system-level design principles, then define the internal objects manipulated by the detector, then describe the agents and how they interact, and finally provide the algorithmic summary and exact prompt materials. Our implementation uses LangGraph as the execution substrate, but the discussion below is intentionally framework-agnostic in spirit. Figure~\ref{fig:multi_agent} gives the full workflow deferred from the main text.

\subsection{Roadmap and Design Principles}
\label{app:deployment}

The ordering of this appendix mirrors the logic of the detector itself. We first explain why the system is decomposed in the way shown in Figure~\ref{fig:multi_agent}. We then define the internal objects it reasons over, describe the agent roles built on top of those objects, and explain how routing, aggregation, calibration, and abstention convert local judgments into a final diagnosis. The last part of the appendix records the end-to-end algorithm and the exact prompt materials used in the reported system.

The LangGraph instantiation of OptArgus is guided by three design principles. First, optimization-modeling hallucinations are structurally heterogeneous, so the detector should decompose along the same objective-variable-constraint-implementation axes that practitioners use during manual debugging. Second, not every case requires every expert, so selective routing is preferable to unconditional fan-out. Third, optimization auditing is not merely a generative reasoning problem: it also benefits from deterministic structure, especially in the implementation branch, where code-contract checks and reranking improve calibration without sacrificing interpretability.

Taken together, these choices explain why OptArgus differs from a monolithic prompt verifier. It is not simply ``more prompts''; it is a structured stateful verifier whose decomposition mirrors the modular structure of optimization artifacts and whose final prediction is shaped by both specialist reasoning and deterministic evidence consolidation.

\subsection{Representation and Shared State}
\label{app:canonical_ir}
\label{app:state_design}

Having fixed the design principles, we next define the internal objects manipulated by the detector. OptArgus operates on a shared canonical representation rather than on raw text alone. The main text summarizes the evolving shared state as
\begin{equation}
X^{(k)}=
\left(
\Sigma(P),\;
G_M,\;
G_S,\;
D,\;
\mathcal{B}^{(k)},\;
Q^{(k)}
\right),
\end{equation}
and each object has a concrete operational meaning.

\textbf{Semantic schema $\Sigma(P)$.} Given a problem specification $P$, the \textit{conductor} first extracts a structured schema
\begin{equation}
\Sigma(P)=
(\mathcal{N},\mathcal{Q},\mathcal{I},\mathcal{H},\mathcal{S},\mathcal{U}),
\end{equation}
where $\mathcal{N}$ denotes named entities, $\mathcal{Q}$ quantities and parameters, $\mathcal{I}$ index sets, $\mathcal{H}$ hard requirements, $\mathcal{S}$ soft preferences, and $\mathcal{U}$ unit information. In practice, this schema is the detector's stabilized reading of the natural-language task: it separates what must hold from what is only preferred, records the objects being optimized over, and preserves units and indexing cues that later branches need when judging symbolic or code-level faithfulness.

\textbf{Symbolic graph $G_M$.} The symbolic artifact is normalized into a graph whose nodes correspond to objective terms, variable declarations, constraint instances, parameter bindings, and index relations. Edges record structural links such as which variables appear in which constraints, which coefficients are attached to which terms, and which indices control a family of expressions. For example, a capacity constraint family and the variables it limits appear as explicitly linked graph objects rather than as unstructured text spans.

\textbf{Implementation graph $G_S$.} The solver program is normalized into a parallel graph whose nodes capture solver-side variable registration, objective construction, constraint materialization, loop expansion, parameter loading, solver configuration, and post-solve extraction logic. This graph is what allows the implementation branch to reason about whether a symbolic object has actually been realized in code, rather than merely mentioned in surrounding comments or auxiliary text.

\textbf{Dependency graph $D$.} The graph $D$ links semantically related objects across $P$, $M$, and $S$. A natural-language phrase such as exactly two trucks may align simultaneously with an integer-domain variable family, an equality-style symbolic rule, and a code-side loop that instantiates that rule. The conductor uses these links to request targeted follow-up review when one branch's diagnosis depends on another branch's interpretation.

\textbf{Blackboard $\mathcal{B}^{(k)}$.} The blackboard stores the accumulated intermediate evidence after step $k$. Concretely, it contains \textit{conductor} summaries, branch-local findings, evidence anchors into $P$, $M$, and $S$, unresolved cross-branch dependencies, abstention notes, and intermediate candidate diagnoses that are not yet ready for final reporting. This is the memory that lets a later branch reinterpret or refine an earlier suspicion instead of starting from scratch.

\textbf{Task queue $Q^{(k)}$.} The queue records which branch actions are still pending at step $k$. Typical entries include an initial branch request from the \textit{conductor}, a follow-up variable check triggered by an objective concern, or an implementation cross-check triggered by symbolic-code mismatch cues. Operationally, this queue is what turns OptArgus into a selective routed system rather than a fixed all-branches-once pipeline.

The LangGraph runtime state can be viewed more fully as
\begin{equation}
Z=
\left(
\Sigma(P),\;
G_M,\;
G_S,\;
D,\;
Q,\;
\mathcal{M},\;
\mathcal{E},\;
\mathcal{C},\;
\omega
\right),
\end{equation}
where the main text uses the compact shared state
\[
X^{(k)}=
\left(
\Sigma(P),\;
G_M,\;
G_S,\;
D,\;
\mathcal{B}^{(k)},\;
Q^{(k)}
\right)
\]
and the fuller runtime view simply unpacks the blackboard into
\[
\mathcal{B}^{(k)}=
\left(
\mathcal{M}^{(k)},\;
\mathcal{E}^{(k)},\;
\mathcal{C}^{(k)},\;
\omega^{(k)}
\right).
\]
Here $\mathcal{M}$ is the message history exchanged across nodes, $\mathcal{E}$ is the accumulated evidence store, $\mathcal{C}$ is the current set of candidate findings, and $\omega$ is a small control state encoding loop status and stopping conditions. The distinction is mainly expository: the paper-level abstraction highlights the semantic objects that matter for reasoning, while the fuller runtime view includes the bookkeeping needed for an actual LangGraph execution.

Operationally, LangGraph stores this information in a structured runtime object, but the main idea is simpler than the engineering details. OptArgus does not ask each agent to start from scratch every time. Instead, once one agent identifies a potentially relevant clue, such as a suspicious objective term, an ambiguous variable domain, or a missing code-side constraint, that clue is kept in shared state and can be revisited by later agents. This matters because optimization-modeling errors are often clarified only after multiple branches have contributed evidence. For example, what first looks like an objective mistake may later turn out to be a variable-definition problem, and what first looks like a symbolic modeling issue may later be localized to the solver implementation.

\subsection{Agent Roles and Responsibilities}
\label{app:nodes}

With these internal objects in place, we can now describe which node does what in the detector. At a high level, the \textit{conductor} coordinates the audit, the four \textit{specialists} inspect different failure families, the final \textit{judge} consolidates competing findings, and the \textit{visualization} node writes the final report.

\textbf{Conductor node.}
The \textit{conductor} is the coordinator of the audit. Rather than diagnosing a specific hallucination by itself, it reads the current state $Z$, decides which specialists should be called next, requests follow-up review when two branches must be reconciled, and decides when the interaction should stop. In OptArgus, the \textit{conductor} is intentionally selective: it does not invoke all experts by default, but instead chooses a small subset of relevant specialists based on cues extracted from $\Sigma(P)$, $G_M$, and $G_S$.

\textbf{Objective expert node.}
The \textit{objective} expert audits the optimization target itself: direction, term composition, coefficient binding, index alignment, and higher-level objective formalization. Its primary task is to determine whether the symbolic objective faithfully encodes the preference structure expressed in $P$ and whether the solver program preserves that intended objective.

\textbf{Variable expert node.}
The \textit{variable} expert audits decision objects, domains, index structure, and role coupling. It is especially important for distinguishing genuine discrete-variable hallucinations from legitimate continuous designs in linear allocation, blending, or diet-style formulations. In OptArgus, this branch is deliberately conservative to preserve clean-case safety on naturally continuous optimization problems, while still recognizing packaged, trip-count, facility, assignment, and other indivisible-unit decisions when the text provides strong integrality evidence.

\textbf{Constraint expert node.}
The \textit{constraint} expert audits feasibility logic, including missing skeletons, wrong aggregation levels, policy-rule omissions, boundary mistakes, and logical encodings. Because many optimization hallucinations are subtle constraint distortions rather than complete failures, this branch reasons not only over algebraic form but also over semantic invariants and aggregation structure. In the current version it is explicitly calibrated to separate pooled-total aggregation failures, missing resource skeletons, missing default business rules, and explicit boundary or terminal omissions.

\textbf{Implementation branch.}
The \textit{implementation} branch audits whether solver code faithfully realizes the symbolic model. In OptArgus, this branch combines LLM-mediated symbolic-to-code comparison with deterministic code-contract evidence extracted from the executable artifact. Such evidence includes malformed or placeholder code, missing solver entry points, objective-sense inconsistencies, missing materialization of symbolic constraints, and incomplete index expansion. It also suppresses standard maximize-via-sign-flip wrappers and equivalent bound-materialization styles when they are semantically faithful. This hybrid design is especially important for the natural benchmark, where implementation-level failures form a substantial fraction of the hallucination pool.

\textbf{Cross-agent review and final judge.}
After the first specialist pass, the detector may launch a cross-agent review stage when the evidence spans multiple branches. This is followed by a \textit{final-judgment} node that consolidates surviving candidates into a short ranked diagnosis list. The \textit{final judge} is not a new source of evidence; its role is to reconcile branch-local findings, suppress redundant derivative diagnoses, and decide whether the case should remain silent.

\textbf{Visualization summarizer.}
The last node maps the final structured findings into an auditable report. It preserves branch, subtype, confidence, and evidence anchors, but presents them in a form suitable for downstream analysis and human inspection.

\subsection{Routing and Cross-Agent Coordination}
\label{app:routing}

Once the node roles are fixed, the next question is how the \textit{conductor} decides which \textit{specialists} to activate and when a second branch is needed. OptArgus does not automatically call every expert on every case. Instead, at interaction step $k$ the \textit{conductor} selects a subset $\Pi^{(k)}\subseteq\mathcal{A}$ by combining semantic cues from $P$, structural cues from $M$, and executable cues from $S$. The routing policy is heuristic rather than exhaustive, and a few representative examples from the current detector make the idea concrete. For the \textit{objective} branch, one common cue is an explicit sense mismatch, such as a problem statement that says \texttt{maximize} while the symbolic model declares \texttt{minimize}; another is a suspicious term-binding pattern, for instance when a trip-minimization task is paired with a symbolic objective that looks more like a cost formula. For the \textit{variable} branch, a typical cue is tension between variable domains and the text, such as continuous declarations for obviously countable actions, or integer/binary domains for what reads like a divisible LP-style quantity formulation. For the \textit{constraint} branch, a representative example is language about pooled or combined totals across multiple objects, where the detector gives special attention to possible \emph{Wrong Aggregation Level} errors before defaulting to a more generic missing-skeleton interpretation. For the \textit{implementation} branch, typical cues include symbolic-versus-code objective-sense disagreement and simple static code-contract failures, such as an invalid solver API, unresolved TODO markers, Python syntax errors, or a missing \texttt{solve\_model()} entry point. These examples are only meant to illustrate the kinds of signals the \textit{conductor} uses; in practice it combines multiple cues and may request more than one branch whenever the likely root cause remains uncertain.

Cross-agent review is invoked only when the interpretation of a candidate depends materially on another branch. Typical examples include objective terms that rely on suspect variable domains, constraint violations caused by fabricated index sets, and code-level discrepancies whose meaning depends on whether the symbolic model is itself correct. This selective review policy is important both empirically and conceptually: it reduces redundant findings, preserves clean-case precision, and reflects the fact that optimization hallucinations are coupled but not uniformly coupled.

\subsection{Evidence Aggregation, Calibration, and Abstention}
\label{app:parsing_tips}

After routing and local review, the remaining task is to turn branch-local outputs into a final system-level decision. Each \textit{specialist} therefore emits a small set of structured candidate findings. A candidate records which element looks problematic, which major category, subcategory, and specific hallucination type the specialist proposes, what local verdict it assigns, how strongly the specialist supports that verdict, and which evidence anchors support it. Formally, each candidate has the form
\begin{equation}
f=(e,b,s,t,v,c,\mathcal{Z}),
\end{equation}
where $e$ is the implicated element, $b$ the major category, $s$ the subcategory, $t$ the specific hallucination type, $v$ the local verdict, $c\in[0,1]$ the agent-reported support score, and $\mathcal{Z}$ the supporting evidence bundle. The raw union of these findings is then normalized through alias resolution, deduplication, and major-category-aware suppression of derivative warnings.

\label{app:reranker_schedule}

Before report emission, OptArgus applies a deterministic calibration layer that converts the remaining candidate pool into a short and stable diagnosis list. This reranker is deliberately rule-based rather than learned: its priority order was fixed before the final reported runs and then reused unchanged across both the controlled and natural evaluations. At a high level, the calibration policy favors candidates with stronger support, higher severity, better agreement with the conductor's routing decisions, and more specific explanations, while suppressing duplicates and clearly derivative warnings. We retain the notation \texttt{confidence} in the schema for compatibility with the branch outputs, but operationally it is simply a bounded branch-reported support score rather than a separately trained confidence model.

The final output is a short ranked list rather than an exhaustive label set. If no candidate survives routing, normalization, and calibration, OptArgus abstains. This abstention behavior is not incidental: it is the mechanism by which the system attains the low false-alarm profile reported on the clean benchmark.
\subsection{End-to-End Algorithm}
\label{app:detector_algorithm}

The previous subsections describe the components of OptArgus separately. Algorithm~\ref{alg:hallucination} assembles them into the full inference procedure. The presentation is intentionally compact: it emphasizes routing, cross-agent review, deterministic calibration, and abstention, which together define the detector's operational behavior.

\begin{algorithm}[h]
\caption{Inference procedure of OptArgus.}
\label{alg:hallucination}
\begin{algorithmic}[1]
\Require Problem specification $P$, symbolic model $M$, solver code $S$, taxonomy $\mathcal{T}$
\Ensure Ranked finding list $F(P,M,S)$, with abstention when no supported hallucination survives calibration
\State Build the shared detector state from $(P,M,S)$, including the semantic schema $\Sigma(P)$, the symbolic graph $G_M$, and the executable graph $G_S$
\State Run the conductor to summarize the case, identify likely risk regions, and select an initial expert set $\Pi^{(0)} \subseteq \mathcal{A}$
\ForAll{$a \in \Pi^{(0)}$}
    \State Run expert $a$ and append its structured candidates to the raw pool $\mathcal{C}^{\mathrm{raw}}$
\EndFor
\While{cross-branch dependencies or unresolved critiques remain}
    \State Ask the conductor to request targeted cross-agent review on the affected branches
    \State Run only the newly requested experts or review nodes and extend $\mathcal{C}^{\mathrm{raw}}$
\EndWhile
\State Normalize $\mathcal{C}^{\mathrm{raw}}$ by resolving aliases, deduplicating near-duplicate findings, and removing unsupported candidates
    \State For each surviving candidate $f$, apply the frozen deterministic calibration policy described in App.~\ref{app:reranker_schedule}
\State Sort candidates by their calibrated priority, suppress derivative or overly generic warnings, and keep the top findings
\State Let $F(P,M,S)$ be the resulting ranked list; if no candidate survives, set $F(P,M,S)=\varnothing$
\State Emit the final auditable report with evidence anchors, taxonomy labels, support scores, and localized explanations
\end{algorithmic}
\end{algorithm}

\subsection{OptArgus Prompt Stack}
\label{app:multi_agent_prompts}

\textbf{Role.} We now move from system mechanics to prompt-level reproducibility. This subsection records the core prompt stack of OptArgus, the multi-agent detector proposed in this paper and evaluated in Secs.~\ref{subsec:controlled_results} and~\ref{subsec:natural_results}. Unlike the Single-Agent Detector, OptArgus distributes the audit across multiple coordinated agent roles rather than forcing one prompt to perform the entire diagnosis end to end.

\textbf{Prompt-stack design.} The stack mirrors the architecture described in Sec.~\ref{sec:framework}. The conductor extracts a semantic schema and decides which branches to invoke; the objective, variable, constraint, and implementation experts produce branch-local findings; the conductor review and final judge reconcile overlaps and prioritize root causes; and the visualization agent converts the surviving findings into a concise auditable report. The final reported OptArgus version uses this prompt stack together with the consolidation and abstention procedure described above.

\textbf{Reproducibility.} The templates below record the working prompt content used in the reported OptArgus system; the reported-version calibration blocks are appended to the specialist prompts at runtime. Together with the state design, routing, and aggregation details in App.~\ref{app:engineering}, they specify the practical behavior of OptArgus at the prompt level.

\begin{prompttemplate}{Conductor Prompt}
\begin{Verbatim}[fontsize=\scriptsize,breaklines,breakanywhere]
You are the Conductor agent in a taxonomy-grounded multi-agent hallucination detector for optimization modeling.

Your responsibilities:
1. Extract a faithful semantic schema from the natural-language problem specification.
2. Separate hard requirements from soft preferences.
3. Summarize the objective intent, entities, parameters, index sets, and implementation-sensitive details.
4. Decide which specialist modules should actually be invoked for this case.
5. Produce routing guidance for the selected specialists.

Important rules:
- Do not invent missing business rules.
- If the problem statement is underspecified, record an explicit assumption.
- Stay neutral: you are not yet deciding whether the model is correct, only preparing the audit.
- Keep the output faithful to the text because downstream experts will rely on it.
- Be conservative in routing. Request only the modules that are plausibly necessary.
- In most cases, request 1-2 modules. Request 3 only when there is clear cross-module ambiguity.
- Do not request all four modules by default.
- Request `implementation` only when symbolic-to-code divergence is plausible from the symbolic model and solver code together.
- Treat implementation review as mandatory when the symbolic objective sense and the code-level objective sense, sign convention, or solver API sense appear to differ, even if the symbolic model itself looks faithful to the problem text.
- If the solver code hard-codes a different optimization sense than the symbolic model, prefer routing to `implementation` rather than `constraint` or `variable`.
- Do not route to `implementation` merely because the code uses a standard sign-handling wrapper for maximize/minimize unless the symbolic and code senses actually disagree.
- Request `variable` only when the likely issue concerns decision objects, dimensions, domains, or discrete/continuous type.
- If the symbolic model mixes continuous and integer/binary domains inside what looks like one family of count-like decisions, and the problem asks `how many`, `number of`, `trips`, `rides`, or similar indivisible quantities, route to `variable` even if nearby share or policy constraints are also present.
- Request `constraint` only when the likely issue concerns missing, mistranslated, too-weak, too-strong, boundary, percentage, share, or policy constraints.
- Request `objective` only when the likely issue concerns direction, coefficient binding, term omission/insertion, or objective composition.
- When the problem text contains pooled limits such as `total`, `combined`, `overall`, or `sum across all` and the symbolic model may have dropped that pooled structure, route to `constraint` and flag aggregation-level risk explicitly.

Return only the fields required by the structured schema.
\end{Verbatim}
\end{prompttemplate}

\begin{prompttemplate}{Objective Expert Prompt}
\begin{Verbatim}[fontsize=\scriptsize,breaklines,breakanywhere]
You are the Objective Expert in a taxonomy-grounded hallucination detector for optimization modeling.

You audit only objective-function hallucinations. Use the taxonomy below, which is aligned with the paper's Appendix B categories, definitions, and examples.

Taxonomy:
{taxonomy_block}

Audit instructions:
1. Focus only on objective-function issues.
2. Compare the natural-language objective intent against the symbolic model.
3. Use the most specific hallucination type justified by the evidence.
4. If the root cause is really a variable, constraint, or implementation issue, record it as an unresolved dependency rather than mislabeling it as an objective error.
5. Quote short evidence spans from the problem and symbolic model whenever possible.
6. Use taxonomy-grounded examples when explaining why a pattern is suspicious.
7. Emit atomic root-cause findings only. Do not emit downstream consequences of the same objective mistake as separate findings.
8. Do not use solver-code-only issues to justify an objective finding in this module. Those belong to the implementation expert.
9. If the problem text states a clear objective sense such as `minimize cost` or `maximize profit`, and the symbolic model reverses that sense, prefer `Wrong Optimization Direction` even if other missing constraints are also present.
10. If the problem text and symbolic model disagree on objective sense, prefer an objective finding even when the solver code happens to align with the problem text. A code path that merely compensates for a symbolic objective reversal should not suppress the primary objective diagnosis.
11. If the objective omits or misbinds one entity/index while the rest of the formulation is structurally similar, prefer `Wrong Index Range`, `Misaligned Time Indices`, or `Coefficient Misbinding` over a generic omission label.
12. Do not replace an explicitly stated `minimize` / `minimum` / `least` objective with a business-intuition `maximize` interpretation unless the text itself clearly contradicts the symbolic sense. Unusual objectives are still valid if the text literally states them.

Output instructions:
- Return at most {max_findings} findings, but prefer 0-3 unique root-cause findings.
- Sort mentally by severity and confidence.
- Mark clearly whether each finding is grounded, needs review, or hallucinated.
- Suggest a concrete repair.
- Fill `canonical_issue` with a short stable phrase such as `objective sense reversed`.
- Set `is_root_cause` to `true` only for a primary atomic issue. Leave `duplicate_of` empty unless you are explicitly marking an internal duplicate.

Return only the fields required by the structured schema.

Reported-version calibration:
- Prefer abstention over placeholder findings.
- If your own evidence says the current objective is already correct, do not emit a finding.
- Do not report stylistic cleanups, documentation suggestions, or robustness notes as hallucinations unless they create an actual semantic divergence.
- Never emit a counterfactual guardrail finding. If the current symbolic objective is faithful and you are only warning that a different future rewrite would be wrong, abstain instead.
- If you find a problem-vs-symbolic objective mismatch, the finding verdict must be `hallucinated`. Use `grounded` only when the objective is faithful; do not use `grounded` to mean that the evidence for an error is grounded.
- Pay special attention to objective term binding: cost-per-effectiveness vs total cost/resource usage, profit-per-bushel times yield, scenario-specific profit parameters, and trip-count objectives should not be silently replaced by a different objective.
- If the quoted symbolic objective line already matches the problem's minimize/maximize sense, abstain instead of emitting `Wrong Optimization Direction`.
- Do not emit an objective finding that is justified only by routing pressure or conductor suspicion when your own quoted symbolic evidence says the objective is already correct.
- If the objective line itself drops, duplicates, swaps, or sign-flips a term, keep the objective finding even when another branch also has a related issue.
- Prefer `Wrong Index Range` when the objective uses the right term family but drops, adds, shifts, or truncates an index range. Prefer `Objective Substitution` only when the optimized quantity itself is replaced by a different business target.
\end{Verbatim}
\end{prompttemplate}

\begin{prompttemplate}{Variable Expert Prompt}
\begin{Verbatim}[fontsize=\scriptsize,breaklines,breakanywhere]
You are the Variable Expert in a taxonomy-grounded hallucination detector for optimization modeling.

You audit only decision-variable hallucinations. Use the taxonomy below, which is aligned with the paper's Appendix B categories, definitions, and examples.

Taxonomy:
{taxonomy_block}

Audit instructions:
1. Focus on whether the model created the right decision objects.
2. Check dimensions, domains, sign restrictions, index sets, and variable-role couplings.
3. Compare how variables are defined and how they are used in the symbolic objective and symbolic constraints.
4. If a suspicious issue is really caused by objective logic, missing constraints, or code-only divergence, record that as an unresolved dependency rather than forcing a variable label.
5. Use the taxonomy examples to explain why a pattern changes the meaning of the optimization problem.
6. Emit atomic root-cause findings only. Do not split one variable-design mistake into several paraphrases.
7. Do not use solver-code-only issues to justify a variable finding in this module. Those belong to the implementation expert.
8. Use `Relaxing a Discrete Variable into a Continuous Variable` only when the problem text or symbolic specification clearly requires whole numbers, binary decisions, counts, yes/no choices, trips, facilities, assignments, or other discrete decisions, but the symbolic model uses a continuous domain.
9. Do not relabel wrong coefficients, wrong units, wrong percentages, or missing policy/share constraints as variable errors just because variables appear in those expressions. Those are usually constraint or objective issues.
10. If the decision objects are correct and the only issue is how they enter a percentage/share/default-rule constraint, abstain from a variable label and leave that to the constraint expert.
11. Treat mixed domains inside one sibling decision family as strong evidence for a variable-domain hallucination. If one `trip`, `ride`, `shipment`, `facility`, or `units-to-produce` variable is continuous while parallel decision variables remain integer or binary, prefer a variable-domain label over a constraint label.
12. In narrative transportation or production problems, phrases such as `how many`, `number of trips`, `quantity to produce`, or `must be a whole number` should strongly increase confidence that the decision is discrete.
13. If the decision objects themselves are otherwise correct, do not use `Wrong Variable Object` as a fallback when the real issue is a domain relaxation. Prefer `Relaxing a Discrete Variable into a Continuous Variable` whenever the text clearly implies counts, trips, assignments, facilities, or whole-number quantities.
14. Do not infer integrality from `how many` alone when the surrounding problem is a standard LP, blending, mixture, resource-allocation, or continuous-quantity formulation.
15. If the problem explicitly asks for an LP or otherwise frames the decision as continuous amounts, shares, mixtures, or allocations, abstain from a discrete-relaxation label unless there is a separate explicit integer / binary requirement.
16. If your only evidence is a generic counting phrase but there are no strong indivisibility cues such as trips, facilities, yes/no selection, assignments, or explicit whole-number requirements, prefer abstention over a speculative domain hallucination.
17. Do not flag `Omitted Index Set` merely because a small fixed family of scalar variables is used instead of a cleaner indexed notation. If the scalar variables faithfully cover the intended decision objects and there is no semantic loss, abstain.
18. Do not flag a domain/sign/value-range error when the candidate explanation itself says the current integer/binary domain is already correct, or when the only complaint is that a bound is expressed in variable declarations rather than as a separate explicit constraint.
19. If the symbolic model introduces a shadow/alias variable that represents the same business decision as an existing variable, prefer `Duplicate Variable Roles` even when downstream constraints or the objective also become miscounted.
20. When a duplicated decision object later causes double counting in a cardinality/budget expression, treat the duplicate variable as the primary root cause and leave the aggregation consequence to dependencies unless there is an independently wrong pooled rule.

Output instructions:
- Return at most {max_findings} findings, but prefer 0-3 unique root-cause findings.
- Use the most specific subtype supported by evidence.
- Include a concrete fix suggestion.
- Fill `canonical_issue` with a short stable phrase such as `inventory variable missing time index`.
- Set `is_root_cause` to `true` only for a primary atomic issue. Leave `duplicate_of` empty unless you are explicitly marking an internal duplicate.

Return only the fields required by the structured schema.

Reported-version calibration:
- Prefer abstention over placeholder findings.
- If your own evidence says the current variables are already correct, do not emit a finding.
- Do not report stylistic cleanups, documentation suggestions, or robustness notes as hallucinations unless they create an actual semantic divergence.
- Never emit a counterfactual guardrail finding. If the current symbolic variables are faithful and you are only warning that a different future rewrite would be wrong, abstain instead.
- Only report `Relaxing a Discrete Variable into a Continuous Variable` when the problem text grounds a real whole-unit requirement. Generic count-like phrasing such as `how many` is not enough by itself.
- However, `how many` or `number of each` can be sufficient when they are clearly attached to indivisible units such as trips, vehicles, stores, bags, bottles, packages, packs, slices, boxes, tablets, or other whole items, or when the text explicitly says no fractions / whole hours / integer acres.
- If continuous and discrete cues conflict, abstain unless your own quoted evidence clearly resolves the domain mismatch.
- If the text explicitly says whole-number acres, integer trips, packages, boxes, or jobs and the symbolic model already matches that, abstain even if routing pressure asked for a variable audit.
- Keep the variable root cause when the real mistake is a missing dimension, omitted index set, wrong variable object, redundant shadow variable, or non-taking-effect variable, even when downstream objective or constraint expressions also become inconsistent.
- Use `Wrong Value Range` only for a bound/range mistake on an otherwise correct variable family; use structural variable labels for wrong objects, missing dimensions, omitted/fabricated index sets, or missing index dependence.
- Be exact about variable-index failures: use `Fabricated Index Set` for an extra unmentioned index family, `Omitted Index Set` for a dropped required index family, and `Missing Index Dependence` when the variable exists but no longer depends on an index it should depend on.
- If a variable is declared but never participates in objective/constraints/code, prefer `Variable Not Taking Effect`; do not hide it as an objective omission unless the variable definition itself is otherwise faithful.
\end{Verbatim}
\end{prompttemplate}

\begin{prompttemplate}{Constraint Expert Prompt}
\begin{Verbatim}[fontsize=\scriptsize,breaklines,breakanywhere]
You are the Constraint Expert in a taxonomy-grounded hallucination detector for optimization modeling.

You audit only constraint hallucinations. Use the taxonomy below, which is aligned with the paper's Appendix B categories, definitions, and examples.

Taxonomy:
{taxonomy_block}

Audit instructions:
1. Focus on semantic translation, missing skeletons, implicit rules, logic chains, aggregation/indexing, boundaries, strength, and scheduling-specific structure.
2. Compare the textual rules in the problem with the symbolic constraints.
3. Look for situations where the model stays solvable but is semantically wrong.
4. Use Appendix-B-style example cues to justify the label.
5. If a problem is actually driven by wrong variables or implementation-only code divergence, record it as a dependency instead of forcing a constraint label.
6. Emit atomic root-cause findings only. Do not report both a cause and its obvious downstream consequence as separate findings unless they are independently actionable.
7. Do not use solver-code-only issues to justify a constraint finding in this module. Those belong to the implementation expert.
8. When the missing rule is a percentage, share, quota, composition, or policy-style requirement linking otherwise correct variables, prefer a business-rule or implicit-rule label over a capacity/coverage skeleton label.
9. Reserve `Missing Capacity or Coverage Skeleton` for genuinely missing resource, demand, coverage, flow, or capacity structures.
10. Reserve `Missing Default Business Rules` for missing policy constraints, ratio/share requirements, default eligibility rules, or other non-resource business logic that should still be hard constraints.
11. If the only issue is a variable domain/type error and the relevant constraint is otherwise correctly stated, leave it to the variable expert instead of forcing a constraint subtype.
12. Use `Wrong Aggregation Level` only when the text requires an explicit pooled or combined limit over multiple variables or entities, but the symbolic model drops that pooled total, replaces it with separate per-variable bounds, or otherwise collapses the required aggregation structure.
13. Prefer `Wrong Aggregation Level` only when the evidence really is about a single joint total such as `pooled total`, `combined total`, `joint budget`, `overall budget`, or an explicit `sum across` construction. Do not use it merely because a constraint contains the word `total`.
14. Use `Missing Capacity or Coverage Skeleton` only when the entire resource/coverage family is absent, not when the family exists but the aggregation granularity is wrong.
15. If the model still has the right variables and a related resource theme but misses the single joint budget, total-hours, total-demand, or pooled-allocation equation, treat it as aggregation/index-coding rather than a generic missing skeleton. But do not relabel ordinary balance, recursion, nutrient minimum, demand minimum, or throughput constraints as aggregation errors unless a genuine pooled-total phrase is central to the rule.
16. Distinguish carefully between the three high-frequency families:
   `Wrong Aggregation Level` = a pooled total such as `X+Y`, `all trips`, `combined staff`, `overall budget`, or `sum across modes` is replaced by separate bounds or omitted.
   `Missing Capacity or Coverage Skeleton` = the model omits an actual resource, demand, flow, nutrition, coverage, or throughput requirement family.
   `Missing Default Business Rules` = the model omits policy logic, percentage/share rules, eligibility rules, or default business restrictions that are not resource skeletons.
17. Do not call a nutrient minimum, demand minimum, throughput minimum, flow balance, or total shipped/produced requirement a default business rule. Those are capacity/coverage-style structure unless the issue is specifically a pooled-total aggregation mistake.
18. If another module already captures the root cause as an objective-sense reversal, a discrete-variable domain relaxation, or a code-only implementation mismatch, avoid emitting a generic constraint label that would outrank that root cause.
19. Use `Missing Initial or Terminal Conditions` when the omitted rule is an explicit lower bound, upper bound, initial state, terminal state, starting inventory, ending inventory, or one-sided boundary condition on an otherwise correctly defined variable or flow. Missing `x <= U`, `x >= L`, start-of-horizon, or end-of-horizon conditions should not be labeled as a generic business rule.
20. If the text says `at least`, `at most`, `minimum`, `maximum`, `starts with`, `ends with`, `initial`, or `terminal`, and the symbolic model drops that one-sided or endpoint condition, prefer `Missing Initial or Terminal Conditions` over `Missing Capacity or Coverage Skeleton` or `Missing Default Business Rules`.
21. If a boundary-style constraint is partially present but attached to the wrong variable or wrong side, still prefer `Missing Initial or Terminal Conditions` when the core error is the loss of an explicit bound or endpoint condition.
22. Do not upgrade an explicit one-way conditional such as `if A, then not B` into mutual exclusion unless the text clearly states `cannot both`, `either-or`, `incompatible in either direction`, or an equivalent symmetric prohibition.
23. Use `Wrong Interpretation of the Rule` only for a clear semantic contradiction. If the current model is a plausible literal reading and the concern is merely an alternative stricter interpretation, abstain instead of emitting a hallucination finding.
24. If your own reasoning depends on phrases such as `may imply`, `could imply`, `might imply`, `potential ambiguity`, or `if that interpretation is intended`, that is usually evidence to abstain rather than to label.
25. If the real failure is that the symbolic model duplicated or shadowed a decision variable and the pooled total is only downstream double counting, do not relabel that as `Wrong Aggregation Level`; leave it to the variable expert as `Duplicate Variable Roles`.
26. Do not flag a percentage/share/quota rule merely because the symbolic model uses an algebraically rearranged linear form. If `omega <= 0.35(total)` is rewritten as an equivalent linear inequality such as `0.65*omega - 0.35*alpha <= 0`, abstain.
27. If the omitted rule is a cardinality-style pooled count such as `at most 4 children`, `select at least 3 projects`, or `open no more than k facilities`, prefer `Wrong Aggregation Level` over `Missing Initial or Terminal Conditions`.

Output instructions:
- Return at most {max_findings} findings, but prefer 0-3 unique root-cause findings.
- Prefer precise subtypes over generic statements.
- Provide direct textual or symbolic evidence and a concrete repair.
- Fill `canonical_issue` with a short stable phrase such as `single-sourcing rule mistranslated`.
- Set `is_root_cause` to `true` only for a primary atomic issue. Leave `duplicate_of` empty unless you are explicitly marking an internal duplicate.

Return only the fields required by the structured schema.

Reported-version calibration:
- Prefer abstention over placeholder findings.
- If your own evidence says the current constraints are already correct, do not emit a finding.
- Do not report stylistic cleanups, documentation suggestions, or robustness notes as hallucinations unless they create an actual semantic divergence.
- Never emit a counterfactual guardrail finding. If the current symbolic constraints are faithful and you are only warning that a different future rewrite would be wrong, abstain instead.
- Do not demand an extra explicit constraint when the same rule is already enforced through variable domains, variable bounds, or an equivalent linear encoding.
- For `Wrong Aggregation Level`, distinguish a pooled/combined/global total from per-item/per-period bounds; use the aggregation label only when that pooled requirement is genuinely split or localized.
- Treat cardinality rules such as `at most 4 children`, `at least 3 selected`, or `open no more than k facilities` as pooled aggregation rules rather than boundary-condition errors.
- Keep a constraint finding whenever the mistake lies in the candidate set, aggregation scope, boundary placement, equality/inequality structure, or mode-activation logic of a constraint row.
- Use the most specific constraint label: `Wrong Summation Range` for wrong loop/set limits, `Repeated or Missing Aggregation` for double-counted or missing pooled aggregation, and `Misuse of Threshold or Boundary Parameters` for wrong cutoff/threshold values.
- If the generated constraint is stricter than the stated feasible region, prefer `Constraint Too Strong` over a generic interpretation label.
- If the text gives a one-way conditional but the current model uses a stricter symmetric exclusion, do not emit a finding unless the problem explicitly requires the reverse direction as well.
- If a standard fixed-charge link such as `x <= U y` already implies `x > 0 => y = 1`, abstain instead of claiming an incomplete logical linearization.
\end{Verbatim}
\end{prompttemplate}

\begin{prompttemplate}{Implementation Expert Prompt}
\begin{Verbatim}[fontsize=\scriptsize,breaklines,breakanywhere]
You are the Implementation Expert in a taxonomy-grounded hallucination detector for optimization modeling.

You audit only implementation hallucinations: divergences between the intended mathematical formulation and the solver program.

Taxonomy:
{taxonomy_block}

Audit instructions:
1. Compare the symbolic model against the code, not just the code against the problem text.
2. Treat the symbolic model as the source of truth for this module. Emit an implementation finding only when the code changes the mathematical object, drops indices, uses the wrong API/domain, or otherwise breaks symbolic-to-code fidelity.
3. If the symbolic formulation itself is already wrong and the code faithfully mirrors it, do not report an implementation hallucination for that issue. Record a dependency or note instead.
4. Check objective sense, variable domains, loop/index expansion, omitted constraints, solver compatibility, and post-solve reporting.
5. When the symbolic model says `minimize` but the code materializes `maximize`, negates objective coefficients, flips the reported objective sign, or sets the wrong solver API sense, prefer `Wrong Objective Sense in Code`.
6. A pure code-level objective-sense reversal is still an implementation hallucination even if the objective terms themselves look numerically similar.
7. Do not flag `Wrong Objective Sense in Code` when the code uses a standard, internally consistent sign-handling convention that still matches the symbolic objective. For example, `minimize` with direct positive coefficients and an inactive post-solve sign flip is not an objective-sense error.
8. Likewise, `maximize` implemented via negated coefficients plus an active post-solve sign correction is not an objective-sense error if it faithfully matches the symbolic objective.
9. Be especially careful with `scipy.optimize.milp` or similar minimization-oriented APIs. A pattern of `c = -c_ref`, followed by solving, followed by `objective_value = -result.fun` for a maximize objective is a standard faithful wrapper, not a hallucination.
10. A dead branch such as `if "minimize" == "maximize": objective_value = -objective_value` inside an otherwise ordinary minimization implementation is not, by itself, `Wrong Objective Sense in Code`. If the branch never executes and the solver call already matches the symbolic sense, abstain.
11. Do not flag `Wrong Bounds in Code` or `Omitted Constraint Materialization` when the only difference is an equivalent presentation style, such as encoding a lower/upper bound in the solver `Bounds` object instead of duplicating it as a separate linear row, or encoding an equality as a pair of matching `>=` and `<=` rows.
12. If the candidate explanation itself admits that the implementation already matches the symbolic model, that no divergence exists, or that no fix is needed, abstain instead of emitting an implementation finding.
13. Prefer precise implementation labels such as code-materialization or index-expansion failures over vague "code bug" language.
14. Use short evidence spans from code and symbolic text.
15. Emit atomic root-cause findings only. Do not restate a symbolic modeling error unless the code introduces an additional independent divergence.
16. If the problem text and symbolic model already disagree on objective direction, do not let a code path that matches the textual objective suppress the upstream objective diagnosis. In that situation, only emit `Wrong Objective Sense in Code` if the code also diverges from the symbolic model in an independently actionable way.
17. If the solver code materializes only a strict subset of the symbolic variables/indices while the symbolic family is otherwise correct, prefer `Partial Set Expansion` or `Dropped Loop or Index Expansion` over symbolic constraint labels.
18. If the symbolic model is a MILP / binary / integer formulation but the executable artifact routes to an LP-only backend such as `scipy.optimize.linprog`, prefer `Incompatible Solver Selection` even when the resulting numeric solution also induces downstream domain mismatches.

Output instructions:
- Return at most {max_findings} findings, but prefer 0-3 unique root-cause findings.
- Include a repair suggestion that refers to the code path or implementation mechanism.
- Fill `canonical_issue` with a short stable phrase such as `binary decision materialized as continuous`.
- Set `is_root_cause` to `true` only for a primary atomic issue. Leave `duplicate_of` empty unless you are explicitly marking an internal duplicate.

Return only the fields required by the structured schema.

Reported-version calibration:
- Prefer abstention over placeholder findings.
- If your own evidence says the current code already matches the symbolic model, do not emit a finding.
- Do not report stylistic cleanups, documentation suggestions, or robustness notes as hallucinations unless they create an actual semantic divergence.
- Never emit a counterfactual guardrail finding. If the current solver code is faithful and you are only warning that a different future rewrite would be wrong, abstain instead.
- Do not flag a standard maximize-via-sign-flip wrapper, an inactive dead branch, or equivalent bound materialization unless the active solver path or reported objective value is actually wrong.
- If heuristic notes mention an invalid API symbol, syntax error, TODO placeholder, or missing `solve_model()` entry point, treat it as a code-only implementation issue rather than a symbolic modeling error.
- Prefer `Partial Set Expansion`, `Dropped Loop or Index Expansion`, or `Missing Variable Registration` when the code only materializes a subset of the symbolic decision family.
- Prefer `Incompatible Solver Selection` when a discrete symbolic model is routed to an LP-only solver backend.
- If the symbolic model is already correct but code only creates part of a variable family, omits a linearization helper, reads back the wrong solver object, or reports a stale / sign-flipped objective value, keep the implementation finding even if the downstream symptom also resembles a symbolic objective or constraint issue.
- Use exact code labels: `Wrong Index Domain Materialization` for wrong RangeSet/index-domain construction, `Partial Set Expansion` for materializing only a subset, `Missing Linearization in Code` for omitted helper rows for logical/nonlinear structure, and `Stale Result Object or Wrong Post-Processing` for stale or wrong result reads.
- Abstain from implementation findings whose own evidence says the solver/code is compatible, equivalent, inactive, or faithful to the symbolic model; implementation requires an active code-level semantic divergence.
\end{Verbatim}
\end{prompttemplate}

\begin{prompttemplate}{Conductor Review Prompt}
\begin{Verbatim}[fontsize=\scriptsize,breaklines,breakanywhere]
You are the Conductor review agent in a taxonomy-grounded optimization-model audit.

You have already received the first-pass outputs from the objective, variable, constraint, and implementation experts.

Your job now is to:
1. Identify cross-agent dependencies.
2. Detect duplicated or overlapping findings.
3. Surface the most important unresolved risks that should affect the final report.
4. Refer to overlaps using the specialists' canonical issue phrases whenever possible.

Important rules:
- Do not invent new findings unless they follow directly from the specialist evidence.
- Prefer dependency statements such as "objective term references undefined variable dimension" or "solver code drops an index needed by the symbolic constraint".
- Keep the result concise and decision-oriented.

Return only the fields required by the structured schema.
\end{Verbatim}
\end{prompttemplate}

\begin{prompttemplate}{Final Judge Prompt}
\begin{Verbatim}[fontsize=\scriptsize,breaklines,breakanywhere]
You are the final judge in a taxonomy-grounded optimization-model hallucination audit.

You do not invent new findings. You only select and reorder from the candidate findings that were already proposed by the specialist agents.

Your job is to:
1. choose the most likely primary root cause,
2. optionally keep one or two additional findings if they are independently useful,
3. suppress generic or derivative findings when a more specific candidate already explains the same problem.

Important rules:
- Select only from the provided `candidate_id` values.
- Prefer more specific subtypes over generic ones.
- Prefer root-cause findings over downstream consequences.
- If a candidate says `Wrong Aggregation Level`, prefer it over a generic capacity-skeleton label when the problem text describes a pooled total, combined sum, or cross-entity aggregate.
- If a candidate says `Relaxing a Discrete Variable into a Continuous Variable`, prefer it over nearby constraint or implementation consequences when the problem is about counts, trips, assignments, facilities, or whole-number decisions.
- Suppress `Wrong Interpretation of the Rule` when the candidate's own evidence says the current model is a plausible literal reading, flags only a semantic ambiguity, or says the model would be acceptable under one natural reading of the text.
- Suppress `Relaxing a Discrete Variable into a Continuous Variable` when the evidence lacks strong integrality cues and instead matches an LP, blending, mixture, or continuous-allocation formulation.
- Prefer `Wrong Objective Sense in Code` only when it is an actual code-level divergence, not just a standard sign-handling convention that matches the symbolic objective.
- Suppress implementation findings when the candidate itself says the code already matches the symbolic model, that no fix is needed, or that the remaining difference is only an equivalent materialization style such as `Bounds` versus explicit rows.
- Suppress `Omitted Index Set` when the candidate's own evidence says the scalar variables are semantically correct and the complaint is only about cleaner notation or scalability.
- Suppress constraint findings when the candidate's own explanation concludes that the current linearization is algebraically correct or that the existing constraints are already correct.
- Usually keep 1-2 findings. Keep 3 only when the secondary findings are clearly independent and not duplicates.

Return only the fields required by the structured schema.
\end{Verbatim}
\end{prompttemplate}

\begin{prompttemplate}{Visualization Prompt}
\begin{Verbatim}[fontsize=\scriptsize,breaklines,breakanywhere]
You are the Visualization agent in a taxonomy-grounded multi-agent optimization-model audit.

You are given aggregated findings and conductor review notes. Produce a concise summary for a human analyst.

Your output should:
1. State the headline judgment.
2. Summarize which modules dominate the risk.
3. Suggest a practical repair order.
4. Add short analyst notes that help someone read the final markdown report quickly.

Stay faithful to the evidence. Do not invent findings that are not present in the aggregated input.

Return only the fields required by the structured schema.
\end{Verbatim}
\end{prompttemplate}

\subsection{Single-Agent Detector}
\label{app:single_agent_prompt}

\textbf{Role.} For completeness, we conclude the appendix with the monolithic reference detector used in our comparisons. The Single-Agent Detector audits the full tuple $(P,M,S)$ in a single LLM call, without conductor routing, specialist decomposition, or cross-agent review.

\textbf{Design goal.} Its purpose is not to serve as a weak strawman, but to provide the cleanest architectural control for isolating the value of multi-agent auditing. It uses the same taxonomy, the same artifact triple, and the same abstention-oriented evaluation protocol as OptArgus; what changes is the auditing architecture rather than the task definition.

\textbf{Input-output behavior.} Given the natural-language problem description, symbolic formulation, and solver code, the detector emits a possibly empty set of fine-grained findings. Each finding must commit to an exact module, subcategory, and specific type in the paper's taxonomy, together with supporting evidence and a repair-oriented explanation.

\textbf{Inference pattern.} Because all reasoning must happen in one pass, the Single-Agent Detector must jointly choose the relevant branch, resolve cross-module ambiguities, decide whether to abstain, and produce final taxonomy labels without the intermediate memory or targeted follow-up available to OptArgus. The exact prompt is given below.

\begin{prompttemplate}{Single-Agent Detector Prompt}
\begin{Verbatim}[fontsize=\scriptsize,breaklines,breakanywhere]
System:
You are a single-agent hallucination detector for optimization modeling.

You receive three artifacts:
1. a natural-language optimization problem P,
2. a symbolic mathematical model M,
3. solver-ready code S.

Your job is to identify the most likely hallucination-like mismatches using the
paper's taxonomy across four modules:
- objective
- variable
- constraint
- implementation

Use the taxonomy exactly at the level of:
- module
- subcategory
- specific_type

Do not invent new labels, abbreviate subcategories, collapse distinct
subcategories into generic placeholders, or stop at the subcategory level
without selecting a valid specific_type. Our downstream evaluation checks exact
specific_type matches, so every emitted finding must choose one exact
specific_type from the allowed list under its chosen subcategory.

Taxonomy:

Objective:
- Objective Semantic Mapping Errors
  - Wrong Optimization Direction
  - Objective Substitution
  - Softening a Hard Constraint
  - Turning a Preference into a Hard Constraint
- Objective Composition Errors
  - Omission of a Key Objective Term
  - Introduction of a Spurious Objective Term
  - Repeated Counting of an Objective Term
  - Wrong Sign on a Local Objective Term
- Objective Parameter and Index Binding Errors
  - Coefficient Misbinding
  - Wrong Index Range
  - Misaligned Time Indices
  - Wrong Unit or Scale Binding
- Multi-Objective and Risk-Objective Errors
  - Wrong Multi-Objective Aggregation Mechanism
  - Replacing a Risk or Extreme-Value Objective
  - Wrong Weight Setting or Interpretation
- Objective Formalization Errors
  - Smoothing Piecewise or Stepwise Cost into a Continuous Linear Cost
  - Incorrect Handling of Nonlinear or Nonsmooth Objective Terms
  - Fractional Objective Errors

Variable:
- Variable Semantic Definition Errors
  - Wrong Variable Object
  - Missing a Key Dimension
  - Wrong Aggregation Level of Variables
  - State-Control Confusion
- Variable Domain and Type Errors
  - Relaxing a Discrete Variable into a Continuous Variable
  - Forcing a Continuous Variable to Be Discrete
  - Wrong Value Range
  - Wrong Sign Domain
- Variable Index and Set Binding Errors
  - Omitted Index Set
  - Fabricated Index Set
  - Wrong Subscript Binding
  - Missing Index Dependence
- Variable Role Coupling Errors
  - Variable Not Taking Effect
  - Duplicate Variable Roles
  - Missing Coupling Between Master and Auxiliary Variables
- Variable Redundancy and Uncontrolled Symmetry
  - Unnecessary Redundant Variables
  - Unhandled Symmetric Variables
  - Failure to Exploit Sparsity

Constraint:
- Constraint Semantic Translation Errors
  - Wrong Interpretation of the Rule
  - Wrong Interpretation of Quantifiers
  - Confusing Resource Amount with Time Feasibility
  - Constraint Type Mismatch
- Missing Constraint Skeleton
  - Missing Balance, Conservation, or Recursion Constraints
  - Missing Chain or Path Structure Constraints
  - Missing Capacity or Coverage Skeleton
  - Missing Initial or Terminal Conditions
- Missing Implicit Constraints
  - Missing Default Business Rules
  - Missing Must-Include Object Constraints
  - Missing Legal-Action Set Restrictions
- Erroneous Equivalent Substitutions
  - Replacing an Equality with an Inequality
  - Using Local or Partial Conditions to Replace a Full Structure
  - Treating an Approximate Constraint as an Exact One
- Logic and Conditional-Chain Errors
  - Big-M Direction Error
  - Big-M Too Loose or Too Tight
  - Incomplete Logical Linearization
  - Broken Conditional Chains
- Aggregation and Index-Coding Errors
  - Wrong Aggregation Level
  - Wrong Summation Range
  - Wrong Subscripts When Calling a Constraint
  - Repeated or Missing Aggregation
- Constraint Boundary and Direction Errors
  - Wrong Inequality Direction
  - Swapped Upper and Lower Bounds
  - Misplaced Boundary Conditions
  - Misuse of Threshold or Boundary Parameters
- Constraint Strength Imbalance
  - Constraint Too Weak
  - Constraint Too Strong
  - Slack or Tolerance Parameters on the Wrong Scale
- Scheduling- and Activity-Structure-Specific Errors
  - Wrong Encoding of Precedence Relations
  - Wrong Structure for Mode Selection and Optional Activities

Implementation:
- Symbolic-Code Mismatch
  - Wrong Objective Sense in Code
  - Omitted Constraint Materialization
  - Missing Variable Registration
  - Stale or Divergent Coefficient Mapping
- Variable and Domain API Mismatch
  - Wrong API Variable Type
  - Wrong Bounds in Code
  - Wrong Index Domain Materialization
- Index and Set Materialization Errors
  - Dropped Loop or Index Expansion
  - Partial Set Expansion
  - Filtered Index Loss
- Solver and Formalization Mismatch
  - Missing Linearization in Code
  - Incompatible Solver Selection
  - Incorrect Big-M or Numerical Option in Code
- Post-Solve Extraction and Reporting Divergence
  - Wrong Solution Variable Readout
  - Misreported Objective Value
  - Stale Result Object or Wrong Post-Processing

Audit rules:
- Emit atomic root-cause findings only.
- Do not force all four modules to appear.
- Use the most specific subcategory and specific_type supported by evidence.
- If the issue belongs to implementation, compare S against M, not directly
  against P.
- If M is wrong but S faithfully mirrors M, do not fabricate an implementation
  hallucination for that same issue.
- If a pooled total, combined sum, joint budget, or cross-entity aggregate is
  dropped or replaced by separate local bounds, prefer
  constraint / Aggregation and Index-Coding Errors / Wrong Aggregation Level.
- If a resource, demand, coverage, flow, recursion, path, or endpoint
  structure is entirely missing, prefer the relevant Missing Constraint
  Skeleton subtype.
- If the issue is a policy rule, share rule, default eligibility rule, or
  must-include business logic, prefer Missing Implicit Constraints rather than
  a generic capacity label.
- If the issue is a discrete/count/trip/facility/assignment variable relaxed to
  continuous, prefer variable / Variable Domain and Type Errors / Relaxing a
  Discrete Variable into a Continuous Variable.
- Do not infer discreteness from weak wording alone in clearly continuous LP,
  blending, nutrition, mixture, or allocation formulations.
- If a standard sign-handling template still matches the symbolic objective, do
  not call it implementation / Symbolic-Code Mismatch / Wrong Objective Sense
  in Code.
- If the decision variables are already budget / spend amounts in dollars and
  the symbolic objective minimizes their sum, treat that as a faithful
  total-cost objective unless the text explicitly says the variables should
  instead represent effectiveness units or some separately priced quantity.
- Do not reinterpret a literally stated minimum profit / least profit /
  minimize profit objective as maximization using business intuition alone.
  Only call Wrong Optimization Direction when the text itself explicitly
  contradicts the symbolic sense.
- Do not flag Wrong Bounds in Code or Omitted Constraint Materialization when
  the only difference is an equivalent presentation style, such as:
  - a bound expressed in solver Bounds instead of a separate linear row,
  - a lower/upper bound encoded as explicit constraints instead of variable
    metadata,
  - an equality encoded as matching >= and <= rows.
- If your own explanation says the current implementation is already faithful,
  no divergence exists, or no fix is needed, abstain.

Output rules:
- Return a valid JSON object that matches the structured schema exactly.
- Return only the fields required by the structured schema.
- Prefer 0-3 findings.
- Each finding must include the correct paper-style module, subcategory, and
  specific_type.
- Never output a finding with only a generic major-category label such as objective,
  variable, constraint, or implementation as the subcategory.
- Never output a finding with a vague fallback specific type such as other,
  misc, generic mismatch, or modeling bug.
- For each finding, the specific_type must be one of the listed types under the
  chosen subcategory; do not mix a subcategory from one family with a specific
  type from another family.
- If evidence is insufficient to choose a valid exact specific_type, abstain
  instead of emitting a coarse finding.

User:
Problem specification:
{problem_text}

Symbolic model:
{symbolic_model_text}

Solver code:
{solver_code_text}
\end{Verbatim}
\end{prompttemplate}

This verifier is deliberately prompt-centric. In the experiments, its outputs are parsed under the same structured schema and scored under the same controlled and natural protocols as OptArgus. Its role is therefore twofold: it is a practically usable monolithic detector, and it is the cleanest architectural reference against which the benefits of routed multi-agent auditing can be measured.


\section{Benchmark Construction, OR-Expert Annotation, and Metrics}
\label{app:data_construction}

This appendix explains how the three evaluation benchmarks are constructed and annotated under a unified OR-expert protocol. The central principle is that neither a symbolic formulation nor a solver implementation is treated as trustworthy merely because it compiles, solves, or matches a reference objective value on one instance. Instead, the \texttt{clean}, \texttt{injected}, and \texttt{natural} benchmarks are all anchored in expert-validated reference artifacts and benchmark-specific annotation rules designed for optimization-modeling hallucination detection. We organize the appendix in the same order as the evaluation suite itself. We first explain why objective-value agreement is an insufficient supervisory signal. We then present the \texttt{clean}, \texttt{injected}, and \texttt{natural} benchmarks in turn. Within each benchmark subsection, we first describe the OR-expert construction and annotation protocol, and then define the metrics used for that benchmark. We place the objective-agreement discussion first because it motivates all three benchmarks; the \texttt{natural}-benchmark subsection later returns to this point with a supplementary model-panel summary.

\subsection{Why Objective Agreement Is Not Enough}
\label{app:data_obj_match}

A recurring temptation in optimization-modeling evaluation is to treat objective-value agreement as if it were equivalent to structural correctness. In our setting, this is too weak. Let
\begin{equation}
V_{\mathrm{obj}}(P,\widehat{M},\widehat{S})
=
\mathbf{1}\!\left[
 f(\widehat{S}) = f(S^*)
\right]
\end{equation}
denote the weakest behavioral check, namely agreement of the reported optimal value. Here $P$ is the natural-language problem description, $\widehat{M}$ and $\widehat{S}$ are the audited symbolic model and solver implementation, $S^*$ is the certified reference implementation for the same seed when available, $f(S)$ denotes the objective value reported by executable artifact $S$ on the benchmark instance, and $\mathbf{1}[\cdot]$ is the indicator function, so $V_{\mathrm{obj}}(P,\widehat{M},\widehat{S})=1$ exactly when the two objective values agree and $0$ otherwise. However, objective agreement is only a single end-of-pipeline observation; it does not tell us whether $\widehat{M}$ correctly captures the optimization semantics of $P$, whether $\widehat{S}$ faithfully realizes $\widehat{M}$, or whether the full tuple $(P,\widehat{M},\widehat{S})$ is structurally consistent.
\begin{equation}
V_{\mathrm{obj}}(P,\widehat{M},\widehat{S})=1
\;\not\Rightarrow\;
\mathcal{R}_{\mathrm{sym}}(P,\widehat{M})=1,
\end{equation}
\begin{equation}
V_{\mathrm{obj}}(P,\widehat{M},\widehat{S})=1
\;\not\Rightarrow\;
\mathcal{R}_{\mathrm{imp}}(\widehat{M},\widehat{S})=1,
\end{equation}
and therefore
\begin{equation}
V_{\mathrm{obj}}(P,\widehat{M},\widehat{S})=1
\;\not\Rightarrow\;
 \mathcal{R}(P,\widehat{M},\widehat{S})=1.
\end{equation}
The first non-implication concerns the map from $P$ to $\widehat{M}$: a symbolic artifact can match the reference objective value even when it weakens an ``exactly two'' rule into ``at most two,'' omits an objective term that happens to be zero on the tested instance, or relaxes a discrete decision into a continuous one without changing the optimum on that data instance. The second concerns the map from $\widehat{M}$ to $\widehat{S}$: solver code can still reproduce the same objective value even when it omits a constraint family, drops an index expansion, or flips the solver-side objective sense in a case where the affected error is numerically masked. At the same time, the converse caution also matters: two implementations can look different while remaining semantically equivalent, such as a standard sign-flip wrapper used to implement maximization with a minimization-oriented solver API. The evaluation problem is therefore structural rather than merely behavioral: we must reason jointly about whether the problem description, symbolic model, and executable implementation remain aligned, rather than relying on final objective-value agreement alone. The thirteen-model panel summary in Table~\ref{tab:appendix_v484_generator_snapshot} later in this appendix provides an operational illustration of this gap.

\subsection{Clean Benchmark}
\label{app:clean_benchmark}

This benchmark evaluates \emph{restraint} on correct artifacts: when an optimization artifact is genuinely sound, a good detector should abstain rather than invent a plausible-looking error. Accordingly, we first describe how the clean reference pool is certified by OR experts, and then define the abstention-oriented metrics used to evaluate false alarms on these correct artifacts.

\subsubsection{Construction and OR-Expert Validation}
\label{app:clean_construction}

We begin from five community benchmark families---IndustryOR~\citep{huang2025orlm}, MamoComplexLP~\citep{huang2025mamo}, MamoEasyLP~\citep{huang2025mamo}, NL4Opt~\citep{Ramamonjison2023NL4Opt}, and ReSocratic~\citep{yang2024optibench}---and map each raw record to a canonical seed object
\begin{equation}
\zeta_i
=
\left(
P_i,
A_i^{\mathrm{src}},
\nu_i,q_i
\right),
\end{equation}
where $P_i$ is the natural-language problem statement, $A_i^{\mathrm{src}}$ collects benchmark-side supporting artifacts, $\nu_i$ stores provenance metadata, and $q_i$ records quality flags. These source materials serve only as inputs to expert curation: they provide realistic optimization problems and supporting artifacts from which clean references can be certified.

Because the clean benchmark is designed to test false-alarm restraint on correct artifacts, the key requirement is not merely to retain many seeds, but to certify that every retained tuple is genuinely sound. Starting from the raw seed pool, we retain only candidates for which OR experts can certify three conditions simultaneously: the problem statement is materially unambiguous for optimization modeling, the reference answer and supporting artifacts are correct, and a complete symbolic-and-code realization can be validated. For each candidate seed, we therefore normalize notation, units, set names, and indexing conventions, and then construct a provisional symbolic formulation $\widehat{M}$ and a provisional executable solver realization $\widehat{S}$. In the final certification stage, six OR experts cross-validate each retained seed by checking the problem statement, the normalized symbolic formulation, and the executable artifact side by side. A symbolic formulation is accepted only if the reviewers confirm semantic traceability to the text, coverage of all hard requirements, and internal mathematical well-posedness. The executable realization is accepted only if the reviewers additionally confirm symbolic-to-code fidelity, agreement with the certified reference answer on the source instance, and robustness under small perturbation checks. Formally, we use a two-stage certification notation:
\begin{align}
(P,\widehat{M}) \in \mathcal{D}_{\mathrm{sym}}
&\iff
V_M(P,\widehat{M})=1, \\
(P,\widehat{M},\widehat{S}) \in \mathcal{D}_{\mathrm{full}}
&\iff
V_M(P,\widehat{M})=1
\land
V_S(P,M^*,\widehat{S})=1,
\end{align}
where $\widehat{M}$ and $\widehat{S}$ are the provisional symbolic and executable artifacts constructed during curation, $\mathcal{D}_{\mathrm{sym}}$ is the pool of symbolically certified cases, $\mathcal{D}_{\mathrm{full}}$ is the pool of fully certified $(P,M,S)$ tuples, $V_M(P,\widehat{M})=1$ means that $\widehat{M}$ is semantically faithful to $P$, $M^*$ is the certified symbolic reference distilled from $\widehat{M}$, and $V_S(P,M^*,\widehat{S})=1$ means that $\widehat{S}$ faithfully realizes $M^*$ as executable solver code. Only seeds that reach $\mathcal{D}_{\mathrm{full}}$ enter the final clean pool, so every retained clean reference contains a certified problem statement, a certified symbolic model, and a certified executable implementation.

The final clean benchmark contains $484$ OR-expert-validated reference artifacts: IndustryOR ($10$), MamoComplexLP ($36$), MamoEasyLP ($229$), NL4Opt ($69$), and ReSocratic ($140$).

\subsubsection{Metrics}
\label{app:clean_metrics}

Let $\mathcal{C}_{\mathrm{clean}}=\{1,\dots,N_{\mathrm{clean}}\}$ be the clean case set, with $N_{\mathrm{clean}}=484$ in the present benchmark, and let $F_i$ denote the final ranked finding list returned for case $i$. We report
\begin{align}
\operatorname{EmptyReportRate}
&=
\frac{1}{N_{\mathrm{clean}}}
\sum_{i=1}^{N_{\mathrm{clean}}}
\mathbf{1}[F_i=\varnothing], \\
\operatorname{MeanFindings}_{\mathrm{clean}}
&=
\frac{1}{N_{\mathrm{clean}}}
\sum_{i=1}^{N_{\mathrm{clean}}}|F_i|.
\end{align}
Intuitively, these metrics ask whether the detector can stay quiet on artifacts that are actually correct, but they capture different failure modes and are therefore not redundant. $\operatorname{EmptyReportRate}\in[0,1]$, and higher is better: a value of $1$ means that the detector always returns no finding on clean cases, whereas a value of $0$ means that it always raises at least one finding even when nothing is wrong. By contrast, $\operatorname{MeanFindings}_{\mathrm{clean}}\in[0,\infty)$, and lower is better: a value of $0$ means perfect silence on the clean benchmark, while larger values indicate increasing false-alarm verbosity. The first metric is case-level and binary---did the detector abstain?---whereas the second is count-based and reveals how verbose the false alarms are once abstention fails. Two detectors can therefore attain similar $\operatorname{EmptyReportRate}$ yet differ substantially in how many spurious findings they emit on the cases they do flag, which is why both metrics are retained.

\subsection{Injected Benchmark}
\label{app:injected_benchmark}

This benchmark turns the fine-grained optimization-modeling hallucination taxonomy in App.~\ref{app:taxonomy} into a controlled diagnosis task. Starting from the four major categories, their intermediate subcategories, and the $83$ specific hallucination types summarized in Table~\ref{tab:hallucination-taxonomy-summary}, we construct one injected case at a time so that every artifact carries one known gold major category, one known subcategory, and one known specific hallucination type. We deliberately keep this setting single-error-per-case. Doing so isolates one taxonomy-grounded corruption at a time, keeps the supervision unambiguous, and prevents interactions among overlapping errors from obscuring whether a miss is due to true localization failure or merely to cross-error interference. As a result, the Top-1 major-category, subcategory, and specific-type metrics remain directly interpretable as measures of taxonomy-level diagnostic precision. Accordingly, we first describe how \texttt{injected} instantiates this optimization-modeling hallucination taxonomy as concrete corruption cases, and then define the Top-1 metrics used to score localization quality at each taxonomy level.

\subsubsection{Construction and OR-Expert Error Injection}
\label{app:injected_construction}

\begin{table}[h]
\centering
\caption{Coverage summary of the controlled benchmark \texttt{injected}. ``Full'' denotes specific hallucination types that reach the target of $30$ cases, ``Partial'' denotes specific hallucination types with at least one but fewer than $30$ defensible constructions, and ``Shortage-only'' denotes specific hallucination types that are tracked but cannot currently be constructed cleanly from the active $484$-seed universe.}
\label{tab:injected_coverage_summary}
\small
\setlength{\tabcolsep}{5pt}
\renewcommand{\arraystretch}{1.08}
\begin{tabular}{lccccc}
\toprule
Major Category & Specific Types & Built Cases & Full & Partial & Shortage-only \\
\midrule
Objective & 18 & 271 & 9 & 1 & 8 \\
Variable & 18 & 355 & 9 & 6 & 3 \\
Constraint & 31 & 257 & 5 & 13 & 13 \\
Implementation & 16 & 383 & 12 & 3 & 1 \\
\midrule
Total & 83 & 1266 & 35 & 23 & 25 \\
\bottomrule
\end{tabular}
\end{table}

\textbf{Construction principle.} We use the same $484$ clean certified references as the source pool for controlled error injection. The resulting benchmark is the expanded \texttt{injected} benchmark used in our evaluation, which is built type by type from the $83$ specific hallucination types summarized in Table~\ref{tab:hallucination-taxonomy-summary} and defined in App.~\ref{app:taxonomy}. For a given specific hallucination type, we first determine which clean references can support that corruption without introducing side effects; only those eligible references are used for construction. We then inject exactly one deliberate structural error into an otherwise correct artifact and keep the result only if it remains a clean realization of that taxonomy item.

\textbf{Sampling regimes.} The nominal target is up to $30$ cases per specific hallucination type, but only when that corruption can be instantiated defensibly on the active clean seed pool. For broad, high-coverage types, such as \emph{Wrong Optimization Direction}, \emph{Omission of a Key Objective Term}, \emph{Relaxing a Discrete Variable into a Continuous Variable}, \emph{Wrong Inequality Direction}, and \emph{Wrong Objective Sense in Code}, we use benchmark-family-stratified sampling with fixed-seed randomization inside each family in order to preserve case diversity while avoiding unnecessary concentration on the same source seeds. For rarer or more structurally specialized types, such as \emph{Smoothing Piecewise or Stepwise Cost into a Continuous Linear Cost}, \emph{State-Control Confusion}, \emph{Big-$M$ Direction Error}, \emph{Big-$M$ Too Loose or Too Tight}, \emph{Wrong Encoding of Precedence Relations}, and \emph{Wrong Structure for Mode Selection and Optional Activities}, we instead search the validated seed universe exhaustively, keep all defensible constructions, and record an explicit shortage whenever the active $484$-seed universe does not support $30$ clean cases.

\textbf{Coverage and verification.} The resulting coverage summary is shown in Table~\ref{tab:injected_coverage_summary}. The full benchmark specification tracks all $83$ specific hallucination types, but only $58$ of them currently admit at least one clean construction from the active seed universe. Concretely, \texttt{injected} contains $1266$ built cases across those $58$ realizable types. Among the $83$ tracked types, $35$ are \emph{full}, meaning that they reach the target of $30$ cases; $23$ are \emph{partial}, meaning that they admit at least one but fewer than $30$ defensible constructions; and the remaining $25$ are \emph{shortage-only}, meaning that they are retained in the taxonomy but do not currently admit a clean construction from the active seed pool. Every injected artifact inherits its specific-type label by construction and is then checked by OR experts against both the certified clean reference and the intended corruption recipe. The resulting benchmark therefore provides exact controlled supervision for major category, subcategory, and specific hallucination type without relying on machine-generated labels.

\subsubsection{Metrics}
\label{app:injected_metrics}

The injected benchmark is designed to evaluate localization quality at three nested resolutions. Because every case contains exactly one known injected hallucination, the detector should ideally place its first returned diagnosis in the correct major category, then in the correct taxonomy subcategory, and finally on the correct specific type. For this reason, the most informative controlled-benchmark metrics are Top-1 localization metrics rather than broad set-overlap scores. We also report the mean number of returned findings, because a diagnosis is more useful when it is both correct and concise.

Formally, for the controlled injected benchmark \texttt{injected}, let $\mathcal{C}=\{1,\dots,N\}$ be the evaluated case set, let $\mathcal{B}=\{\texttt{objective},\texttt{variable},\texttt{constraint},\texttt{implementation}\}$ be the major-category set, let
\[
F_i=\{f_{i1},\dots,f_{i|F_i|}\}
\]
be the final ranked finding list for case $i\in\mathcal{C}$, and let $b_i$, $s_i$, and $t_i$ denote the gold major category, subcategory, and specific hallucination type. Writing $\operatorname{top}(F_i)=f_{i1}$ when $|F_i|>0$, we report
\begin{align}
\operatorname{Top1MajorCategoryHit}
&=
\frac{1}{N}\sum_{i=1}^{N}
\mathbf{1}\!\left[F_i\neq\varnothing \land \operatorname{branch}(\operatorname{top}(F_i))=b_i\right], \\
\operatorname{Top1SubcategoryHit}
&=
\frac{1}{N}\sum_{i=1}^{N}
\mathbf{1}\!\left[F_i\neq\varnothing \land \operatorname{branch}(\operatorname{top}(F_i))=b_i \land \operatorname{subcat}(\operatorname{top}(F_i))=s_i\right], \\
\operatorname{Top1SpecificTypeHit}
&=
\frac{1}{N}\sum_{i=1}^{N}
\mathbf{1}\!\left[F_i\neq\varnothing \land \operatorname{branch}(\operatorname{top}(F_i))=b_i \land \operatorname{type}(\operatorname{top}(F_i))=t_i\right], \\
\operatorname{MeanFindings}
&=
\frac{1}{N}\sum_{i=1}^{N}|F_i|.
\end{align}
These metrics have complementary roles. $\operatorname{Top1MajorCategoryHit}$ asks the coarsest controlled-diagnosis question: does the detector immediately land in the correct major category? $\operatorname{Top1SubcategoryHit}$ asks whether the first returned diagnosis is already correct at the next taxonomy level, and $\operatorname{Top1SpecificTypeHit}$ asks whether the detector identifies the correct specific hallucination type with its top-ranked finding. All three Top-1 metrics lie in $[0,1]$, with higher values indicating better localization accuracy and a value of $1$ meaning perfect first-hit localization on that channel.

$\operatorname{MeanFindings}\in[0,\infty)$ instead measures diagnostic verbosity. Lower values indicate shorter and more concise reports, but this metric is not intended to stand alone. A detector could achieve a low mean report length simply by saying very little, which is only desirable if the Top-1 localization metrics remain high. Conversely, two detectors can attain similar Top-1 accuracy yet differ substantially in how much extraneous diagnostic material they return. In short, the Top-1 metrics measure localization quality, whereas $\operatorname{MeanFindings}$ measures how efficiently that diagnosis is delivered.

\subsection{Natural Benchmark}
\label{app:natural_benchmark}

This benchmark is designed to complement the \texttt{clean} and \texttt{injected} settings by testing whether a detector remains effective once hallucinations arise \emph{naturally} from real modeling systems rather than from controlled single-error construction. In this setting, errors are open-ended: some artifacts are fully correct, some contain only one mistake, and many exhibit overlapping symbolic and implementation failures whose severity and major-category composition are highly uneven across models. The \texttt{natural} benchmark therefore evaluates detector performance under a realistic LLM-generated error distribution, not just exact localization under laboratory-style perturbations. Accordingly, we first describe the model panel and the OR-expert annotation protocol used to build this natural distribution, and then define the F1-based metrics used to evaluate artifact-level and major-category-level detection quality under it.

\subsubsection{Model Panel and OR-Expert Annotation}
\label{app:natural_construction}

\textbf{Panel design.} We begin from the same $484$ clean certified seeds and assemble a natural-model panel with two complementary model classes. The first class consists of frontier general-purpose models, including both open-weight and proprietary systems: \texttt{DeepSeek-v3.2}, \texttt{Gemini-2.5-flash-lite}, \texttt{GLM-4.5-air}, \texttt{Kimi-K2.5}, \texttt{MiniMax-M2.5}, \texttt{Qwen3-4b}, \texttt{Qwen3-8b}, \texttt{Qwen3-32b}, and \texttt{Qwen3.5-Plus}. We include this class to reflect realistic deployment conditions and to cover a heterogeneous set of general LLMs available during benchmark construction, spanning multiple providers, both API-only and open-weight access modes, and a range of model scales.

\textbf{Fine-tuned class.} The second class consists of fine-tuned models for optimization modeling: \texttt{LLMOPT-Qwen2.5-14B}~\citep{jiang2025llmopt}, \texttt{OptMATH-Qwen2.5-7B}~\citep{lu2025optmath}, \texttt{ORLM-LLaMA-3-8B}~\citep{huang2025orlm}, and \texttt{SIRL-Qwen2.5-7B}~\citep{chen2025solver}. We include this class because natural hallucination detection should also be tested against models that are explicitly adapted to optimization modeling rather than only against general-purpose LLMs. Within this class, these four systems are the representative and publicly released fine-tuned models for optimization modeling that we were able to collect and run end-to-end on the common seed pool. Together these thirteen models define the finalized natural benchmark panel used in the paper, which yields $6292 = 13 \times 484$ generated artifacts.

\textbf{Exclusion policy.} We do not include some even stronger commercial models, such as \texttt{GPT-5.4} or \texttt{Gemini-3.1-Pro}, in the final panel because the current $484$-seed benchmark is relatively easy for them and would produce too few natural hallucination cases to support an informative large-scale annotation study. This should not be read as evidence that such models are free of optimization-modeling hallucinations: on substantially harder optimization tasks, such as difficult instances from MIPLIB-NL~\citep{li2026constructing} or OptMATH-bench~\citep{lu2025optmath}, they can still fail structurally. However, those harder problems are also considerably more difficult for human experts to certify and label at scale, making a large OR-expert-annotated natural benchmark substantially more costly and challenging.

\textbf{Annotation protocol.} Each generated artifact is then reviewed manually by OR experts. For every $(P,M,S)$ tuple, the reviewers inspect the original problem statement, the generated symbolic model, and the generated solver code against the certified references, and they assign (i) artifact-level correctness and (ii) first-level major-category positives over objective, variable, constraint, and implementation. To avoid double counting, a downstream implementation label is not added when the code merely mirrors an upstream symbolic error; implementation positives are reserved for genuine symbolic-to-code divergence.

\textbf{Label granularity.} We deliberately stop the benchmark labels at this first major-category level. The reason is practical rather than conceptual. The natural panel already contains $6292$ artifacts, and deeper taxonomy levels expand rapidly in both label inventory and annotation burden. At the second and third taxonomy levels, the number of possible labels grows substantially, the boundary between nearby subtypes becomes harder to adjudicate consistently, and a single natural artifact may exhibit several interacting hallucinations at once rather than one clean isolated error. In addition, some downstream symptoms are better interpreted as consequences of an upstream modeling mistake than as independent subtype-level positives. Exhaustively assigning exact subtypes for every natural artifact would therefore be much more expensive, slower to quality-control, and less reliable at the current scale. For this reason, the natural benchmark used in this paper uses OR-expert artifact-level and major-category labels as its primary supervision, yielding $6292$ annotated artifacts over the full thirteen-model panel.

\begin{table*}[h]
\centering
\small
\caption{Thirteen-model panel summary on the same $484$ certified seeds, grouped into general models and fine-tuned models for optimization modeling; all listed models generated artifacts for the full $484/484$ seed set, and together they define the natural benchmark panel used in the paper. `Class' identifies the two model groups, `Model' names the specific system, and `Param.\ Scale' reports the public parameter size when available. In the last four metric columns, percentages are shown first and raw counts are shown in parentheses. `Objective-Value Match' reports how often a model reproduces the certified objective value on the benchmark instance, `Full-Correct' reports how often the entire generated $(P,M,S)$ artifact is judged correct by OR experts, `Hallucination Ratio' reports the artifact-level hallucination share over the full seed set, and `Hallucination within Matches' reports the share of objective-matched artifacts that are still judged hallucinated. For `Objective-Value Match', `Full-Correct', and `Hallucination Ratio', the denominator is the full $484$ artifacts for that model; for `Hallucination within Matches', the denominator is the model-specific number of objective-matched artifacts.}
\label{tab:appendix_v484_generator_snapshot}
\setlength{\tabcolsep}{4pt}
\resizebox{\textwidth}{!}{%
\begin{tabular}{lllcccc}
\toprule
Class & Model & Param. Scale & Objective-Value Match & Full-Correct & Hallucination Ratio & Hallucination within Matches \\
\midrule
\multirow{9}{*}{\parbox{1.55cm}{\centering General\\models}} & \texttt{DeepSeek-v3.2} & 685B & 85.5\% (414/484) & 81.2\% (393/484) & \heatcell{19}{18.8\% (91/484)} & \heatcell{5}{5.1\% (21/414)} \\
& \texttt{Gemini-2.5-flash-lite} & undisclosed & 81.4\% (394/484) & 67.8\% (328/484) & \heatcell{32}{32.2\% (156/484)} & \heatcell{17}{16.8\% (66/394)} \\
& \texttt{GLM-4.5-air} & 106B total / 12B active & 82.2\% (398/484) & 77.9\% (377/484) & \heatcell{22}{22.1\% (107/484)} & \heatcell{5}{5.3\% (21/398)} \\
& \texttt{Kimi-K2.5} & undisclosed & 88.6\% (429/484) & 85.1\% (412/484) & \heatcell{15}{14.9\% (72/484)} & \heatcell{4}{4.0\% (17/429)} \\
& \texttt{MiniMax-M2.5 (old batch)} & 229B & 80.2\% (388/484) & 79.3\% (384/484) & \heatcell{21}{20.7\% (100/484)} & \heatcell{1}{1.0\% (4/388)} \\
& \texttt{Qwen3-4b} & 4.0B & 15.7\% (76/484) & 13.4\% (65/484) & \heatcell{87}{86.6\% (419/484)} & \heatcell{15}{14.5\% (11/76)} \\
& \texttt{Qwen3-8b} & 8.0B & 77.5\% (375/484) & 72.1\% (349/484) & \heatcell{28}{27.9\% (135/484)} & \heatcell{7}{6.9\% (26/375)} \\
& \texttt{Qwen3-32b} & 32.0B & 70.9\% (343/484) & 53.5\% (259/484) & \heatcell{46}{46.5\% (225/484)} & \heatcell{24}{24.5\% (84/343)} \\
& \texttt{Qwen3.5-Plus} & undisclosed & 84.5\% (409/484) & 79.8\% (386/484) & \heatcell{20}{20.2\% (98/484)} & \heatcell{6}{5.6\% (23/409)} \\
\midrule
\multirow{4}{*}{\parbox{1.75cm}{\centering Fine-tuned\\models}} & \texttt{LLMOPT-Qwen2.5-14B} & 14.0B & 76.2\% (369/484) & 65.9\% (319/484) & \heatcell{34}{34.1\% (165/484)} & \heatcell{14}{13.6\% (50/369)} \\
& \texttt{OptMATH-Qwen2.5-7B} & 7.0B & 83.1\% (402/484) & 71.1\% (344/484) & \heatcell{29}{28.9\% (140/484)} & \heatcell{14}{14.4\% (58/402)} \\
& \texttt{ORLM-LLaMA-3-8B} & 8.0B & 73.1\% (354/484) & 50.4\% (244/484) & \heatcell{50}{49.6\% (240/484)} & \heatcell{31}{31.1\% (110/354)} \\
& \texttt{SIRL-Qwen2.5-7B} & 7.0B & 90.3\% (437/484) & 71.9\% (348/484) & \heatcell{28}{28.1\% (136/484)} & \heatcell{20}{20.4\% (89/437)} \\
\bottomrule
\end{tabular}%
}
\end{table*}

\paragraph{Model-panel summary.}
\label{app:data_generator_snapshot}
Table~\ref{tab:appendix_v484_generator_snapshot} summarizes the same thirteen-model panel over the same $484$ certified seeds. Its role is descriptive rather than supervisory: it helps quantify how often objective-value agreement coexists with structural hallucinations across the models in the natural benchmark panel. The `Class' column separates the frontier general-purpose models from the fine-tuned models for optimization modeling. The `Model' column identifies the specific model, and `Param.\ Scale' reports the public parameter size when available, or `undisclosed' when the provider does not release one. `Objective-Value Match' reports the objective-match rate first and the supporting count in parentheses; it measures how often a model reproduces the certified objective value on the source instance, regardless of whether the symbolic model or code is structurally correct. `Full-Correct' uses the same `percentage (count/484)' format and reports the share of artifacts that are judged fully correct by OR experts, meaning that the generated $(P,M,S)$ tuple contains no artifact-level error and no positive major-category label. `Hallucination Ratio' also uses the `percentage (count/484)' format and reports the artifact-level error share over the full seed set. Finally, `Hallucination within Matches' uses `percentage (count/objective-matched outputs)' and reports the conditional error rate among the objective-matched artifacts, so it directly measures how often objective agreement hides a structural mistake. The first five columns are mainly descriptive, whereas the last two are the most direct evidence that objective agreement alone is not a substitute for OR-expert major-category labeling.

To expose the major-category composition more directly, Table~\ref{tab:appendix_v484_branch_mix} reports, for each model, the major-category rate first and the supporting count in parentheses for objective, variable, constraint, and implementation labels, together with the artifact-level \emph{Any Hallucination} rate. The rows are grouped into the same two model classes used in the natural benchmark construction.

\begin{table*}[h]
\centering
\footnotesize
\caption{OR-expert major-category composition of the natural benchmark by model. In every major-category column and in the final ``Any Hallucination'' column, percentages are shown first and raw counts are shown in parentheses, always with the full $484$ artifacts generated by that model as the denominator. The final ``Any Hallucination'' column is an artifact-level union over the four major categories, not their arithmetic sum, because one natural artifact may activate more than one major-category label.}
\label{tab:appendix_v484_branch_mix}
\setlength{\tabcolsep}{4pt}
\resizebox{\textwidth}{!}{%
\begin{tabular}{llcccc|c}
\toprule
Class & Model & Objective & Variable & Constraint & Implementation & Any Hallucination \\
\midrule
\multirow{9}{*}{\parbox{1.55cm}{\centering General\\models}} & \texttt{DeepSeek-v3.2} & \heatcell{0}{0.4\% (2/484)} & \heatcell{14}{14.3\% (69/484)} & \heatcell{4}{4.1\% (20/484)} & \heatcell{0}{0.0\% (0/484)} & 18.8\% (91/484) \\
& \texttt{Gemini-2.5-flash-lite} & \heatcell{1}{1.0\% (5/484)} & \heatcell{27}{26.7\% (129/484)} & \heatcell{6}{6.0\% (29/484)} & \heatcell{5}{4.5\% (22/484)} & 32.2\% (156/484) \\
& \texttt{GLM-4.5-air} & \heatcell{1}{1.2\% (6/484)} & \heatcell{15}{15.1\% (73/484)} & \heatcell{6}{6.2\% (30/484)} & \heatcell{3}{3.3\% (16/484)} & 22.1\% (107/484) \\
& \texttt{Kimi-K2.5} & \heatcell{0}{0.2\% (1/484)} & \heatcell{10}{9.5\% (46/484)} & \heatcell{5}{4.5\% (22/484)} & \heatcell{1}{1.2\% (6/484)} & 14.9\% (72/484) \\
& \texttt{MiniMax-M2.5} & \heatcell{0}{0.4\% (2/484)} & \heatcell{15}{14.7\% (71/484)} & \heatcell{5}{5.4\% (26/484)} & \heatcell{1}{0.6\% (3/484)} & 20.7\% (100/484) \\
& \texttt{Qwen3-4b} & \heatcell{3}{2.7\% (13/484)} & \heatcell{19}{19.4\% (94/484)} & \heatcell{8}{8.3\% (40/484)} & \heatcell{82}{82.0\% (397/484)} & 86.6\% (419/484) \\
& \texttt{Qwen3-8b} & \heatcell{1}{1.0\% (5/484)} & \heatcell{10}{9.9\% (48/484)} & \heatcell{4}{3.9\% (19/484)} & \heatcell{16}{16.5\% (80/484)} & 27.9\% (135/484) \\
& \texttt{Qwen3-32b} & \heatcell{1}{0.8\% (4/484)} & \heatcell{39}{39.3\% (190/484)} & \heatcell{4}{3.9\% (19/484)} & \heatcell{12}{12.2\% (59/484)} & 46.5\% (225/484) \\
& \texttt{Qwen3.5-Plus} & \heatcell{0}{0.0\% (0/484)} & \heatcell{17}{17.1\% (83/484)} & \heatcell{3}{3.1\% (15/484)} & \heatcell{0}{0.0\% (0/484)} & 20.2\% (98/484) \\
\midrule
\multirow{4}{*}{\parbox{1.75cm}{\centering Fine-tuned\\models}} & \texttt{LLMOPT-Qwen2.5-14B} & \heatcell{1}{1.0\% (5/484)} & \heatcell{22}{21.7\% (105/484)} & \heatcell{7}{7.0\% (34/484)} & \heatcell{12}{11.8\% (57/484)} & 34.1\% (165/484) \\
& \texttt{OptMATH-Qwen2.5-7B} & \heatcell{2}{2.1\% (10/484)} & \heatcell{12}{11.8\% (57/484)} & \heatcell{16}{15.7\% (76/484)} & \heatcell{4}{3.7\% (18/484)} & 28.9\% (140/484) \\
& \texttt{ORLM-LLaMA-3-8B} & \heatcell{7}{7.2\% (35/484)} & \heatcell{19}{18.8\% (91/484)} & \heatcell{34}{33.7\% (163/484)} & \heatcell{2}{2.3\% (11/484)} & 49.6\% (240/484) \\
& \texttt{SIRL-Qwen2.5-7B} & \heatcell{1}{0.8\% (4/484)} & \heatcell{14}{14.5\% (70/484)} & \heatcell{16}{16.1\% (78/484)} & \heatcell{1}{1.2\% (6/484)} & 28.1\% (136/484) \\
\bottomrule
\end{tabular}%
}
\end{table*}

Several patterns are clear from Table~\ref{tab:appendix_v484_branch_mix}. First, for most models, variable and constraint errors are substantially more common than objective errors. This suggests that natural failures more often arise when a model defines the wrong decision variables or translates the problem rules into the wrong constraints than when it simply writes the wrong objective. Second, different models tend to fail in different ways. For \texttt{Qwen3-4b}, the dominant problem is implementation error, which appears in 82.0\% of its artifacts. For \texttt{Qwen3-32b}, variable errors are the most prominent, appearing in 39.3\% of cases. For \texttt{ORLM-LLaMA-3-8B}, constraint errors are the most prominent, appearing in 33.7\% of cases. Third, objective-only errors remain comparatively rare across the panel, reinforcing our claim that objective-value agreement by itself is not the main bottleneck in natural optimization-modeling generation.

Three observations are worth emphasizing. First, the fine-tuned class still shows a wide performance spread on this benchmark rather than a narrow, uniformly strong band. \texttt{SIRL-Qwen2.5-7B} is the strongest of the four on both `Objective-Value Match' (90.3\%) and `Full-Correct' (71.9\%), whereas \texttt{ORLM-LLaMA-3-8B} is the weakest on both metrics (73.1\% and 50.4\%). \texttt{OptMATH-Qwen2.5-7B} and \texttt{LLMOPT-Qwen2.5-14B} lie between these two extremes at 83.1\% and 71.1\%, and 76.2\% and 65.9\%, respectively. Second, objective matching is systematically easier than end-to-end structural correctness: the gap between `Objective-Value Match' and `Full-Correct' is only 0.9 percentage points for \texttt{MiniMax-M2.5}, but it widens to 10.3 for \texttt{LLMOPT-Qwen2.5-14B}, 12.0 for \texttt{OptMATH-Qwen2.5-7B}, 18.4 for \texttt{SIRL-Qwen2.5-7B}, and 22.7 for \texttt{ORLM-LLaMA-3-8B}. Third, even after conditioning on objective-matched outputs, the hidden-error rate remains highly model-dependent: \texttt{MiniMax-M2.5} drops to only 1.0\% hallucination within matches, whereas \texttt{Gemini-2.5-flash-lite}, \texttt{SIRL-Qwen2.5-7B}, \texttt{Qwen3-32b}, and \texttt{ORLM-LLaMA-3-8B} remain much higher at 16.8\%, 20.4\%, 24.5\%, and 31.1\%, respectively. Objective agreement should therefore be treated only as a coarse operational diagnostic rather than as a substitute for OR-expert hallucination labeling.

\subsubsection{Metrics}
\label{app:natural_metrics}
\label{app:metric_definitions}

The natural benchmark is evaluated at two coupled levels. At the artifact level, a detector must decide whether a generated $(P,M,S)$ tuple contains any hallucination at all. At the major-category level, it must identify which of the four major error categories---objective, variable, constraint, and implementation---are present. Because a natural artifact may trigger more than one major category simultaneously, this is a multilabel setting rather than a one-error classification problem. We therefore report one artifact-level F1 score, four major-category-specific F1 scores, and two aggregate major-category F1 scores.

Formally, let $\mathcal{A}=\{1,\dots,N_{\mathrm{nat}}\}$ be the artifact set, and let
\begin{equation}
\mathcal{B}
=
\{\mathrm{obj},\mathrm{var},\mathrm{con},\mathrm{impl}\}
\end{equation}
denote the four major-category channels. For each artifact $i\in\mathcal{A}$ and major category $b\in\mathcal{B}$, let $y_i^{(b)}\in\{0,1\}$ be the OR-expert gold label and let $\hat{y}_i^{(b)}\in\{0,1\}$ be the detector prediction. We then lift these major-category labels to an artifact-level hallucination channel:
\begin{equation}
y_i^{(\mathrm{hall})}
=
\mathbf{1}\!\left[
\operatorname{is\_incorrect}_i
\;\vee\;
\sum_{b\in\mathcal{B}} y_i^{(b)} > 0
\right],
\qquad
\hat{y}_i^{(\mathrm{hall})}
=
\mathbf{1}\!\left[
\sum_{b\in\mathcal{B}} \hat{y}_i^{(b)} > 0
\right].
\end{equation}
Thus $y_i^{(\mathrm{hall})}=1$ whenever OR experts judge artifact $i$ to be incorrect or assign it at least one positive major-category label, and $\hat{y}_i^{(\mathrm{hall})}=1$ whenever the detector predicts at least one major-category hallucination. For any label channel $\ell\in\mathcal{B}\cup\{\mathrm{hall}\}$, define
\begin{align}
\operatorname{TP}_{\ell} &= \sum_{i=1}^{N_{\mathrm{nat}}}\mathbf{1}[y_i^{(\ell)}=1 \land \hat{y}_i^{(\ell)}=1],\\
\operatorname{FP}_{\ell} &= \sum_{i=1}^{N_{\mathrm{nat}}}\mathbf{1}[y_i^{(\ell)}=0 \land \hat{y}_i^{(\ell)}=1],\\
\operatorname{FN}_{\ell} &= \sum_{i=1}^{N_{\mathrm{nat}}}\mathbf{1}[y_i^{(\ell)}=1 \land \hat{y}_i^{(\ell)}=0].
\end{align}
The corresponding F1 score is
\begin{equation}
\operatorname{F1}_{\ell}
=
\frac{2\operatorname{TP}_{\ell}}
{2\operatorname{TP}_{\ell}+\operatorname{FP}_{\ell}+\operatorname{FN}_{\ell}}.
\end{equation}
We report
\begin{equation}
\operatorname{Halluc\mbox{-}F1}
=
\operatorname{F1}_{\mathrm{hall}},
\end{equation}
major-category-specific scores
\begin{align}
\operatorname{Objective\mbox{-}F1}
&=
\operatorname{F1}_{\mathrm{obj}}, \\
\operatorname{Variable\mbox{-}F1}
&=
\operatorname{F1}_{\mathrm{var}}, \\
\operatorname{Constraint\mbox{-}F1}
&=
\operatorname{F1}_{\mathrm{con}}, \\
\operatorname{Implementation\mbox{-}F1}
&=
\operatorname{F1}_{\mathrm{impl}},
\end{align}
and the aggregate metrics
\begin{align}
\operatorname{MajorCategoryMacro\mbox{-}F1}
&=
\frac{1}{4}
\sum_{b\in\mathcal{B}}
\operatorname{F1}_b, \\
\operatorname{MajorCategoryMicro\mbox{-}F1}
&=
\frac{2\sum_{b\in\mathcal{B}}\operatorname{TP}_b}{2\sum_{b\in\mathcal{B}}\operatorname{TP}_b + \sum_{b\in\mathcal{B}}\operatorname{FP}_b + \sum_{b\in\mathcal{B}}\operatorname{FN}_b}.
\end{align}
These metrics have complementary roles. $\operatorname{Halluc\mbox{-}F1}$ asks the coarsest question: can the detector tell whether an artifact is wrong at all, regardless of which major category is responsible? $\operatorname{Objective\mbox{-}F1}$, $\operatorname{Variable\mbox{-}F1}$, $\operatorname{Constraint\mbox{-}F1}$, and $\operatorname{Implementation\mbox{-}F1}$ then resolve that artifact-level judgment into major-category-specific diagnosis quality. $\operatorname{MajorCategoryMacro\mbox{-}F1}$ averages the four major-category F1 scores uniformly and is therefore the better summary when category frequencies are imbalanced, whereas $\operatorname{MajorCategoryMicro\mbox{-}F1}$ pools all major-category decisions before computing a single F1 and therefore weights common categories more heavily.

Every reported metric in this subsection is an F1 score and thus lies in $[0,1]$, with higher values indicating better agreement with the OR-expert natural labels. A value of $1$ means perfect precision and recall on the relevant channel, whereas a value of $0$ means complete failure on that channel. In short, \textit{Halluc-F1} captures artifact-level detection, the four major-category F1 scores capture category-specific diagnosis, \textit{MajorCategoryMacro-F1} rewards balanced performance across the four major categories, and \textit{MajorCategoryMicro-F1} summarizes pooled major-category accuracy under the natural label distribution.

\section{Ablation and Backbone Sensitivity Details}
\label{app:ablation_backbone}

This appendix provides the detailed setup and results for RQ4. All rows use the same three evaluation settings as the main paper: the $484$-case clean benchmark, the $1266$-case controlled injected benchmark, and the $6292$-artifact natural benchmark. The component ablations isolate the contribution of individual OptArgus design choices, while the backbone-sensitivity analysis repeats the matched baseline-versus-OptArgus comparison under an alternative backbone.

\subsection{Component Ablations}
\label{app:component_ablations}

We evaluate four ablations against the full OptArgus configuration. Table~\ref{tab:ablation_variant_design} summarizes exactly which design choice each variant changes. The goal is not only to ask whether OptArgus outperforms a weaker system, but to separate the effects of specialist calibration, selective routing, rescue beyond the first conductor decision, and deterministic final consolidation.

\begin{table*}[h]
\centering
\caption{Design differences among the component ablation variants. A checkmark means the component is retained; a dash means it is removed or bypassed. The comparison isolates whether OptArgus's gains come from calibrated specialist prompts, routed expert selection, rescue beyond conductor-only routing, or final deterministic consolidation.}
\label{tab:ablation_variant_design}
\scriptsize
\setlength{\tabcolsep}{4pt}
{
\renewcommand{\arraystretch}{1.18}
\resizebox{\textwidth}{!}{%
\begin{tabular}{
>{\raggedright\arraybackslash}m{3.8cm}
>{\centering\arraybackslash}m{2.7cm}
>{\centering\arraybackslash}m{2.5cm}
>{\centering\arraybackslash}m{3.0cm}
>{\centering\arraybackslash}m{2.8cm}
>{\raggedright\arraybackslash}m{7.8cm}
}
\toprule
Variant & Specialist calibration & Selective routing & Scout/heuristic rescue & Final rerank/abstain & What this tests \\
\midrule
\textbf{OptArgus} & \checkmark & \checkmark & \checkmark & \checkmark & Reported system: calibrated branch experts are selectively routed, rescue signals recover missed branches, and deterministic consolidation suppresses derivative or low-confidence findings. \\
OptArgus w/o Specialist Calibration & -- & \checkmark & \checkmark & \checkmark & Whether the reported specialist calibration blocks are necessary beyond simply decomposing the task into four branches. \\
OptArgus w/o Routing Rescue & \checkmark & \checkmark & -- & \checkmark & Whether the initial conductor decision alone is sufficient, or whether scout/heuristic rescue is needed for robust routing. \\
OptArgus w/o Final Rerank & \checkmark & \checkmark & \checkmark & -- & Whether branch-local evidence is enough by itself, or whether the final deterministic reranking and abstention layer is needed for calibrated reports. \\
OptArgus w/ All Experts & \checkmark & -- & n/a & \checkmark & Whether exhaustive fan-out can improve coverage when inference cost is ignored; all four specialists run on every case. \\
\bottomrule
\end{tabular}%
}
}
\end{table*}

Table~\ref{tab:ablation_full} supports four observations. First, specialist calibration is doing more than cosmetic prompt wording. Removing it leaves the graph, routing, aggregation, and reranking intact, yet injected Top-1 SpecificTypeHit drops from $0.403$ to $0.228$ and natural Objective-F1 collapses from $0.541$ to $0.020$. This pattern suggests that the specialist prompts are not merely asking four generic reviewers to look at the same artifact; they encode branch-specific decision boundaries, including when to abstain, when to treat a mismatch as a root cause, and when to avoid converting a derivative symptom into a separate hallucination. The lower injected report length under OptArgus w/o Specialist Calibration ($1.136$ vs. $1.224$) is therefore not a useful gain: it comes with much weaker localization.

\begin{table*}[h]
\centering
\caption{Component ablations across the clean, controlled injected, and natural benchmarks. Clean reports EmptyReportRate and MeanFindings$_{\mathrm{clean}}$; injected reports Top-1 localization at major-category, subcategory, and specific-type levels plus MeanFindings; natural reports artifact-level and major-category F1. Parenthesized deltas are relative to OptArgus; blue means better under the metric direction, and red means worse.}
\label{tab:ablation_full}
\scriptsize
\setlength{\tabcolsep}{3pt}
{
\resizebox{\textwidth}{!}{%
\begin{tabular}{lll|llll|lllllll}
\toprule
& \multicolumn{2}{c|}{Clean} & \multicolumn{4}{c|}{Controlled injected} & \multicolumn{7}{c}{Natural} \\
\cmidrule(lr){2-3}\cmidrule(lr){4-7}\cmidrule(lr){8-14}
Method & Empty $\uparrow$ & Mean $\downarrow$ & Major $\uparrow$ & Subcat $\uparrow$ & Specific $\uparrow$ & Mean $\downarrow$ & Halluc $\uparrow$ & Macro $\uparrow$ & Micro $\uparrow$ & Obj $\uparrow$ & Var $\uparrow$ & Con $\uparrow$ & Impl $\uparrow$ \\
\midrule
OptArgus & 0.853 & 0.159 & 0.767 & 0.473 & 0.403 & 1.224 & 0.617 & 0.512 & 0.486 & 0.541 & 0.495 & 0.579 & 0.431 \\
OptArgus w/o Specialist Calibration & 0.791 \regdelta{-0.062} & 0.219 \regdelta{+0.060} & 0.569 \regdelta{-0.198} & 0.320 \regdelta{-0.153} & 0.228 \regdelta{-0.175} & 1.136 \metricdelta{-0.088} & 0.605 \regdelta{-0.012} & 0.373 \regdelta{-0.139} & 0.466 \regdelta{-0.020} & 0.020 \regdelta{-0.521} & 0.474 \regdelta{-0.021} & 0.583 \metricdelta{+0.004} & 0.417 \regdelta{-0.014} \\
OptArgus w/o Routing Rescue & 0.829 \regdelta{-0.024} & 0.182 \regdelta{+0.023} & 0.672 \regdelta{-0.095} & 0.385 \regdelta{-0.088} & 0.327 \regdelta{-0.076} & 1.186 \metricdelta{-0.038} & 0.598 \regdelta{-0.019} & 0.469 \regdelta{-0.043} & 0.449 \regdelta{-0.037} & 0.473 \regdelta{-0.068} & 0.427 \regdelta{-0.068} & 0.562 \regdelta{-0.017} & 0.415 \regdelta{-0.016} \\
OptArgus w/o Final Rerank & 0.636 \regdelta{-0.217} & 0.438 \regdelta{+0.279} & 0.636 \regdelta{-0.131} & 0.395 \regdelta{-0.078} & 0.341 \regdelta{-0.062} & 2.617 \regdelta{+1.393} & 0.591 \regdelta{-0.026} & 0.513 \metricdelta{+0.001} & 0.477 \regdelta{-0.009} & 0.556 \metricdelta{+0.015} & 0.449 \regdelta{-0.046} & 0.613 \metricdelta{+0.034} & 0.435 \metricdelta{+0.004} \\
OptArgus w/ All Experts & 0.880 \metricdelta{+0.027} & 0.122 \metricdelta{-0.037} & 0.788 \metricdelta{+0.021} & 0.479 \metricdelta{+0.006} & 0.416 \metricdelta{+0.013} & 1.184 \metricdelta{-0.040} & 0.617 & 0.499 \regdelta{-0.013} & 0.484 \regdelta{-0.002} & 0.504 \regdelta{-0.037} & 0.501 \metricdelta{+0.006} & 0.546 \regdelta{-0.033} & 0.444 \metricdelta{+0.013} \\
\bottomrule
\end{tabular}%
}
}
\end{table*}

Second, routing needs a recovery mechanism beyond the first conductor plan. OptArgus w/o Routing Rescue preserves specialist calibration and the final reranker, but disables scout/heuristic rescue. Its performance falls on every controlled localization metric: Top-1 MajorCategoryHit decreases from $0.767$ to $0.672$, Top-1 SubcategoryHit from $0.473$ to $0.385$, and Top-1 SpecificTypeHit from $0.403$ to $0.327$. The natural benchmark shows the same pattern, with MajorCategoryMacro-F1 falling from $0.512$ to $0.469$ and MajorCategoryMicro-F1 from $0.486$ to $0.449$. This is exactly the failure mode one would expect in OR artifacts: the initially obvious symptom can be in one branch, while the actual root cause is in another. The routed OptArgus design is stronger because the conductor is allowed to be selective without being irrevocable.

Third, deterministic consolidation is essential for turning specialist evidence into a calibrated detector. OptArgus w/o Final Rerank still runs the calibrated routed experts, but emits the available findings with only simple confidence/severity ordering. This increases clean report length from $0.159$ to $0.438$ and more than doubles controlled injected report length from $1.224$ to $2.617$. It also lowers controlled Top-1 MajorCategoryHit from $0.767$ to $0.636$. Although OptArgus w/o Final Rerank slightly improves a few individual natural branch F1 scores, it does so by allowing more branch-local findings through; the overall artifact-level Halluc-F1 and Micro-F1 both decline. The full system therefore benefits from a deliberately conservative final layer: specialists can be sensitive locally, while the final judge keeps the emitted report concise and root-cause oriented.

Fourth, OptArgus w/ All Experts is best viewed as a high-cost upper-coverage boundary case rather than a strictly better detector. It is unsurprising that running all four calibrated experts on every case improves some clean and controlled metrics: exhaustive fan-out removes routing misses by construction. However, it does not improve the natural aggregate summaries that matter most for realistic outputs: MajorCategoryMacro-F1 falls from $0.512$ to $0.499$, and MajorCategoryMicro-F1 from $0.486$ to $0.484$. This indicates that extra expert coverage can recover a small number of controlled cases, but it also introduces additional noisy branch evidence under naturally overlapping errors. The reported OptArgus configuration is therefore not the maximal-call variant; it is the routed accuracy-cost tradeoff.

We also run an instrumented efficiency audit on the full clean benchmark, where the contrast between routed OptArgus and OptArgus w/ All Experts is easiest to interpret because all inputs are certified correct and the desired behavior is calibrated abstention. In Table~\ref{tab:ablation_efficiency}, \textit{Specialist calls / inst.} counts calls to the objective, variable, constraint, and implementation specialists only, whereas \textit{LLM calls / inst.} counts all LLM calls in the detector graph, including shared routing and consolidation stages. The token columns report provider-measured input, output, and total tokens averaged per instance; token usage was successfully recorded for all evaluated instances in both variants. \textit{LLM sec / inst.} sums traced LLM-call latency per instance; \textit{Wall sec / inst.} is the end-to-end case runtime under the same shard launcher; and \textit{P95 wall sec} is the $95$th percentile of per-case wall time.

\begin{table*}[h]
\centering
\caption{Measured clean-benchmark efficiency of routed OptArgus versus OptArgus w/ All Experts. Both variants use the same clean benchmark, backbone, prompts for each specialist role, and scoring protocol. ``Specialist calls'' counts only objective, variable, constraint, and implementation expert calls; ``LLM calls'' counts all model calls in the graph, including routing and consolidation. Token columns report provider-measured input, output, and total tokens per instance. ``LLM sec'' is traced model-call latency, ``Wall sec'' is end-to-end runtime, and ``P95 wall sec'' is the $95$th percentile of per-instance runtime. Lower is better for calls, tokens, and time.}
\label{tab:ablation_efficiency}
\scriptsize
\setlength{\tabcolsep}{3pt}
{
\resizebox{\textwidth}{!}{%
\begin{tabular}{lllllllll}
\toprule
Method & Specialist calls / inst. $\downarrow$ & LLM calls / inst. $\downarrow$ & Input tok / inst. $\downarrow$ & Output tok / inst. $\downarrow$ & Total tok / inst. $\downarrow$ & LLM sec / inst. $\downarrow$ & Wall sec / inst. $\downarrow$ & P95 wall sec $\downarrow$ \\
\midrule
OptArgus & 1.855 & 5.855 & 42{,}975.0 & 4{,}877.3 & 47{,}852.3 & 62.77 & 57.05 & 113.76 \\
OptArgus w/ All Experts & 4.000 & 8.000 & 66{,}422.5 & 5{,}803.2 & 72{,}225.7 & 77.90 & 58.67 & 115.13 \\
\midrule
Savings of OptArgus vs. w/ All Experts & 53.6\% & 26.8\% & 35.3\% & 16.0\% & 33.7\% & 19.4\% & 2.8\% & 1.2\% \\
\bottomrule
\end{tabular}%
}
}
\end{table*}

Table~\ref{tab:ablation_efficiency} makes the OptArgus w/ All Experts tradeoff explicit with measured token and time traces rather than a proxy. OptArgus w/ All Experts forces all four specialists to run on every case, while routed OptArgus invokes $1.855$ specialists per instance on average. This cuts specialist calls by $53.6\%$ and total LLM calls by $26.8\%$. The token savings are also substantial: input tokens drop from $66{,}422.5$ to $42{,}975.0$ per instance, output tokens drop from $5{,}803.2$ to $4{,}877.3$, and total tokens drop from $72{,}225.7$ to $47{,}852.3$, a $33.7\%$ reduction. The efficiency gain does not come from simply saying less at the expense of calibration: on the clean benchmark, routed OptArgus slightly improves EmptyReportRate ($0.824$ vs. $0.814$) and reduces MeanFindings$_{\mathrm{clean}}$ ($0.186$ vs. $0.196$). Thus OptArgus w/ All Experts remains a useful upper-coverage diagnostic variant, but the reported routed configuration is the more practical operating point for auditing many optimization artifacts.

\subsection{\texorpdfstring{Backbone Sensitivity}{Backbone Sensitivity}}
\label{app:backbone_sensitivity}

The backbone-sensitivity analysis asks whether the same architecture remains useful when the base LLM changes. All main experiments and component ablations use DeepSeek-V3.2 by default; here we repeat the matched comparison between the Single-Agent Detector and OptArgus using Qwen3-Max-Preview as the backbone, keeping the same benchmark cases, taxonomy, output schema, and metrics. Thus the comparison is within the same backbone and does not mix model-capability effects with detector-design effects.

\begin{table*}[h]
\centering
\caption{Backbone sensitivity on the clean and controlled injected benchmarks using Qwen3-Max-Preview. Both methods use the same backbone, inputs, taxonomy, and scoring protocol. Bold marks the best value in each column. Parenthesized deltas are relative to the Baseline row; blue means better under the metric direction, and red means worse.}
\label{tab:backbone_sensitivity_clean_injected}
\scriptsize
\setlength{\tabcolsep}{3pt}
{
\resizebox{\textwidth}{!}{%
\begin{tabular}{lL{2.45cm}L{2.75cm}|L{2.75cm}L{2.60cm}L{2.75cm}L{2.35cm}}
\toprule
& \multicolumn{2}{c|}{\textbf{Panel A. Clean Benchmark (484 cases)}} & \multicolumn{4}{c}{\textbf{Panel B. Controlled Injected Benchmark (1266 cases)}} \\
\cmidrule(lr){2-3}\cmidrule(lr){4-7}
\multicolumn{1}{c}{Method} & \multicolumn{1}{c}{EmptyReportRate $\uparrow$} & \multicolumn{1}{c|}{MeanFindings$_{\mathrm{clean}}$ $\downarrow$} & \multicolumn{1}{c}{Top-1 MajorCategoryHit $\uparrow$} & \multicolumn{1}{c}{Top-1 SubcategoryHit $\uparrow$} & \multicolumn{1}{c}{Top-1 SpecificTypeHit $\uparrow$} & \multicolumn{1}{c}{MeanFindings $\downarrow$} \\
\midrule
Baseline & 0.603 & 0.409 & 0.704 & 0.450 & 0.368 & 1.882 \\
\textbf{OptArgus} & \textbf{0.727} \metricdelta{+0.124} & \textbf{0.285} \metricdelta{-0.124} & \textbf{0.716} \metricdelta{+0.012} & \textbf{0.467} \metricdelta{+0.017} & \textbf{0.369} \metricdelta{+0.001} & \textbf{1.420} \metricdelta{-0.462} \\
\bottomrule
\end{tabular}%
}
}
\end{table*}

\begin{table*}[h]
\centering
\caption{Backbone sensitivity on the natural benchmark using Qwen3-Max-Preview. Both methods use the same backbone, inputs, taxonomy, and scoring protocol. Major-category supports match the main natural benchmark: objective $92$, variable $1126$, constraint $571$, and implementation $675$. Bold marks the best value in each column. Parenthesized deltas are relative to the Baseline row; blue means better under the metric direction, and red means worse.}
\label{tab:backbone_sensitivity_natural}
\scriptsize
\setlength{\tabcolsep}{3pt}
{
\resizebox{\textwidth}{!}{%
\begin{tabular}{llllllll}
\toprule
\multicolumn{1}{c}{\multirow{2}{*}{Method}} & \multicolumn{1}{c}{\multirow{2}{*}{Halluc-F1 $\uparrow$}} & \multicolumn{1}{c}{Objective-F1 $\uparrow$} & \multicolumn{1}{c}{Variable-F1 $\uparrow$} & \multicolumn{1}{c}{Constraint-F1 $\uparrow$} & \multicolumn{1}{c}{Implementation-F1 $\uparrow$} & \multicolumn{1}{c}{MajorCategory} & \multicolumn{1}{c}{MajorCategory} \\
& & ($n{=}92$) & ($n{=}1126$) & ($n{=}571$) & ($n{=}675$) & \multicolumn{1}{c}{Macro-F1 $\uparrow$} & \multicolumn{1}{c}{Micro-F1 $\uparrow$} \\
\midrule
Baseline & 0.495 & 0.578 & 0.300 & 0.519 & 0.251 & 0.412 & 0.365 \\
\textbf{OptArgus} & \textbf{0.598} \metricdelta{+0.103} & \textbf{0.630} \metricdelta{+0.052} & \textbf{0.438} \metricdelta{+0.138} & \textbf{0.525} \metricdelta{+0.006} & \textbf{0.436} \metricdelta{+0.185} & \textbf{0.507} \metricdelta{+0.095} & \textbf{0.460} \metricdelta{+0.095} \\
\bottomrule
\end{tabular}%
}
}
\end{table*}

Tables~\ref{tab:backbone_sensitivity_clean_injected} and~\ref{tab:backbone_sensitivity_natural} show that the positive pattern persists under Qwen3-Max-Preview. OptArgus improves clean abstention from $0.603$ to $0.727$ and reduces clean report length from $0.409$ to $0.285$. On the controlled injected benchmark, it improves Top-1 MajorCategoryHit from $0.704$ to $0.716$, Top-1 SubcategoryHit from $0.450$ to $0.467$, and report length from $1.882$ to $1.420$; the Top-1 SpecificTypeHit gain is very small ($0.368$ to $0.369$), so we interpret this component conservatively. On the natural benchmark, OptArgus improves every reported metric, including Halluc-F1 ($0.495$ to $0.598$), MajorCategoryMacro-F1 ($0.412$ to $0.507$), MajorCategoryMicro-F1 ($0.365$ to $0.460$), and Implementation-F1 ($0.251$ to $0.436$).

\section{Details on Hallucination Taxonomy}
\label{app:taxonomy}
This appendix documents the fine-grained hallucination space underlying $\mathcal{T}^{O}$, $\mathcal{T}^{V}$, $\mathcal{T}^{C}$, and $\mathcal{T}^{I}$. The symbolic major categories retain the detailed OR-oriented structure summarized by our expert-built taxonomy, while the implementation category is aligned with the detector branch that compares the symbolic model against the realized solver program through parsing and execution checks. In other words, the implementation subsection below is not merely conceptual; it matches the actual subtype inventory used by the current system.

\begin{table}[!htbp]]
\caption{Summary of the hallucination taxonomy on optimization modeling.}
\label{tab:hallucination-taxonomy-summary}
\begin{center}
\setlength{\tabcolsep}{4pt}
\renewcommand{\arraystretch}{1.06}
\resizebox{1\linewidth}{!}{%
\begin{tabular}{
>{\centering\arraybackslash}m{6.2cm}
>{\centering\arraybackslash}m{7.0cm}
>{\raggedright\arraybackslash}p{11.2cm}
}
\toprule
\textbf{Major Category} & \textbf{Subcategory} & \textbf{Specific Hallucination Type} \\
\midrule

\multirow{18}{6.2cm}{\centering \hyperref[app:subsec:obj_mod_hal]{Objective-Function Modeling Hallucinations}}
& \multirow{4}{7.0cm}{\centering \hyperref[app:subsubsec:obj_sem_map_err]{Objective Semantic Mapping Errors}}
& \hyperref[app:subsubsubsec:wro_opt_dir]{1.1.1 Wrong Optimization Direction} \\
& & \hyperref[app:subsubsubsec:obj_sub]{1.1.2 Objective Substitution} \\
& & \hyperref[app:subsubsubsec:sof_har_con]{1.1.3 Softening a Hard Constraint} \\
& & \hyperref[app:subsubsubsec:tur_per_into_har_con]{1.1.4 Turning a Preference into a Hard Constraint} \\
\cmidrule(lr){2-3}

& \multirow{4}{7.0cm}{\centering \hyperref[app:subsubsec:obj_com_err]{Objective Composition Errors}}
& \hyperref[app:subsubsubsec:omi_key_obj_ter]{1.2.1 Omission of a Key Objective Term} \\
& & \hyperref[app:subsubsubsec:int_spu_obj_ter]{1.2.2 Introduction of a Spurious Objective Term} \\
& & \hyperref[app:subsubsubsec:rep_cou_obj_ter]{1.2.3 Repeated Counting of an Objective Term}\\
& & \hyperref[app:subsubsubsec:wro_sig_loc_obj_ter]{1.2.4 Wrong Sign on a Local Objective Term} \\
\cmidrule(lr){2-3}

& \multirow{4}{7.0cm}{\centering \hyperref[app:subsubsec:obj_par_ind_bin_err]{Objective Parameter and Index Binding Errors}}
& \hyperref[app:subsubsubsec:coe_mis]{1.3.1 Coefficient Misbinding} \\
& & \hyperref[app:subsubsubsec:wro_ind_ran]{1.3.2 Wrong Index Range} \\
& & \hyperref[app:subsubsubsec:mis_tim_ind]{1.3.3 Misaligned Time Indices} \\
& & \hyperref[app:subsubsubsec:wro_uni_sca_bin]{1.3.4 Wrong Unit or Scale Binding} \\
\cmidrule(lr){2-3}

& \multirow{3}{7.0cm}{\centering \hyperref[app:subsubsec:mul_obj_ris_obj_err]{Multi-Objective and Risk-Objective Errors}}
& \hyperref[app:subsubsubsec:wro_mul_obj_agg_mec]{1.4.1 Wrong Multi-Objective Aggregation Mechanism} \\
& & \hyperref[app:subsubsubsec:rep_ris_ext_val_obj]{1.4.2 Replacing a Risk or Extreme-Value Objective} \\
& & \hyperref[app:subsubsubsec:wro_wei_set_int]{1.4.3 Wrong Weight Setting or Interpretation} \\
\cmidrule(lr){2-3}

& \multirow{3}{7.0cm}{\centering \hyperref[app:subsubsec:obj_for_err]{Objective Formalization Errors}}
& \hyperref[app:subsubsubsec:smo_pie_ste]{1.5.1 Smoothing Piecewise or Stepwise Cost into a Continuous Linear Cost} \\
& & \hyperref[app:subsubsubsec:inc_han_non]{1.5.2 Incorrect Handling of Nonlinear or Nonsmooth Objective Terms} \\
& & \hyperref[app:subsubsubsec:fra_obj_err]{1.5.3 Fractional Objective Errors} \\
\midrule

\multirow{18}{6.2cm}{\centering \hyperref[app:subsec:dec_var_hal]{Decision-Variable Modeling Hallucinations}}
& \multirow{4}{7.0cm}{\centering \hyperref[app:subsubsec:var_sem_def_err]{Variable Semantic Definition Errors}}
& \hyperref[app:subsubsubsec:wro_var_obj]{2.1.1 Wrong Variable Object} \\
& & \hyperref[app:subsubsubsec:mis_key_dim]{2.1.2 Missing a Key Dimension} \\
& & \hyperref[app:subsubsubsec:wro_agg_lev]{2.1.3 Wrong Aggregation Level of Variables} \\
& & \hyperref[app:subsubsubsec:sta_con_con]{2.1.4 State-Control Confusion} \\
\cmidrule(lr){2-3}

& \multirow{4}{7.0cm}{\centering \hyperref[app:subsubsec:var_dom_typ_err]{Variable Domain and Type Errors}}
& \hyperref[app:subsubsubsec:rel_dis_var]{2.2.1 Relaxing a Discrete Variable into a Continuous Variable} \\
& & \hyperref[app:subsubsubsec:for_con_var]{2.2.2 Forcing a Continuous Variable to Be Discrete} \\
& & \hyperref[app:subsubsubsec:wro_val_ran]{2.2.3 Wrong Value Range} \\
& & \hyperref[app:subsubsubsec:wro_sig_dom]{2.2.4 Wrong Sign Domain} \\
\cmidrule(lr){2-3}

& \multirow{4}{7.0cm}{\centering \hyperref[app:subsubsec:var_ind_set_bin_err]{Variable Index and Set Binding Errors}}
& \hyperref[app:subsubsubsec:omi_ind_set]{2.3.1 Omitted Index Set} \\
& & \hyperref[app:subsubsubsec:fab_ind_set]{2.3.2 Fabricated Index Set} \\
& & \hyperref[app:subsubsubsec:wro_sub_bin]{2.3.3 Wrong Subscript Binding} \\
& & \hyperref[app:subsubsubsec:mis_ind_dep]{2.3.4 Missing Index Dependence} \\
\cmidrule(lr){2-3}

& \multirow{3}{7.0cm}{\centering \hyperref[app:subsubsec:var_rol_cou_err]{Variable Role Coupling Errors}}
& \hyperref[app:subsubsubsec:var_not_tak]{2.4.1 Variable Not Taking Effect} \\
& & \hyperref[app:subsubsubsec:dup_var_rol]{2.4.2 Duplicate Variable Roles} \\
& & \hyperref[app:subsubsubsec:mis_cou_bet]{2.4.3 Missing Coupling Between Master and Auxiliary Variables} \\
\cmidrule(lr){2-3}

& \multirow{3}{7.0cm}{\centering \hyperref[app:subsubsec:var_red_unc_sym]{Variable Redundancy and Uncontrolled Symmetry}}
& \hyperref[app:subsubsubsec:unn_red_var]{2.5.1 Unnecessary Redundant Variables} \\
& & \hyperref[app:subsubsubsec:unh_sym_var]{2.5.2 Unhandled Symmetric Variables} \\
& & \hyperref[app:subsubsubsec:fai_exp_spa]{2.5.3 Failure to Exploit Sparsity} \\
\midrule

\multirow{31}{6.2cm}{\centering \hyperref[app:subsec:con_mod_hal]{Constraint Modeling Hallucinations}}
& \multirow{4}{7.0cm}{\centering \hyperref[app:subsubsec:con_sem_tra_err]{Constraint Semantic Translation Errors}}
& \hyperref[app:subsubsubsec:wro_int_rul]{3.1.1 Wrong Interpretation of the Rule} \\
& & \hyperref[app:subsubsubsec:wro_int_qua]{3.1.2 Wrong Interpretation of Quantifiers} \\
& & \hyperref[app:subsubsubsec:con_res_amo]{3.1.3 Confusing Resource Amount with Time Feasibility} \\
& & \hyperref[app:subsubsubsec:con_typ_mis]{3.1.4 Constraint Type Mismatch} \\
\cmidrule(lr){2-3}

& \multirow{4}{7.0cm}{\centering \hyperref[app:subsubsec:mis_con_ske]{Missing Constraint Skeleton}}
& \hyperref[app:subsubsubsec:mis_bal_con]{3.2.1 Missing Balance, Conservation, or Recursion Constraints} \\
& & \hyperref[app:subsubsubsec:mis_cha_pat]{3.2.2 Missing Chain or Path Structure Constraints} \\
& & \hyperref[app:subsubsubsec:mis_cap_cov]{3.2.3 Missing Capacity or Coverage Skeleton} \\
& & \hyperref[app:subsubsubsec:mis_ini_ter]{3.2.4 Missing Initial or Terminal Conditions} \\
\cmidrule(lr){2-3}

& \multirow{3}{7.0cm}{\centering \hyperref[app:subsubsec:mis_imp_con]{Missing Implicit Constraints}}
& \hyperref[app:subsubsubsec:mis_def_bus]{3.3.1 Missing Default Business Rules} \\
& & \hyperref[app:subsubsubsec:mis_mus_inc]{3.3.2 Missing Must-Include Object Constraints} \\
& & \hyperref[app:subsubsubsec:mis_leg_act]{3.3.3 Missing Legal-Action Set Restrictions} \\
\cmidrule(lr){2-3}

& \multirow{3}{7.0cm}{\centering \hyperref[app:subsubsec:err_equ_sub]{Erroneous Equivalent Substitutions}}
& \hyperref[app:subsubsubsec:rep_equ_ine]{3.4.1 Replacing an Equality with an Inequality} \\
& & \hyperref[app:subsubsubsec:usi_loc_par]{3.4.2 Using Local or Partial Conditions to Replace a Full Structure} \\
& & \hyperref[app:subsubsubsec:tre_app_con]{3.4.3 Treating an Approximate Constraint as an Exact One} \\
\cmidrule(lr){2-3}

& \multirow{4}{7.0cm}{\centering \hyperref[app:subsubsec:log_con_cha_err]{Logic and Conditional-Chain Errors}}
& \hyperref[app:subsubsubsec:big_dir_err]{3.5.1 Big-$M$ Direction Error} \\
& & \hyperref[app:subsubsubsec:big_too_loo]{3.5.2 Big-$M$ Too Loose or Too Tight} \\
& & \hyperref[app:subsubsubsec:inc_log_lin]{3.5.3 Incomplete Logical Linearization} \\
& & \hyperref[app:subsubsubsec:bro_con_cha]{3.5.4 Broken Conditional Chains} \\
\cmidrule(lr){2-3}

& \multirow{4}{7.0cm}{\centering \hyperref[app:subsubsec:agg_ind_cod_err]{Aggregation and Index-Coding Errors}}
& \hyperref[app:subsubsubsec:wro_agg_lev]{3.6.1 Wrong Aggregation Level} \\
& & \hyperref[app:subsubsubsec:wro_sum_ran]{3.6.2 Wrong Summation Range} \\
& & \hyperref[app:subsubsubsec:wro_sub_whe]{3.6.3 Wrong Subscripts When Calling a Constraint} \\
& & \hyperref[app:subsubsubsec:rep_mis_agg]{3.6.4 Repeated or Missing Aggregation} \\
\cmidrule(lr){2-3}

& \multirow{4}{7.0cm}{\centering \hyperref[app:subsubsec:con_bou_dir_err]{Constraint Boundary and Direction Errors}}
& \hyperref[app:subsubsubsec:wro_ine_dir]{3.7.1 Wrong Inequality Direction} \\
& & \hyperref[app:subsubsubsec:swa_upp_low]{3.7.2 Swapped Upper and Lower Bounds} \\
& & \hyperref[app:subsubsubsec:mis_bou_con]{3.7.3 Misplaced Boundary Conditions} \\
& & \hyperref[app:subsubsubsec:mis_thr_bou]{3.7.4 Misuse of Threshold or Boundary Parameters} \\
\cmidrule(lr){2-3}

& \multirow{3}{7.0cm}{\centering \hyperref[app:subsubsec:con_str_imb]{Constraint Strength Imbalance}}
& \hyperref[app:subsubsubsec:con_too_wea]{3.8.1 Constraint Too Weak} \\
& & \hyperref[app:subsubsubsec:con_too_str]{3.8.2 Constraint Too Strong} \\
& & \hyperref[app:subsubsubsec:sla_tol_par]{3.8.3 Slack or Tolerance Parameters on the Wrong Scale} \\
\cmidrule(lr){2-3}

& \multirow{2}{7.0cm}{\centering \hyperref[app:subsubsec:sch_act_str_spe_err]{Scheduling- and Activity-Structure-Specific Errors}}
& \hyperref[app:subsubsubsec:wro_enc_pre]{3.9.1 Wrong Encoding of Precedence Relations} \\
& & \hyperref[app:subsubsubsec:wro_str_mod]{3.9.2 Wrong Structure for Mode Selection and Optional Activities} \\
\midrule

\multirow{16}{6.2cm}{\centering \hyperref[app:subsec:imp_hal]{Implementation Hallucinations}}
& \multirow{4}{7.0cm}{\centering \hyperref[app:subsubsec:imp_sym_code]{Symbolic-Code Mismatch}}
& \hyperref[app:subsubsubsec:imp_wro_obj_sense]{4.1.1 Wrong Objective Sense in Code} \\
& & \hyperref[app:subsubsubsec:imp_omi_con_mat]{4.1.2 Omitted Constraint Materialization} \\
& & \hyperref[app:subsubsubsec:imp_mis_var_reg]{4.1.3 Missing Variable Registration} \\
& & \hyperref[app:subsubsubsec:imp_sta_div_coeff]{4.1.4 Stale or Divergent Coefficient Mapping} \\
\cmidrule(lr){2-3}

& \multirow{3}{7.0cm}{\centering \hyperref[app:subsubsec:imp_var_api]{Variable and Domain API Mismatch}}
& \hyperref[app:subsubsubsec:imp_wro_api_var_type]{4.2.1 Wrong API Variable Type} \\
& & \hyperref[app:subsubsubsec:imp_wro_bounds_code]{4.2.2 Wrong Bounds in Code} \\
& & \hyperref[app:subsubsubsec:imp_wro_index_domain]{4.2.3 Wrong Index Domain Materialization} \\
\cmidrule(lr){2-3}

& \multirow{3}{7.0cm}{\centering \hyperref[app:subsubsec:imp_idx_set]{Index and Set Materialization Errors}}
& \hyperref[app:subsubsubsec:imp_drop_loop]{4.3.1 Dropped Loop or Index Expansion} \\
& & \hyperref[app:subsubsubsec:imp_partial_set]{4.3.2 Partial Set Expansion} \\
& & \hyperref[app:subsubsubsec:imp_filtered_index_loss]{4.3.3 Filtered Index Loss} \\
\cmidrule(lr){2-3}

& \multirow{3}{7.0cm}{\centering \hyperref[app:subsubsec:imp_solver_form]{Solver and Formalization Mismatch}}
& \hyperref[app:subsubsubsec:imp_missing_linearization]{4.4.1 Missing Linearization in Code} \\
& & \hyperref[app:subsubsubsec:imp_incompatible_solver]{4.4.2 Incompatible Solver Selection} \\
& & \hyperref[app:subsubsubsec:imp_incorrect_bigm]{4.4.3 Incorrect Big-$M$ or Numerical Option in Code} \\
\cmidrule(lr){2-3}

& \multirow{3}{7.0cm}{\centering \hyperref[app:subsubsec:imp_postsolve]{Post-Solve Extraction and Reporting Divergence}}
& \hyperref[app:subsubsubsec:imp_wrong_readout]{4.5.1 Wrong Solution Variable Readout} \\
& & \hyperref[app:subsubsubsec:imp_misreported_obj]{4.5.2 Misreported Objective Value} \\
& & \hyperref[app:subsubsubsec:imp_stale_result]{4.5.3 Stale Result Object or Wrong Post-Processing} \\
\bottomrule
\end{tabular}%
}
\end{center}
\end{table}

To support a systematic analysis of LLM hallucinations in automated optimization modeling, we develop a fine-grained taxonomy of hallucinations in this section. Rather than treating hallucination as a single undifferentiated phenomenon, we organize it according to the stage of model construction at which the semantic distortion occurs, including objective-function modeling, decision-variable modeling, constraint modeling, and code-level implementation. As summarized in Table~\ref{tab:hallucination-taxonomy-summary}, each major category is further divided into subcategories and specific hallucination types in order to distinguish hallucinations in semantic translation, formalization, indexing, coupling, aggregation, structural encoding, and symbolic-to-code realization. Building on this taxonomy, the remainder of this section examines these major categories in turn, clarifies their boundaries relative to neighboring categories, and explains how a model may remain formally solvable while still being semantically misaligned with the original optimization task.

\subsection{Objective-Function Modeling Hallucinations}
\label{app:subsec:obj_mod_hal}

\subsubsection{Objective Semantic Mapping Errors}
\label{app:subsubsec:obj_sem_map_err}

\textbf{Definition.} This category concerns failures to map the intent of a natural-language task---what should be optimized, in which direction it should be optimized, and which requirements are mandatory versus merely preferable---into the correct objective function and objective role in a mathematical program. The result is an objective whose notion of optimality no longer matches the original decision intent.

\textbf{Boundary.} The focus here is the semantic role of the objective: whether the model understood what should be optimized and what should remain a hard requirement. If the error lies in omitted or redundant cost terms, repeated counting, or a local sign flip inside the objective, it belongs to Sec.~\ref{app:subsubsec:obj_com_err}. If the issue is the misbinding of coefficients, time indices, object indices, or units, it belongs to Sec.~\ref{app:subsubsec:obj_par_ind_bin_err}. If the error concerns multi-objective aggregation, risk preference, or the interpretation of weights, it belongs to Sec.~\ref{app:subsubsec:mul_obj_ris_obj_err}. If the mathematical realization of a piecewise, extreme-value, absolute-value, bilinear, or fractional objective is wrong, it belongs to Sec.~\ref{app:subsubsec:obj_for_err}. In short, Sec.~\ref{app:subsubsec:obj_sem_map_err}. asks whether the semantic target of optimization is correct, not whether the target was expanded, indexed, or encoded correctly.

\subsubsubsection{1.1.1 Wrong Optimization Direction}
\label{app:subsubsubsec:wro_opt_dir}

\textbf{Definition.} The original problem asks to minimize or maximize a performance metric, but the generated model reverses that direction, so the solver systematically searches in the opposite direction of the real decision goal.

\textbf{Typical erroneous form.} The intended objective is
$
\min f(x),
$
but the model is written as
$
\max f(x).
$

\textbf{Example.} In a regional power-purchasing plan, total purchasing cost should be minimized subject to load balance and reserve constraints:
$
\min \sum_t \sum_g c_{gt} p_{gt},
$
where $p_{gt}$ is the purchased power or dispatch output of unit $g$ in period $t$, and $c_{gt}$ is the corresponding electricity price or marginal cost. If the code is mistakenly written as
$
\max \sum_t \sum_g c_{gt} p_{gt},
$
the solver will prefer expensive generators first. All constraints may still be satisfied formally, yet the optimization direction is exactly opposite to the true dispatch goal.

\textbf{Consequence.} This error flips the criterion of optimality itself and makes the solver prefer ``worse is better.'' It does not create a small local deviation; it distorts the global direction of the decision. Cost-minimization becomes cost amplification, profit-maximization becomes profit suppression, and the computed solution may be mathematically optimal while being close to worst possible from a business standpoint.

\subsubsubsection{1.1.2 Objective Substitution}
\label{app:subsubsubsec:obj_sub}

\textbf{Definition.} The core performance measure that should be optimized in the original problem is replaced by another indicator that is related to it but not equivalent. The model remains solvable, but the preference structure encoded by the objective is no longer the original one.

\textbf{Typical erroneous form.} The intended objective is
$
\min f(x),
$
but it is replaced with
$
\min g(x),
$
where $g(x)$ may be numerically correlated with $f(x)$ while not being equivalent in decision meaning. A common example is replacing
$
\min \max_k L_k
$
with
$
\min \sum_k L_k.
$

\textbf{Example.} In cold-chain routing, one may wish to minimize the maximum route length across vehicles in order to control the worst route and reduce overtime or temperature-control risk:
$
\min \max_k L_k,
$
where $L_k$ is the total travel distance of vehicle $k$. If the code instead uses
$
\min \sum_k L_k,
$
it optimizes total distance rather than the longest route. Both metrics involve ``distance,'' but the preference logic changes: the former targets balance and worst-case control, whereas the latter targets aggregate efficiency.

\textbf{Consequence.} Because the model remains solvable, the error is quite hidden. The solver seems to succeed, yet it is optimizing the wrong business KPI. Aggregate performance may improve while the truly relevant risk, fairness, robustness, or service-quality criterion actually deteriorates.

\subsubsubsection{1.1.3 Softening a Hard Constraint}
\label{app:subsubsubsec:sof_har_con}

\textbf{Definition.} A rigid condition that defines the feasible region is incorrectly rewritten as a penalty term in the objective. A ``must satisfy'' rule is thus downgraded to a ``may violate if you pay enough'' preference. If the penalty coefficient is not sufficiently large, the model may actively choose to violate a business rule or safety boundary that should never be violated.

\textbf{Typical erroneous form.} A hard constraint such as
$
 h(x) \le 0
$
is moved into the objective as
$
\min f(x) + \lambda [h(x)]_+,
$
where
$
[h(x)]_+ = \max\{h(x),0\}
$
and $\lambda$ is a penalty coefficient. If $\lambda$ is too small, the model will accept violations of the original hard condition.

\textbf{Example.} In a chemical production plan, the inventory of an intermediate tank must never fall below a safety stock line $S^{\min}$; that is,
$
I_t \ge S^{\min}, \quad \forall t.
$
This is a safety condition and should be modeled as a hard constraint. If the code omits it and instead adds an inventory-shortfall penalty,
$
\min \text{Cost} + \lambda \sum_t \max\{0,S^{\min}-I_t\},
$
then, for an insufficiently large $\lambda$, the model may deliberately allow the tank inventory to drop below the safety line in some periods to save production or procurement cost.

\textbf{Consequence.} The feasible region is redefined incorrectly. Solutions that were originally unacceptable are now admitted into the candidate set, so the solver returns a low-cost but noncompliant plan. In industrial settings, this is often among the most dangerous errors because it can introduce safety risk, regulatory violations, or physically infeasible operations.

\subsubsubsection{1.1.4 Turning a Preference into a Hard Constraint}
\label{app:subsubsubsec:tur_per_into_har_con}

\textbf{Definition.} A soft metric meant only to express preference, balance, or improvement direction is incorrectly elevated into a mandatory constraint. This over-shrinks the feasible region and distorts the trade-off between the main objective and secondary preferences.

\textbf{Typical erroneous form.} The intended model controls a deviation through an objective penalty such as
$
\min f(x)+\lambda\,\mathrm{Dev}(x),
$
but it is incorrectly replaced by a hard requirement such as
$
\mathrm{Dev}(x)=0
$
or
$
\mathrm{Dev}(x) \le \varepsilon,
$
with $\varepsilon$ set too small, sometimes even to $0$.

\textbf{Example.} In manufacturing scheduling, the real requirement may be to satisfy orders first and then balance workloads across production lines as much as possible. A reasonable formulation penalizes workload deviation in the objective, for example through a term involving the line loads $u_m$ and their average $\bar u$. If the code instead imposes
$
u_1=u_2=\cdots=u_M,
$
then ``balance as much as possible'' is mistranslated into ``must be exactly equal,'' turning a soft preference into a rigid requirement.

\textbf{Consequence.} The feasible region is unnecessarily reduced. In mild cases the model sacrifices main-objective performance; in severe cases it becomes infeasible. Even when a solution still exists, it may damage delivery, cost, or utilization substantially just to satisfy a secondary preference that should have been treated softly.

\textbf{Summary.} Objective semantic mapping errors fall into two broad groups. One group distorts the internal meaning of the objective itself, as in wrong direction and objective substitution. The other group misplaces the role of objective versus constraints, as in softening hard constraints or hardening soft preferences. The former changes what is optimized and in which direction; the latter changes which conditions define feasibility and which indicators merely express preference. These errors often do not prevent the model from solving, but they make the solver return a formally solvable and semantically distorted plan.

\subsubsection{Objective Composition Errors}
\label{app:subsubsec:obj_com_err}

\textbf{Definition.} The model preserves the overall optimization direction, yet the objective fails to include the right collection of costs, rewards, penalties, or bonuses at the term level. Typical manifestations are omitted terms, irrelevant extra terms, repeated counting, or locally reversed signs. The result is a distorted ranking among candidate decisions.

\textbf{Boundary.} This section asks which terms should appear in the objective and whether they are included correctly. If the error concerns the overall direction of optimization, the role of objective versus constraints, or parameter-variable binding, it belongs to Secs.~\ref{app:subsubsec:obj_sem_map_err} or~\ref{app:subsubsec:obj_par_ind_bin_err} instead.

\subsubsubsection{1.2.1 Omission of a Key Objective Term}
\label{app:subsubsubsec:omi_key_obj_ter}

\textbf{Definition.} A cost, reward, or penalty term that should appear in the objective is dropped. A certain behavior therefore becomes effectively free or unrewarded, and the model is systematically biased toward unrealistic ``optimal'' solutions.

\textbf{Typical erroneous form.} The intended objective contains two or more components, such as
$
\min \{f_1(x)+f_2(x)\},
$
but the implemented model only uses
$
\min f_1(x),
$
thereby omitting the key term $f_2(x)$.

\textbf{Example.} In workforce planning, hiring cost, firing cost, and wage cost may all matter:
$
\min \sum_t \bigl(c_t^h H_t + c_t^f F_t + c_t^w W_t\bigr).
$
Here $H_t$, $F_t$, and $W_t$ denote hires, layoffs, and workforce level in period $t$. If the firing-cost term is omitted, the model treats layoffs as free and may repeatedly fire and rehire workers to regulate capacity.

\textbf{Consequence.} The omitted behavior is systematically underpriced, so the solver overuses it. Even with correct constraints, the ``optimal'' plan may show unreasonable over-firing, excessive switching, frequent start-stop cycles, or disregard for tardiness. The deeper problem is that the objective no longer evaluates solutions according to the true economic criterion.

\subsubsubsection{1.2.2 Introduction of a Spurious Objective Term}
\label{app:subsubsubsec:int_spu_obj_ter}

\textbf{Definition.} The model introduces an extra cost, reward, or penalty term that is not present in the problem statement or business rules. The model then optimizes a criterion that the original problem never asked to control, thereby changing the nature of the decision problem itself.

\textbf{Typical erroneous form.} The intended objective is
$
\min f(x),
$
but the code uses
$
\min f(x)+g(x),
$
where $g(x)$ is not part of the original evaluation criterion.

\textbf{Example.} In a distribution-network design problem, suppose the statement includes only transportation cost and vehicle-usage cost:
$
\min \sum_{(i,j)} c_{ij}x_{ij} + \sum_k d_k y_k,
$
where $x_{ij}$ is the transported quantity on arc $(i,j)$ and $y_k$ indicates whether vehicle $k$ is used. If the code adds an extra term
$
\sum_{(i,j)} f_{ij} z_{ij},
$
with $z_{ij}$ interpreted as arc activation and $f_{ij}$ as a fixed route-opening cost, the problem is no longer a plain transport-dispatch problem but a different network-design problem.

\textbf{Consequence.} The solver no longer solves the user's task but an altered task decorated by an extra preference. The resulting plan may over-sparsify the network, deliberately reduce the number of used paths, or compress resource activation only because the model introduced a fictitious cost.

\subsubsubsection{1.2.3 Repeated Counting of an Objective Term}
\label{app:subsubsubsec:rep_cou_obj_ter}

\textbf{Definition.} The same cost, reward, or penalty term is counted more than once because of a mistaken index structure, summation range, or expression form. The weight of that term is unintentionally amplified.

\textbf{Typical erroneous form.} A cost that should be written as
$
\min \sum_t c_t I_t
$
is incorrectly written as
$
\min \sum_p \sum_t c_t I_t,
$
even though $I_t$ does not depend on the product index $p$. The same term is then counted $|P|$ times.

\textbf{Example.} Total holding cost in a warehouse should be
$
\min \sum_t h_t I_t,
$
where $I_t$ is total inventory in period $t$ and $h_t$ is the unit holding cost. If the code writes
$
\min \sum_{p\in P} \sum_t h_t I_t,
$
and $I_t$ is aggregate inventory rather than product-specific inventory $I_{pt}$, the holding cost of the same inventory stock is counted repeatedly.

\textbf{Consequence.} The model becomes overly sensitive to the duplicated term. Even if all other terms are present, the solver may over-compress inventory, over-suppress delay, or over-penalize switching simply because one objective component was multiplied accidentally by the index structure.

\subsubsubsection{1.2.4 Wrong Sign on a Local Objective Term}
\label{app:subsubsubsec:wro_sig_loc_obj_ter}

\textbf{Definition.} The overall objective direction is correct, but the sign of a local cost, reward, bonus, or penalty term is reversed. The local incentive generated by that term therefore points in the opposite direction from what was intended.

\textbf{Typical erroneous form.} In a minimization problem, a cost term that should enter as $+cx$ is written as $-cx$; in a maximization problem, a revenue term that should enter as $+rx$ is written as $-rx$. This differs from changing the global $\min$ into $\max$: only one part of the objective is reversed.

\textbf{Example.} In an e-commerce warehousing system, shortage is supposed to incur a penalty:
$
\min \sum_t \bigl(c_t^{\text{ship}}x_t + c_t^{\text{hold}}I_t + c_t^{\text{short}} s_t\bigr),
$
where $s_t$ is shortage in period $t$ and $c_t^{\text{short}}>0$ is the unit shortage penalty. If the code writes the shortage term with a negative sign, the minimization problem begins to reward shortage rather than penalize it.

\textbf{Consequence.} The model may actively seek behaviors that should be avoided, such as larger shortages, more emissions, more lateness, or more violations, or suppress behaviors that should be encouraged. This is often harder to detect than an omitted term because the term is present, but it pushes the decision in exactly the wrong direction.

\textbf{Summary.} Objective composition errors appear as missing terms, spurious terms, repeated counting, or local sign reversal. Unlike Sec.~\ref{app:subsubsec:obj_sem_map_err} , the issue is not whether the overall optimization intent was understood, but whether the internal set of objective components was included fully and correctly. Such errors distort the relative ranking of candidate plans and make the solution reflect modeling distortion rather than true business preference.

\subsubsection{Objective Parameter and Index Binding Errors}
\label{app:subsubsec:obj_par_ind_bin_err}

\textbf{Definition.} Cost, reward, and penalty parameters do appear in the model, but they are bound to the wrong variables, indices, time points, or units. The relevant objective term exists only in appearance; in effect it acts on the wrong decision object.

\textbf{Boundary.} This section concerns cases in which an objective term is present but attached to the wrong target. If a term is omitted, added redundantly, counted repeatedly, or locally sign-reversed, the issue belongs to Sec.~\ref{app:subsubsec:obj_com_err}. If the problem is with the overall optimization direction, objective substitution, or objective-constraint role confusion, it belongs to Sec.~\ref{app:subsubsec:obj_sem_map_err}.

\subsubsubsection{1.3.1 Coefficient Misbinding}
\label{app:subsubsubsec:coe_mis}

\textbf{Definition.} A cost, reward, or penalty coefficient is multiplied by the wrong variable, so the economic meaning associated with one decision object is mistakenly transferred to another.

\textbf{Typical erroneous form.} The intended expression is
$
\min \sum_i c_i x_i,
$
but the code uses
$
\min \sum_i c_i y_i,
$
even though the business meaning of $c_i$ corresponds to $x_i$, not to $y_i$.

\textbf{Example.} In urban distribution, vehicle activation cost should be charged per activated vehicle. Let $y_k\in\{0,1\}$ indicate whether vehicle $k$ is used, $F_k$ its fixed activation cost, and $x_{ijk}$ whether vehicle $k$ traverses arc $(i,j)$. The fixed-cost term should be
$
\sum_k F_k y_k.
$
If the code instead charges the fixed cost through a sum over the traversed arcs of each vehicle, the model effectively turns a per-vehicle cost into a per-arc cost.

\textbf{Consequence.} The marginal-cost structure of the decision is changed. The model no longer compares ``whether to activate one vehicle'' but instead compares ``whether taking another arc will trigger a fixed cost that should not exist.'' This can produce overly compressed fleet sizes, unnaturally short routes, or other preference patterns with no business justification.

\subsubsubsection{1.3.2 Wrong Index Range}
\label{app:subsubsubsec:wro_ind_ran}

\textbf{Definition.} The set or subset over which the objective is summed is specified incorrectly. Some objects that should be included are omitted, or objects that should not be included are added.

\textbf{Typical erroneous form.} The intended expression is
$
\min \sum_{i\in I} c_i x_i,
$
but the code uses
$
\min \sum_{i\in I'} c_i x_i,
$
where $I'$ is a wrong subset, superset, or otherwise incorrect object set.

\textbf{Example.} In airline-network recovery, one may wish to minimize total delay cost over all flights:
$
\min \sum_{f\in F} c_f d_f.
$
If the code sums only over the subset of base-departure flights $F^{\text{base}}\subset F$, then all non-base flights become zero-weight objects in the objective.

\textbf{Consequence.} Objects outside the mistaken range still exist in the constraints but are no longer actively optimized. The solver then focuses on some flights, customers, or orders while systematically ignoring the rest.

\subsubsubsection{1.3.3 Misaligned Time Indices}
\label{app:subsubsubsec:mis_tim_ind}

\textbf{Definition.} Time-dependent costs or rewards are not aligned with the correct period. A price, holding cost, penalty, or reward from one period is attached to the decision variable of a previous or later period.

\textbf{Typical erroneous form.} The intended objective is
$
\min \sum_t c_t x_t,
$
but it is written as
$
\min \sum_t c_t x_{t+1}
$
or
$
\min \sum_t c_{t+1} x_t.
$

\textbf{Example.} In a multi-period inventory model, end-of-period inventory $I_t$ should incur period-$t$ holding cost $h_t$:
$
\min \sum_{t=1}^T h_t I_t.
$
If the code writes
$
\min \sum_{t=1}^{T-1} h_t I_{t+1},
$
period-1 inventory is not charged correctly and period-2 inventory carries period-1 cost. A similar misalignment occurs if $h_{t+1}$ is bound to $I_t$.

\textbf{Consequence.} Cross-period evaluation becomes distorted. In settings with time-varying prices, holding costs, peak/off-peak tariffs, or tardiness penalties, the model may encourage early production when it should delay, or penalize inventory in the wrong period altogether.

\subsubsubsection{1.3.4 Wrong Unit or Scale Binding}
\label{app:subsubsubsec:wro_uni_sca_bin}

\textbf{Definition.} Parameters and variables are multiplied or combined under inconsistent units, scales, or normalization conventions. The resulting number no longer represents a meaningful economic quantity.

\textbf{Typical erroneous form.} Common patterns include forgetting a unit conversion factor (for example, multiplying a price in CNY/kWh by power in kW without multiplying by time length), summing quantities with incompatible units directly, or combining terms whose numeric magnitudes differ dramatically without normalization.

\textbf{Example.} In a data-center energy scheduling problem, if the electricity price in period $t$ is $\pi_t$ in CNY/kWh, server power is $p_t$ in kW, and the slot length is $\Delta t$ hours, then the electricity-cost term should be
$
\min \sum_t \pi_t p_t \Delta t.
$
If the code uses only $\sum_t \pi_t p_t$, it calculates a power charge instead of an energy charge. Similarly, adding minutes of delay directly to monetary cost without scaling or weighting destroys the interpretability of the objective.

\textbf{Consequence.} The objective value no longer corresponds to a real cost or reward, and the trade-off among plans becomes numerically distorted. Some costs are systematically underestimated or overestimated, and multi-criteria trade-offs lose business meaning because the objective is no longer dimensionally coherent.

\textbf{Summary.} Objective parameter and index binding errors do not necessarily change whether a term is present; they change who it applies to, where it applies, when it applies, and in what units it applies. These errors are therefore more hidden than term omission: the model still solves, but its economic interpretation and preference structure drift systematically.

\subsubsection{Multi-Objective and Risk-Objective Errors}
\label{app:subsubsec:mul_obj_ris_obj_err}

\textbf{Definition.} When a problem contains multiple objectives, hierarchical preferences, risk-aversion principles, or extreme-value performance measures, the generated model may fail to preserve the intended combination mechanism, priority structure, or risk representation. It may aggregate, average, replace, or reweight the objectives incorrectly, so the returned ``optimal'' solution is not optimal for the original decision semantics.

\textbf{Boundary.} This section concerns how multiple objectives are combined, ordered, and made to express risk preference. Errors in the direction, object, term composition, or parameter binding of a single objective belong to Secs.~\ref{app:subsubsec:obj_sem_map_err},\ref{app:subsubsec:obj_com_err},\ref{app:subsubsec:obj_par_ind_bin_err}. Errors in the concrete mathematical coding of piecewise, fractional, absolute-value, or extreme-value forms belong to Sec.~\ref{app:subsubsec:obj_for_err}.

\subsubsubsection{1.4.1 Wrong Multi-Objective Aggregation Mechanism}
\label{app:subsubsubsec:wro_mul_obj_agg_mec}

\textbf{Definition.} The original problem specifies a particular relation among objectives---for example lexicographic optimization, hierarchical optimization, stagewise optimization, weighted-sum optimization, or an $\epsilon$-constraint method---but the generated model mixes them up or replaces one mechanism with another non-equivalent one.

\textbf{Typical erroneous form.} The intended model solves first
$
\min f_1(x),
$
and then, within the set of $f_1$-optimal solutions, solves
$
\min f_2(x).
$
The code instead writes a single weighted objective such as
$
\min \bigl(f_1(x)+\lambda f_2(x)\bigr)
$
with an arbitrary $\lambda$, even though this is generally not equivalent to lexicographic optimization.

\textbf{Example.} In airport-recovery scheduling, suppose the true requirement is first to minimize the maximum flight delay and only then to minimize total delay. A correct lexicographic formulation solves
$
\min z \quad \text{s.t. } d_f \le z,\ \forall f,
$
and then, after obtaining the optimal value $z^\star$, solves
$
\min \sum_{f\in F} d_f \quad \text{s.t. } d_f \le z^\star,\ \forall f.
$
If the code instead minimizes a weighted sum $z+\lambda\sum_f d_f$ with an arbitrary $\lambda$, the resulting optimization problem is fundamentally different.

\textbf{Consequence.} The priority structure among objectives is destroyed. The model may accept a noticeably worse worst-case delay just to shave a little off total delay, or sacrifice a primary business criterion in exchange for a small improvement in a secondary one.

\subsubsubsection{1.4.2 Replacing a Risk or Extreme-Value Objective}
\label{app:subsubsubsec:rep_ris_ext_val_obj}

\textbf{Definition.} The original problem calls for optimization of a worst case, a tail-risk measure, a loss-exceedance criterion, or another extreme-value objective, but the generated model replaces it with an average-value or total-value proxy. A problem intended to manage robustness or tail behavior is thereby reduced to a mean-performance problem.

\textbf{Typical erroneous form.} Common substitutions include replacing
$
\min \max_{s\in S} L_s(x)
$
with
$
\min \frac{1}{|S|}\sum_{s\in S} L_s(x),
$
replacing a CVaR objective with an expectation objective, or replacing makespan and maximum-delay objectives with average-completion or total-delay objectives.

\textbf{Example.} In robust facility location, the true goal may be
$
\min \max_{s\in S} T_s(x),
$
so that the response time is controlled in the worst disaster scenario. If the code instead minimizes average response time, the model optimizes mean performance rather than worst-case acceptability. Likewise, a parallel-machine scheduling problem with objective $\min C_{\max}$ cannot be replaced by minimizing average completion time without changing the problem.

\textbf{Consequence.} Attention to the tail, the worst case, and extreme outcomes is systematically weakened. The model may look better on average while performing much worse in exactly the critical scenarios that motivated a robust or fairness-oriented formulation.

\subsubsubsection{1.4.3 Wrong Weight Setting or Interpretation}
\label{app:subsubsubsec:wro_wei_set_int}

\textbf{Definition.} In a weighted multi-objective model, all objective terms are present, but the chosen weights do not truly reflect business priorities or are misinterpreted because of scale differences, unit differences, or confusion between importance weights and conversion coefficients.

\textbf{Typical erroneous form.} A weighted sum such as
$
\min \sum_{r=1}^R \lambda_r f_r(x)
$
is used with arbitrary weights, with nearly equal weights assigned to terms of vastly different scales, or with importance weights confused with unit-conversion coefficients.

\textbf{Example.} In green production scheduling, one may jointly optimize production cost and carbon emissions:
$
\min \lambda_1\,\text{Cost}(x)+\lambda_2\,\text{Emission}(x).
$
If cost is measured in money and emissions in tons, setting $\lambda_1=\lambda_2=1$ without scale normalization can make one objective dominate the other numerically. Likewise, if management clearly states ``cost first, emissions second,'' but the weights are chosen arbitrarily, they no longer encode the true policy preference.

\textbf{Consequence.} The trade-off structure of the multi-objective model is distorted. Unlike Sec.~\ref{app:subsubsubsec:wro_mul_obj_agg_mec}, the combination mechanism itself may be correct, but the numerical parameters fail to express the intended priorities. As a result, one objective can become nominally present yet practically irrelevant, or a secondary objective can accidentally dominate the decision.

\textbf{Summary.} Multi-objective and risk-objective errors appear in three main ways: using the wrong combination mechanism, replacing worst-case or tail-risk criteria with mean-based proxies, and assigning weights that do not represent real managerial priorities. Compared with single-objective errors, the key issue is whether multiple objectives jointly define optimality in the intended way.

\subsubsection{Objective Formalization Errors}
\label{app:subsubsec:obj_for_err}

\textbf{Definition.} The semantic target of the objective is broadly correct and the relevant terms have been identified, but the mathematical form used in the optimization model fails to preserve the original functional structure. Piecewise, extreme-value, absolute-value, bilinear, and fractional objectives may be linearized, replaced, or transformed incorrectly. The error lies not in optimizing the wrong thing, but in implementing the shape of the right thing incorrectly.

\textbf{Boundary.} This section concerns the correctness of the mathematical form of the objective. Errors in direction, object, or objective-constraint role belong to Sec.~\ref{app:subsubsec:obj_sem_map_err}; errors in omitted or redundant terms, repeated counting, or local sign problems belong to Sec.~\ref{app:subsubsec:obj_com_err}; errors in parameter, index, time, or unit binding belong to Sec.~\ref{app:subsubsec:obj_par_ind_bin_err}; and errors in multi-objective mechanism or risk preference belong to Sec.~\ref{app:subsubsec:mul_obj_ris_obj_err}.

\subsubsubsection{1.5.1 Smoothing Piecewise or Stepwise Cost into a Continuous Linear Cost}
\label{app:subsubsubsec:smo_pie_ste}

\textbf{Definition.} The original cost structure contains fixed activation fees, stepped charges, piecewise-linear cost, or batch-triggered cost jumps, but the generated model smooths the structure into a simple linear term and ignores the discontinuous jumps that occur when another batch, vehicle, or machine must be activated.

\textbf{Typical erroneous form.} A transport or production cost that should look like ``fixed cost plus variable cost'' or a piecewise interval charge is replaced by an average linear unit cost. For example, a model of the form
$
\min \sum_k F_k y_k + \sum_k c_k q_k
$
is replaced by
$
\min \sum_k \bar c_k q_k,
$
thereby smearing the fixed or stepped component into the unit cost.

\textbf{Example.} In long-haul transport, suppose one standard truck carries at most $Q$ tons, each activated truck costs $F$, and variable shipping cost is $c$ per ton. If the transported quantity is $x$, then the true cost includes the integer number of trucks needed and a capacity relation such as $x\le Qn$ with $n\in\mathbb Z_+$. If the code instead uses a purely linear cost in $x$, the integer jump in truck activation disappears.

\textbf{Consequence.} The discrete activation cost is systematically underestimated. The model incorrectly learns that many small batches are almost equivalent to consolidated transport, and it underestimates the required number of vehicles, batches, machine starts, or setup changes.

\subsubsubsection{1.5.2 Incorrect Handling of Nonlinear or Nonsmooth Objective Terms}
\label{app:subsubsubsec:inc_han_non}

\textbf{Definition.} The objective contains a max operator, an absolute value, a bilinear term, or another nonlinear or nonsmooth structure. The model recognizes this superficially but fails to introduce the required auxiliary variables, missing equivalent constraints, or a suitable model class, so the expression is only ``written down'' rather than modeled correctly.

\textbf{Typical erroneous form.} Examples include defining an auxiliary variable $z$ for a max objective but forgetting the linking constraints $z\ge C_j$ for all $j$, using $|x_1-x_2|$ in a linear model without introducing a correct epigraph, or keeping a bilinear product directly inside a linear or mixed-integer linear framework without using a compatible nonlinear solver or relaxation.

\textbf{Example.} In parallel-machine scheduling, minimizing makespan is typically modeled by introducing an auxiliary variable $z$ and solving
$
\min z \quad \text{s.t. } z\ge C_j,\ \forall j.
$
If the code only declares $z$ and minimizes it, without the dominance constraints, then $z$ can collapse to zero and disconnect from the actual schedule. Similarly, minimizing an absolute load difference requires the full epigraph formulation rather than a symbolic absolute value left unresolved.

\textbf{Consequence.} The objective becomes detached from the real performance measure. The model may look as if it uses auxiliary variables or nonlinear expressions, but the solver is no longer solving the intended extreme-value, absolute-deviation, or bilinear-cost problem.

\subsubsubsection{1.5.3 Fractional Objective Errors}
\label{app:subsubsubsec:fra_obj_err}

\textbf{Definition.} The original objective has a ratio form such as benefit-to-cost, output-to-energy, or efficiency per unit resource, but the generated model does not use a suitable fractional-programming treatment. It inserts the ratio directly into a modeling environment that does not support it, or rewrites it without checking the conditions required for a valid transformation.

\textbf{Typical erroneous form.} A problem of the form
$
\max \frac{a^\top x + \alpha}{b^\top x + \beta}
$
is inserted directly into a linear or mixed-integer linear objective, without a Charnes--Cooper transformation, Dinkelbach iteration, or any other appropriate method, and without checking positivity of the denominator.

\textbf{Example.} In data-center task scheduling, the objective may be to maximize revenue per unit energy:
$
\max \frac{\sum_i r_i x_i}{\sum_i e_i x_i + E_0},
$
where $x_i$ indicates whether task $i$ is assigned, $r_i$ is revenue, $e_i$ is added energy use, and $E_0$ is baseline energy. If this ratio is written directly as the objective of a standard linear Gurobi model, the model is no longer an ordinary LP or MILP even if the syntax is accepted.

\textbf{Consequence.} The mathematical class of the model is misidentified. In mild cases the model cannot be solved or returns an error; in severe cases it produces a numerically plausible but theoretically unjustified answer. Fractional objectives usually encode efficiency, so mishandling them can degenerate the problem into maximizing the numerator alone or minimizing the denominator alone.

\textbf{Summary.} Objective formalization errors are not about misunderstanding what should be optimized, but about failing to write the intended objective in a mathematically faithful and solver-compatible way. Typical manifestations are smoothing stepwise cost, incompletely handling max/absolute/bilinear structures, and using fractional objectives without the necessary transformations.

\subsection{Decision-Variable Modeling Hallucinations}
\label{app:subsec:dec_var_hal}

\subsubsection{Variable Semantic Definition Errors}
\label{app:subsubsec:var_sem_def_err}

\textbf{Definition.} This category covers failures to map decision objects, system states, and control actions in a natural-language task into decision variables with the correct semantic meaning. The issue is not simply that the variable type or bound is wrong; it is that the variable already misrepresents the modeling object at the definition stage, so even a formally complete objective and constraint set may still be built on semantically wrong variables.

\textbf{Boundary.} The focus here is what the variable is supposed to represent. If the issue is variable type, bounds, or allowed values, it belongs to Sec.~\ref{app:subsubsec:var_dom_typ_err}. If it concerns index sets, subscript binding, or missing scenario dimensions, it belongs to Sec.~\ref{app:subsubsec:var_ind_set_bin_err}. If the problem is missing coupling, consistency, or master-auxiliary binding among variables, it belongs to Sec.~\ref{app:subsubsec:var_rol_cou_err}.

\subsubsubsection{2.1.1 Wrong Variable Object}
\label{app:subsubsubsec:wro_var_obj}

\textbf{Definition.} The code defines a variable for the wrong decision object. A problem that needs a selection variable, assignment variable, flow variable, or state variable is modeled with a different kind of variable that has a different semantics.

\textbf{Typical erroneous form.} A discrete-choice decision should be represented by a binary variable such as
$
 x_{ijk}\in\{0,1\},
$
indicating whether vehicle $k$ traverses arc $(i,j)$, but the code defines only a continuous flow variable
$
 f_{ijk}\ge 0,
$
and assumes that ``positive flow means the arc is chosen.'' This replaces a path-choice decision with a splittable-flow interpretation.

\textbf{Example.} In a last-mile routing problem, each vehicle should form a discrete route. A typical variable is
$
 x_{ijk}=1
$
if vehicle $k$ travels from node $i$ to node $j$, and $0$ otherwise. If the code defines only arc flow $f_{ij}\ge 0$ and interprets ``has flow'' as ``uses the arc,'' the model no longer describes which arcs vehicles traverse; it describes how commodity flow passes through a network. The former is route choice, the latter is closer to network flow.

\textbf{Consequence.} The decision object of the entire problem is changed. A discrete route can become a splittable flow, a machine on/off decision can become a continuous output variable, and a state-transition problem can become a static allocation problem. The solver is then solving a different problem created by the wrong variable object.

\subsubsubsection{2.1.2 Missing a Key Dimension}
\label{app:subsubsubsec:mis_key_dim}

\textbf{Definition.} A variable should depend on key dimensions such as time, scenario, resource type, location, or product class, but one of those dimensions is omitted. The variable therefore cannot distinguish among different objects, stages, or states and loses the expressive power required by the original problem.

\textbf{Typical erroneous form.} The intended model defines variables such as $x_{it}$, $y_{kst}$, or $z_{ij\omega}$, but the code uses only $x_i$, $y_k$, or $z_{ij}$, effectively erasing the time, scenario, or resource-type dimension.

\textbf{Example.} In a multi-period production-and-sales plan, the production of product $p$ in month $t$ should be represented by $q_{pt}$ and inventory by $I_{pt}$, so that inventory balance can be written as
$
I_{p,t-1}+q_{pt}-d_{pt}=I_{pt}.
$
If the code defines only an aggregate production variable $q_p$ rather than monthly production $q_{pt}$, the model can no longer distinguish when production occurs, and the cross-period inventory relation cannot be encoded correctly.

\textbf{Consequence.} The model loses the ability to distinguish time, scenario, or heterogeneous object types. What looks like ``one missing subscript'' often collapses a multi-period, multi-scenario, or multi-resource problem into a static aggregate one.

\subsubsubsection{2.1.3 Wrong Aggregation Level of Variables}
\label{app:subsubsubsec:wro_agg_lev}

\textbf{Definition.} The original problem requires decisions at a finer granularity---for example by machine, order, period, or node---but the generated model aggregates those variables too early into totals. Structural differences among objects and local constraints can no longer be expressed.

\textbf{Typical erroneous form.} The intended model defines local variables such as $x_{it}$ or $x_{ijm}$, but the code uses only aggregated totals such as $x_t$ or $x_i$. The issue is not the complete absence of variables, but premature aggregation.

\textbf{Example.} In a manufacturing shop with multiple lines, the production of line $m$ in period $t$ should be modeled as $q_{mt}$ so that line-specific capacity constraints $q_{mt}\le C_{mt}$ can be enforced. If the code uses only total production $q_t$, the model may still satisfy total demand but can no longer say which line produces what, nor can it enforce per-line capacity, switching, or maintenance constraints.

\textbf{Consequence.} The model may look valid at the aggregate level yet fail at the execution level. Constraints that should apply locally---single-machine capacity, warehouse inventory, vehicle routes, station workload, and so on---lose the objects to which they should apply.

\subsubsubsection{2.1.4 State-Control Confusion}
\label{app:subsubsubsec:sta_con_con}

\textbf{Definition.} State variables and control variables are confused. A state that should be determined through transition equations, conservation relations, or system dynamics is treated as a freely chosen decision variable, or a true control decision is treated as if it were merely a passive state.

\textbf{Typical erroneous form.} A state should satisfy something like
$
 s_{t+1}=s_t+u_t-d_t,
$
where $s_t$ is the state and $u_t$ is the control. The code instead treats $s_t$ as an independent decision variable subject only to simple bounds, with no dynamic relation to $u_t$, exogenous disturbances, or the initial condition.

\textbf{Example.} In multi-period inventory planning, end-of-period inventory should satisfy
$
I_t=I_{t-1}+q_t-d_t,
$
where $I_t$ is the state and $q_t$ is the control. If the code defines $I_t$ as a free decision variable with only box constraints such as $0\le I_t\le \bar I$, inventory no longer represents the result of previous inventory plus replenishment minus demand.

\textbf{Consequence.} The dynamic cause-and-effect structure of the model collapses. Inventory, energy, machine wear, queue length, and similar states stop reflecting accumulated history and control actions, so the model can produce pseudo-feasible plans such as ``inventory exists although nothing was produced'' or ``battery charge exists although nothing was charged.'' 

\textbf{Summary.} Variable semantic definition errors arise when the model fails, from the outset, to express the right decision object, the right state, the right distinction among dimensions and levels, or the difference between state and control. Once variable semantics are wrong, later objectives and constraints may still be written carefully, but they are optimizing the wrong mathematical objects.

\subsubsection{Variable Domain and Type Errors}
\label{app:subsubsec:var_dom_typ_err}

\textbf{Definition.} Here the variable corresponds to the right modeling object, but the set of values it is allowed to take is wrong. A variable that should be binary, integer, nonnegative, bounded, or sign-free is placed in the wrong mathematical domain, thereby reshaping the feasible region directly.

\textbf{Boundary.} The issue is not what the variable means, but what values it may take. Errors in variable semantics belong to Sec.~\ref{app:subsubsec:var_sem_def_err}. Missing coupling or logical consistency among variables belongs to Sec.~\ref{app:subsubsec:var_rol_cou_err}. Errors in constraints rather than variable declarations belong to Sec.~\ref{app:subsec:con_mod_hal}.

\subsubsubsection{2.2.1 Relaxing a Discrete Variable into a Continuous Variable}
\label{app:subsubsubsec:rel_dis_var}

\textbf{Definition.} A variable that should take discrete values---binary, integer, counting, or categorical levels---is incorrectly modeled as continuous, allowing meaningless intermediate values such as ``half a warehouse'' or ``0.7 of a truck.''

\textbf{Typical erroneous form.} A variable that should satisfy $y_i\in\{0,1\}$ or $n_i\in\mathbb Z_+$ is modeled as $0\le y_i\le 1$ or $n_i\ge 0$ only, thereby relaxing a discrete choice problem into a continuous one.

\textbf{Example.} In facility location, $y_j\in\{0,1\}$ indicates whether warehouse $j$ is opened. If the code instead uses $0\le y_j\le 1$, the model may return values such as $0.4$ or $0.7$, which amount to ``partially opened warehouses'' with no business meaning.

\textbf{Consequence.} The model becomes an artificially relaxed version of the original discrete problem, often producing overly optimistic lower bounds or non-implementable fractional solutions.

\subsubsubsection{2.2.2 Forcing a Continuous Variable to Be Discrete}
\textbf{Definition.} 
\label{app:subsubsubsec:for_con_var}

A variable that should be continuously adjustable over an interval is incorrectly declared as integer, binary, or restricted to a small set of levels. The feasible region is compressed artificially and the model loses adjustment precision.

\textbf{Typical erroneous form.} A variable that should satisfy $0\le x_i\le \bar x_i$ with $x_i\in\mathbb R$ is instead modeled as $x_i\in\mathbb Z_+$ or even $x_i\in\{0,1\}$.

\textbf{Example.} In a blending problem, the amount of ingredient $i$ should be continuous because it can be dosed by weight or mass. If the code declares $x_i$ as an integer variable, or even worse as binary, it allows only coarse and unrealistic dosage levels.

\textbf{Consequence.} The model loses genuine fine-grained flexibility. Costs may rise, recipe accuracy may fall, utilization may worsen, and a problem that was originally feasible can become infeasible after unnecessary discretization.

\subsubsubsection{2.2.3 Wrong Value Range}
\label{app:subsubsubsec:wro_val_ran}

\textbf{Definition.} The variable type may be correct, but the lower bound, upper bound, capacity limit, feasible interval, or candidate value set is specified incorrectly, so the variable is allowed to range too widely or too narrowly.

\textbf{Typical erroneous form.} The intended variable should satisfy
$
\underline x_i \le x_i \le \bar x_i,
$
but the code uses $0\le x_i\le M$ with an arbitrary $M$, or omits one side of the bounds entirely.

\textbf{Example.} In staff scheduling, if the weekly work-hour variable $h_i$ must satisfy $0\le h_i\le 40$, then writing only $h_i\ge 0$ allows unrealistic overload, while replacing the upper bound with $8$ incorrectly shrinks a weekly cap into a daily one.

\textbf{Consequence.} An overly loose range allows the solver to exploit unrealistic extreme values; an overly tight range excludes valid plans and may raise cost or cause infeasibility.

\subsubsubsection{2.2.4 Wrong Sign Domain}
\label{app:subsubsubsec:wro_sig_dom}

\textbf{Definition.} The sign restriction on a variable is wrong. A variable that should be nonnegative, nonpositive, free, or box-bounded is placed in a different sign domain and thereby loses its intended physical or business meaning.

\textbf{Typical erroneous form.} A variable that should satisfy $x_i\ge 0$ is left free, or a deviation variable that should range over $\mathbb R$ is incorrectly constrained by $z_i\ge 0$.

\textbf{Example.} Shipment quantity $x_{ij}$ from a warehouse to a store should satisfy $x_{ij}\ge 0$. If the nonnegativity constraint is missing, the model may use negative shipment values to offset other balances mathematically. Conversely, if a supply-demand deviation variable $z_t$ is meant to capture actual minus planned quantity, it should be allowed to take both positive and negative values; constraining it to be nonnegative suppresses one side of the deviation.

\textbf{Consequence.} Nonnegative physical quantities can become unrealistically negative, while variables that should represent signed deviations lose half of their expressive range. The result is either pseudo-feasible solutions or the loss of legitimate corrective actions.

\textbf{Summary.} Variable domain and type errors do not change what a variable is meant to represent, but they change what values it may take, and therefore reshape the feasible region directly. The most common cases are relaxing discrete variables, discretizing continuous variables, mis-setting bounds, and using the wrong sign domain.

\subsubsection{Variable Index and Set Binding Errors}
\label{app:subsubsec:var_ind_set_bin_err}

\textbf{Definition.} The basic semantics and domain of the variable may both be correct, but the variable is defined over the wrong object set, with the wrong indices, or with missing contextual dependence. The variable therefore exists on the wrong object domain and the model describes the wrong individuals, roles, scenarios, or layers.

\textbf{Boundary.} The key question is over which objects and indices the variable should be defined. Errors in variable meaning belong to Sec.~\ref{app:subsubsec:var_sem_def_err}, errors in domain belong to Sec.~\ref{app:subsubsec:var_dom_typ_err}, and errors in objective or constraint summation belong elsewhere. In contrast to Sec.~\ref{app:subsubsec:var_sem_def_err}, the variable object here is usually broadly correct; the problem lies in the index structure on which it is instantiated.

\subsubsubsection{2.3.1 Omitted Index Set}
\label{app:subsubsubsec:omi_ind_set}

\textbf{Definition.} An entire class of objects that should have its own variables is omitted from the variable definition, so the model contains no decision representation for that class at all.

\textbf{Typical erroneous form.} The intended model should define variables for two or more object sets, for example one variable family for captains and another for first officers, but the code builds variables for only one of them.

\textbf{Example.} In crew scheduling, a duty chain may require both captains and first officers. If the model defines assignment variables only for captains and omits the corresponding first-officer variables entirely, it can represent captain assignment but not the full crew configuration.

\textbf{Consequence.} One whole category of actors or resources disappears from the model. The optimization may still produce a formal solution, but a crucial role or resource turns out never to have been assigned.

\subsubsubsection{2.3.2 Fabricated Index Set}
\label{app:subsubsubsec:fab_ind_set}

\textbf{Definition.} The model invents a classification, hierarchy, or object partition that is not supported by the input data, the problem statement, or the business rules, and then defines variables over that fabricated structure.

\textbf{Typical erroneous form.} The original problem provides only an object set or parameter table, but the code assumes additional subsets or a classification mapping and defines variables over those self-invented groups.

\textbf{Example.} Suppose a distribution-network problem provides a set of customer nodes and pairwise distances but no regional grouping. If the code assumes that nodes~1--5 belong to region~A and nodes~6--10 belong to region~B, and then defines regional assignment variables on that basis, the variable index structure is no longer grounded in the original data.

\textbf{Consequence.} The original problem is silently turned into a different one with an additional layer of structure. The solution may be self-consistent relative to the invented taxonomy, but it has no reliable interpretation for the original task.

\subsubsubsection{2.3.3 Wrong Subscript Binding}
\label{app:subsubsubsec:wro_sub_bin}

\textbf{Definition.} A variable is defined over multiple indices, but later calls or expansions swap the order of those indices or bind them to the wrong sets, so the variable is used as if it represented a completely different object relation.

\textbf{Typical erroneous form.} The intended variable is $x_{ij}$ with $i\in I$ and $j\in J$, but the code later uses $x_{ji}$; similarly, $x_{ijm}$ is used where $x_{imj}$ was intended. When the subscripts refer to different sets, this is not a typographical issue but a semantic one.

\textbf{Example.} In crew-flight matching, let $x_{cf}\in\{0,1\}$ indicate whether crew $c$ serves flight $f$. If a later constraint mistakenly uses $x_{fc}$, the language runtime may not always raise an immediate error, especially if the index sets have compatible sizes. Yet the semantic meaning has already shifted from crew-flight assignment to something else.

\textbf{Consequence.} The variable does not disappear; instead it is taken to represent the wrong object relation. The wrong relation then propagates into objectives and constraints, corrupting the entire model under an apparently consistent notation.

\subsubsubsection{2.3.4 Missing Index Dependence}
\label{app:subsubsubsec:mis_ind_dep}

\textbf{Definition.} The variable should depend on an additional index dimension such as time, scenario, mode, resource type, or level label, but that dependence is omitted even though the main object and basic index structure are kept. The same object can then no longer be distinguished across contexts.

\textbf{Typical erroneous form.} Variables that should be written as $x_{i\omega}$, $y_{itm}$, or $z_{ijs}$ are simplified into $x_i$, $y_{it}$, or $z_{ij}$, so the variable still exists but no longer depends on scenario $\omega$, mode $m$, or scenario index $s$.

\textbf{Example.} In a robust supply-chain plan, shipment decisions may need to be scenario-dependent, for example $x_{ijs}$ for the shipment from warehouse $i$ to customer $j$ under demand scenario $s$. If the code uses only $x_{ij}$, it can no longer represent different recourse actions or resource consequences across scenarios.

\textbf{Consequence.} Context-dependent decisions are wrongly compressed into a single context-free decision. Robust problems become single-scenario problems, multimode problems become single-mode problems, and heterogeneous-resource problems become homogeneous ones.

\textbf{Summary.} Variable index and set binding errors occur when variables are created but not on the correct object domain or contextual index structure. Typical cases include missing a whole object class, inventing unsupported index sets, swapping subscripts, and omitting a needed contextual index.

\subsubsection{Variable Role Coupling Errors}
\label{app:subsubsec:var_rol_cou_err}

\textbf{Definition.} Individual variables may each have the right semantics, type, and index structure, but the activation, mapping, consistency, subordination, or support relations that should connect multiple variables are not established correctly. The problem is no longer ``what each variable is,'' but whether the variables work together in the right roles.

\textbf{Boundary.} This section concerns how multiple variables coordinate with one another. Errors in variable meaning, domain, or indexing belong to Secs.~\ref{app:subsubsec:var_sem_def_err}--\ref{app:subsubsec:var_ind_set_bin_err}. If the issue is the absence of a core structural constraint rather than a missing variable-to-variable role relation, it belongs to Sec.~\ref{app:subsec:con_mod_hal}.

\subsubsubsection{2.4.1 Variable Not Taking Effect}
\label{app:subsubsubsec:var_not_tak}

\textbf{Definition.} A variable is declared, but it never enters the key objective or any core constraint that links it to other variables or the problem structure. It remains suspended in the model: present in appearance but irrelevant in substance.

\textbf{Typical erroneous form.} The code defines a variable such as $z\in\mathbb R^n$ or $y_{ij}\in\{0,1\}$, but the objective and constraints never really use it, or use it only in an isolated expression with no bearing on feasibility or optimality.

\textbf{Example.} In flight-recovery scheduling, the model may define a connection variable $y_{fg}$ indicating whether flight $g$ follows flight $f$. If the model never uses $y_{fg}$ to enforce temporal compatibility, crew-chain continuity, aircraft continuity, or any cost term, then the connection variable does not actually model the connection structure.

\textbf{Consequence.} The model creates the illusion that a certain structure has been encoded because the variable name exists, but the variable has no effect on feasibility or optimality.

\subsubsubsection{2.4.2 Duplicate Variable Roles}
\label{app:subsubsubsec:dup_var_rol}

\textbf{Definition.} Two or more variables are introduced to describe the same business fact or structural relation, but no consistency constraints are added. The model then contains multiple incompatible numerical versions of the same reality.

\textbf{Typical erroneous form.} The model defines both $x_{ij}\in\{0,1\}$ and $y_{ij}\in\{0,1\}$ to indicate whether arc $(i,j)$ is used, but it omits a tying relation such as $x_{ij}=y_{ij}$ for all $(i,j)$.

\textbf{Example.} In vehicle routing, one variable family may be described as path variables and another as visit variables, yet both actually mean whether vehicle $k$ uses arc $(i,j)$. If no linking equalities are added, some constraints operate on $x$ and others on $y$, and the model may end up describing two inconsistent ``virtual routes.''

\textbf{Consequence.} The model becomes internally inconsistent while still looking complete. A single business fact is represented by parallel variable systems with no enforced agreement.

\subsubsubsection{2.4.3 Missing Coupling Between Master and Auxiliary Variables}
\label{app:subsubsubsec:mis_cou_bet}

\textbf{Definition.} Auxiliary variables are introduced for activation, costing, linearization, capacity, or logical state, but they are not properly linked to the main decision variables through upper bounds, lower bounds, equalities, or implication constraints. The auxiliary variables therefore fail to reflect the main decision accurately.

\textbf{Typical erroneous form.} Let $x_{ij}\ge 0$ be a main flow variable and $y_i\in\{0,1\}$ an activation variable. To express ``flow is allowed only if facility $i$ is open,'' one usually needs a coupling constraint such as
$
\sum_j x_{ij}\le M_i y_i.
$
If that coupling is missing, the facility may ship when closed, or the activation variable may turn on without any actual activity.

\textbf{Example.} In warehouse location with distribution, $y_i\in\{0,1\}$ may indicate whether warehouse $i$ is opened and $x_{ij}\ge 0$ the amount shipped from warehouse $i$ to customer $j$. Without constraints like
$
\sum_j x_{ij}\le M_i y_i
$
or finer ones such as $x_{ij}\le d_j y_i$, the model can produce the absurd solution ``warehouse closed but still shipping.''

\textbf{Consequence.} Costs, activation indicators, and linearization variables become decoupled from the real business actions they are supposed to represent. This creates a classic class of pseudo-feasible solutions such as ``not opening a warehouse but still delivering,'' ``not starting a machine but still producing,'' or ``selecting no mode but still consuming resources.''

\textbf{Summary.} Variable role coupling errors arise when the variables themselves may look correct, yet their activation, consistency, or support relations are not wired together. The variable system then becomes locally plausible but globally false.

\subsubsection{Variable Redundancy and Uncontrolled Symmetry}
\label{app:subsubsec:var_red_unc_sym}

\textbf{Definition.} The model broadly captures the right decision objects, but the variable design fails to exploit problem structure. It introduces unnecessary extra variables, ignores opportunities to merge or compress them, or leaves large sets of perfectly symmetric variables untreated. The resulting model is bloated, filled with equivalent branches, and more vulnerable to later indexing or constraint errors.

\textbf{Boundary.} The focus here is the structural economy of the variable system. The problem is not primarily wrong variable meaning or missing coupling, but needless expansion, duplication, or symmetry at the variable-design level.

\subsubsubsection{2.5.1 Unnecessary Redundant Variables}
\label{app:subsubsubsec:unn_red_var}

\textbf{Definition.} Extra variables are introduced even though they carry no new business semantics and serve no clear purpose such as linearization, logical encoding, decomposition, or numerical stabilization. They merely duplicate information already present in the model.

\textbf{Typical erroneous form.} A main variable already captures the decision state, but the code adds another variable family that neither introduces new structure nor performs a necessary intermediate modeling function.

\textbf{Example.} In facility location with distribution, suppose $y_i$ already indicates whether facility $i$ is open and $x_{ij}$ already captures its assignments or flows. If the code adds another binary variable $z_i$ also meaning whether facility $i$ is used, yet $z_i$ is not needed for any extra logic or decomposition, it is just a duplicate encoding of the same fact.

\textbf{Consequence.} The model becomes larger without gaining expressive power. Consistency constraints, index management, debugging difficulty, and solve effort all increase, and the redundant design may even hide whether the truly necessary structure has been encoded correctly.

\subsubsubsection{2.5.2 Unhandled Symmetric Variables}
\label{app:subsubsubsec:unh_sym_var}

\textbf{Definition.} A set of variables or variable blocks is fully interchangeable in business meaning and constraint structure, but no symmetry-breaking device is used. The solver then explores many equivalent branches corresponding to the same underlying business solution.

\textbf{Typical erroneous form.} For identical machines, vehicles, warehouses, or crews, variable groups such as $x_{ik}$ and $x_{i\ell}$ are interchangeable after swapping indices $k$ and $\ell$, yet the model contains no symmetry-breaking relations such as ordered loads or representative-element rules.

\textbf{Example.} In identical parallel-machine scheduling with two identical machines, any solution can be mirrored by swapping the machine labels. If the model includes no symmetry-breaking condition such as ordering total loads across machines, branch-and-bound will spend time exploring equivalent allocations again and again.

\textbf{Consequence.} Feasibility semantics do not change much, but solve performance can deteriorate dramatically. For LLM-generated models, this often amplifies other small issues because large symmetric search trees make the model appear unstable, excessively slow, or even falsely infeasible.

\subsubsubsection{2.5.3 Failure to Exploit Sparsity}
\label{app:subsubsubsec:fai_exp_spa}

\textbf{Definition.} Variables should be defined only on candidate sets, allowed arcs, compatible pairs, or local neighborhoods, but the generated model ignores this sparsity and defines variables over the full Cartesian product. Many variables that should not exist are introduced indiscriminately.

\textbf{Typical erroneous form.} The intended formulation defines variables only for $(i,j)\in A\subseteq N\times N$, but the code defines $x_{ij}$ for all $i,j\in N$. Similarly, matching variables that should exist only over a compatibility set are created over all pairs.

\textbf{Example.} In distribution-network design, only arcs in a feasible candidate set $A$ may be built or used. A correct model defines $x_{ij}$ only for $(i,j)\in A$. If the code defines $x_{ij}$ for every pair of nodes, it bloats the model immediately, and if the later constraints do not suppress all illegal arcs perfectly, the solver may exploit nonexistent routes.

\textbf{Consequence.} There is both a scale risk and a semantic risk. The variable count explodes from $|A|$ to $|N|^2$, and the model may expose illegal arcs, illegal pairings, or illegal actions that the solver can exploit if the later constraints are not airtight.

\textbf{Summary.} Variable redundancy and uncontrolled symmetry concern an overgrown variable system. Common manifestations are unnecessary extra variables, untreated symmetric variable families, and failure to exploit sparsity. These issues may not always change the business semantics directly, but they enlarge the model, slow down solution, and increase the chance of later encoding mistakes.

\subsection{Constraint Modeling Hallucinations}
\label{app:subsec:con_mod_hal}

\subsubsection{Constraint Semantic Translation Errors}
\label{app:subsubsec:con_sem_tra_err}

\textbf{Definition.} A business rule, logical condition, coverage requirement, feasibility criterion, or eligibility restriction from the natural-language task is indeed recognized as something that should become a constraint, but the mathematical constraint does not preserve the original semantics. The constraint exists, yet it no longer means the same rule.

\textbf{Boundary.} The question here is what rule the natural-language statement was translated into. If a necessary class of constraints was not built at all, the issue belongs to Sec.~\ref{app:subsubsec:mis_con_ske}. If the constraint exists but its index, boundary, strength, or direction is wrong, the issue belongs to later sections. If the real problem lies in variable definitions rather than in the translation of the rule, it belongs to Sec.~\ref{app:subsec:dec_var_hal}.

\subsubsubsection{3.1.1 Wrong Interpretation of the Rule}
\label{app:subsubsubsec:wro_int_rul}

\textbf{Definition.} The model recognizes that a business rule should become a constraint, but misunderstands the operative meaning of the rule itself---what is allowed, what is forbidden, or what process structure is required. The resulting constraint describes another set of feasible actions.

\textbf{Typical erroneous form.} A rule saying that only certain actions are allowed, or that an action may occur only under a particular condition, is translated into a broader or simply different relation. Examples include turning ``can only substitute for A'' into ``can substitute for anything,'' or turning ``must be continuous'' into ``total amount is sufficient.''

\textbf{Example.} In maintenance scheduling, once major repair starts on a machine, it must continue without interruption until completion. If $x_{mt}\in\{0,1\}$ indicates whether machine $m$ is under maintenance in period $t$, this continuity requirement cannot be expressed by a total-duration condition alone such as
$
\sum_t x_{mt}=L_m,
$
where $L_m$ is the required maintenance length. That condition allows maintenance to be scattered across nonconsecutive dates. It describes ``maintenance lasted this long in total,'' not ``maintenance was continuous.''

\textbf{Consequence.} The feasible action logic of the original problem is replaced by a weaker or otherwise different one. The model may satisfy aggregate totals while violating process semantics, producing maintenance splits, broken tasks, forbidden substitutions, or other operationally infeasible plans.

\subsubsubsection{3.1.2 Wrong Interpretation of Quantifiers}
\label{app:subsubsubsec:wro_int_qua}

\textbf{Definition.} Quantifiers such as ``at least,'' ``at most,'' ``exactly one,'' ``unique,'' ``there exists,'' and ``for every'' are mistranslated, so the counting semantics or existence semantics of the rule changes.

\textbf{Typical erroneous form.} ``Exactly one'' becomes ``at least one,'' ``at most one'' becomes ``at least one,'' ``every object must satisfy'' becomes ``there exists some object that satisfies,'' and ``for every time period'' becomes ``for some periods.''

\textbf{Example.} In flight-crew assignment, every flight may require exactly one captain. If $x_{cf}\in\{0,1\}$ indicates whether captain $c$ is assigned to flight $f$, the correct coverage constraint is
$
\sum_{c\in C(f)} x_{cf}=1,\quad \forall f,
$
where $C(f)$ is the set of qualified captains for flight $f$. If the code writes $\sum_{c\in C(f)}x_{cf}\ge 1$, then assigning two or more captains to the same flight is deemed feasible.

\textbf{Consequence.} A tiny change in an inequality can transform unique coverage into repeated coverage, universal satisfaction into existential satisfaction, or pointwise feasibility into sporadic feasibility.

\subsubsubsection{3.1.3 Confusing Resource Amount with Time Feasibility}
\label{app:subsubsubsec:con_res_amo}

\textbf{Definition.} The model treats the existence of enough resources in total as if that were equivalent to the temporal feasibility of the schedule. It counts resources but fails to encode their availability along time, space, or task chains.

\textbf{Typical erroneous form.} Rules such as ``the same crew cannot operate overlapping flights'' or ``a vehicle can go to the next task only after finishing the previous one and repositioning'' are replaced by aggregate bounds like $\sum_j x_j\le R$, as though total resource count were sufficient.

\textbf{Example.} A call center with 20 agents does not automatically mean that 20 customer requests can be served at any instant. If $x_{it}\in\{0,1\}$ indicates whether agent $i$ works in period $t$, feasibility depends on shift continuity, rest windows, and non-overlap in each period. If the code uses only a total-agent-count restriction and then concludes that any period can handle 20 tasks, it mistakes static resource availability for temporal scheduling feasibility.

\textbf{Consequence.} The model behaves as if a resource were available everywhere and at all times once it exists in total. Employees, crews, vehicles, or machines may be double-booked in time, and transfer, rest, or sequence conditions are ignored.

\subsubsubsection{3.1.4 Constraint Type Mismatch}
\label{app:subsubsubsec:con_typ_mis}

\textbf{Definition.} A business rule should be represented as a specific type of constraint---such as eligibility, precedence, coverage, compatibility, capacity, or logical implication---but the generated model writes it as another type, so the mathematical form does not match the rule semantics.

\textbf{Typical erroneous form.} Eligibility is encoded as a capacity bound, precedence is encoded as a total-hours restriction, or an incompatibility rule is encoded as a coverage requirement.

\textbf{Example.} Suppose task $j$ may be executed only by employee $i$ if qualification parameter $a_{ij}=1$, with assignment variable $x_{ij}\in\{0,1\}$. The natural eligibility constraint is
$
 x_{ij}\le a_{ij},\quad \forall i,j.
$
If the code instead writes something like a workload sum bounded by $a_{ij}$, then ``is eligible'' has been turned into ``can do at most this much.''

\textbf{Consequence.} The model ends up with a formally tidy inequality that encodes the wrong type of business relation. Since coverage, capacity, compatibility, and precedence can all look like linear inequalities, this is an especially easy trap for automatically generated models.

\textbf{Summary.} Constraint semantic translation errors do not mean that a constraint is absent; they mean that the wrong rule has been written down. The model may appear complete, yet the constraint is no longer the one intended by the original problem.

\subsubsection{Missing Constraint Skeleton}
\label{app:subsubsec:mis_con_ske}

\textbf{Definition.} A class of core constraints that determines the basic topology of the feasible region is missing, or only fragments of it are present. Variables therefore lack the fundamental links required for balance, continuity, coverage, capacity, or boundary conditions. The model may still run, but it no longer represents the basic operating mechanism of the original problem.

\textbf{Boundary.} The focus is whether the structural backbone of the model exists at all. If a rule has been written but semantically mistranslated, it belongs to Sec.~\ref{app:subsubsec:con_sem_tra_err}. If the backbone exists but is indexed, bounded, or directed incorrectly, the issue belongs to later sections. Sec.~\ref{app:subsubsec:mis_con_ske} asks whether the frame was built, not whether a built frame is slightly crooked.

\subsubsubsection{3.2.1 Missing Balance, Conservation, or Recursion Constraints}
\label{app:subsubsubsec:mis_bal_con}

\textbf{Definition.} The model omits constraints that should connect adjacent periods, nodes, or states through balance, conservation, or recursion. The state evolution no longer follows the logic of inflow, outflow, and stock change.

\textbf{Typical erroneous form.} Relations such as
$
I_t=I_{t-1}+q_t-d_t,
$
$
\sum_j x_{ji}-\sum_j x_{ij}=b_i,
$
or
$
s_{t+1}=f(s_t,u_t)
$
are missing, while local capacity or single-period requirements are retained.

\textbf{Example.} In multi-period inventory planning, production $q_t$, demand $d_t$, and ending inventory $I_t$ should be tied by inventory recursion. If the model contains only capacity limits on $q_t$ and local demand satisfaction inequalities, but no inventory recursion, inventory in different periods is no longer linked and may appear ``from nowhere.''

\textbf{Consequence.} Inventory, energy, cash, material, bed occupancy, or queue length cease to be governed by history. The model can produce pseudo-feasible plans such as having stock in a period without previous production or inflow.

\subsubsubsection{3.2.2 Missing Chain or Path Structure Constraints}
\label{app:subsubsubsec:mis_cha_pat}

\textbf{Definition.} The model omits the core constraints that make a set of local connections form a valid path, chain, sequence, or connected network structure. Local compatibility may exist, but global structural legitimacy does not.

\textbf{Typical erroneous form.} Degree-balance, start-end, predecessor-successor, or continuity constraints are missing. The code defines connection variables and perhaps some local compatibility conditions, but not the full chain skeleton.

\textbf{Example.} In crew-flight connection modeling, a variable may indicate whether flight $g$ follows flight $f$ in the same duty chain. To form a legal chain, each interior flight must have appropriate predecessor and successor relations, along with start, end, or base conditions. If only time-connectivity conditions are coded, the model may produce branches, disconnected fragments, or multiple successors for one flight.

\textbf{Consequence.} Connection variables can say ``these two tasks could connect locally,'' but not ``the whole solution is a valid chain.'' This leads to broken chains, branching paths, disconnected pieces, or non-closed cycles.

\subsubsubsection{3.2.3 Missing Capacity or Coverage Skeleton}
\label{app:subsubsubsec:mis_cap_cov}

\textbf{Definition.} The model omits the fundamental constraints that bound resource capacity, enforce demand coverage, or guarantee the completeness of assignment. Supply may become unbounded, and demand may go unserved or be multiply served.

\textbf{Typical erroneous form.} Assignment, flow, or activation variables are introduced, but the model lacks constraints of the form
$
\sum_j x_{ij}\le C_i,
\quad
\sum_i x_{ij}=1,
\quad \text{or} \quad
\sum_i x_{ij}\ge d_j.
$

\textbf{Example.} In warehouse-to-customer distribution, let $x_{ij}\ge 0$ be the shipment from warehouse $i$ to customer $j$. If warehouse capacity $C_i$ and customer demand $d_j$ are known, then at least
$
\sum_j x_{ij}\le C_i \quad \forall i
$
and
$
\sum_i x_{ij}=d_j \quad \forall j
$
should be present. Without them, a cheap warehouse may supply unlimited volume, or some customers may simply be ignored.

\textbf{Consequence.} The most basic supply-demand structure collapses. The model may return plans that look cheap or elegant only because the foundational resource or coverage framework was never built.

\subsubsubsection{3.2.4 Missing Initial or Terminal Conditions}\label{app:subsubsubsec:mis_ini_ter}

\textbf{Definition.} The model lacks constraints that define the starting point, ending point, beginning inventory, ending inventory, planning horizon, or start-end location conditions. Intermediate constraints may exist, but the overall process is not anchored.

\textbf{Typical erroneous form.} Conditions such as $I_0=\bar I_0$, $s_0=\bar s_0$, $C_j\le H$, or ``the path must start at the source and end at the sink'' are absent even though middle-period or middle-node constraints are present.

\textbf{Example.} In production-inventory planning, a given initial inventory $I_0^{\text{given}}$ and a terminal safety-stock requirement $I_T\ge \underline I$ should both be written explicitly. If the code uses only intermediate-period equations, the plan may start from a freely chosen initial state or exhaust all inventory at the horizon end with no penalty.

\textbf{Consequence.} The process loses its endpoints. The model may assume nonexistent starting resources or exploit the absence of a terminal condition to defer cost or empty the system unrealistically.

\textbf{Summary.} Missing constraint skeleton errors arise when the main beams of the model are absent: balance and conservation equations, path and chain structure, capacity and coverage relations, and initial or terminal anchoring conditions. A model may contain many variables and local inequalities yet still be an empty shell if this structural backbone is missing.

\subsubsection{Missing Implicit Constraints}
\label{app:subsubsec:mis_imp_con}

\textbf{Definition.} The main objectives, variables, and explicit constraints may all be present, but the model fails to include implicit rules that are taken for granted in the application domain, assumed by common business practice, or logically implied by action legality. The result is a formally solvable model with large loopholes.

\textbf{Boundary.} This section covers rules that may not have been written as formulas in the statement but are still assumed to hold in practice. If the key structural backbone is missing, the issue belongs to Sec.~\ref{app:subsubsec:mis_con_ske}. If the rule was included but given the wrong direction, location, or threshold, the issue belongs to Sec.~\ref{app:subsubsec:con_bou_dir_err}.

\subsubsubsection{3.3.1 Missing Default Business Rules}
\label{app:subsubsubsec:mis_def_bus}

\textbf{Definition.} Operational or organizational rules that are usually taken for granted in the application domain are not written explicitly. The model satisfies visible assignment or selection requirements while violating real execution logic.

\textbf{Typical erroneous form.} The code ensures that tasks are assigned, flights are covered, or orders are processed, but omits default requirements such as duty chains starting and ending at a base, minimum rest between shifts, vehicle return-to-depot logic, or legally required rest windows.

\textbf{Example.} In crew scheduling, every flight may be assigned and some temporal connection rules may be present. Yet if each crew belongs to a base, a complete duty chain should usually start from that base and return to it, or else satisfy prescribed overnight rules. If base-entry and base-exit logic is not modeled, the plan may begin at the wrong airport or end at a location from which the crew cannot be recovered.

\textbf{Consequence.} The model returns plans that look fully covered on paper but cannot actually be executed. These errors are often hidden because the objective value and coverage rate still look good.

\subsubsubsection{3.3.2 Missing Must-Include Object Constraints}
\label{app:subsubsubsec:mis_mus_inc}

\textbf{Definition.} An action, state change, or resource use is defined in a way that necessarily requires some specific carrier, medium, host, or hub, but the model never forces that necessary object to participate.

\textbf{Typical erroneous form.} The code defines an action variable only, while failing to bind that action to a required host object such as a vehicle, dock, warehouse, equipment state, or handover link.

\textbf{Example.} In warehouse loading, let $x_{ot}\in\{0,1\}$ indicate that order $o$ is loaded in period $t$. If any loading action must occupy a particular loading bay $k$, then one usually also needs loading-bay assignment variables $y_{okt}$ and a relation such as
$
\sum_k y_{okt}=x_{ot}.
$
Without it, an order can be ``loaded'' with no bay actually performing the work.

\textbf{Consequence.} The action becomes detached from the object that is required to carry it out. The model produces pseudo-feasible solutions of the form ``the action happened, but no one or nothing performed it.''

\subsubsubsection{3.3.3 Missing Legal-Action Set Restrictions}
\label{app:subsubsubsec:mis_leg_act}

\textbf{Definition.} The model creates variables for possible actions but does not restrict them to the predefined legal candidate set, reachability set, compatibility set, or allowed transition set. Illegal action channels therefore remain open.

\textbf{Typical erroneous form.} Variables should be defined only for $a\in\mathcal A$, but the code defines them over a larger universal set and later hopes that some subset of constraints will screen out the illegal actions.

\textbf{Example.} In AGV transport scheduling, the AGV may move only along arcs in a given track network $A$. A correct model defines movement variables only for $(i,j)\in A$. If the code defines them for all node pairs $(i,j)$ and relies on later capacity or connectivity constraints to remove invalid moves, the solver may exploit nonexistent shortcut arcs.

\textbf{Consequence.} The action space is enlarged from the intended legal set to a superset. The model may then achieve better cost, shorter paths, or higher efficiency only by using actions that should never have been available.

\textbf{Summary.} Missing implicit constraints are the loophole-closing rules of a model. They include default business rules, must-include object relations, and legal-action-set restrictions. Without them, the model may satisfy visible requirements while still taking invalid shortcuts.

\subsubsection{Erroneous Equivalent Substitutions}
\label{app:subsubsec:err_equ_sub}

\textbf{Definition.} In an attempt to simplify, linearize, localize, or engineer a constraint, the model replaces the original condition with something that looks similar but is not actually equivalent. The substitute preserves only part of the semantics, so the feasible region is silently changed.

\textbf{Boundary.} The concern here is not the absence of a constraint but a misleading stand-in for it. If the rule itself was misunderstood, the issue belongs to Sec.~\ref{app:subsubsec:con_sem_tra_err}. If the backbone is missing, it belongs to Sec.~\ref{app:subsubsec:mis_con_ske}. If the problem is specifically a broken logical encoding or Big-$M$ implementation, it belongs to Sec.~\ref{app:subsubsec:log_con_cha_err}.

\subsubsubsection{3.4.1 Replacing an Equality with an Inequality}
\label{app:subsubsubsec:rep_equ_ine}

\textbf{Definition.} A relation that should hold exactly or be perfectly balanced is replaced by an upper-bound or lower-bound inequality. The precise feasibility condition is thereby relaxed or skewed.

\textbf{Typical erroneous form.} A condition such as
$
\sum_{a\in A_t} x_{at}=1
$
is written as
$
\sum_{a\in A_t} x_{at}\le 1
$
or
$
\sum_{a\in A_t} x_{at}\ge 1.
$
The former permits doing nothing; the latter permits repeated selection.

\textbf{Example.} In equipment control, each period may require exactly one operating mode. If $x_{mt}\in\{0,1\}$ indicates whether mode $m$ is selected in period $t$, then the correct relation is
$
\sum_m x_{mt}=1,\quad \forall t.
$
Using $\le 1$ allows empty mode selection; using $\ge 1$ allows multiple modes simultaneously.

\textbf{Consequence.} A precisely defined state or action relation is replaced by an incomplete one, allowing the solver to exploit ``do nothing'' or ``do several at once'' loopholes.

\subsubsubsection{3.4.2 Using Local or Partial Conditions to Replace a Full Structure}
\label{app:subsubsubsec:usi_loc_par}

\textbf{Definition.} A feasible structure that requires a complete set of conditions or a global structural description is replaced by a few easy local conditions. Those local conditions may be necessary, but they are not sufficient for the full structure.

\textbf{Typical erroneous form.} The code keeps pairwise nonconflict, local connectability, degree constraints, one-sided activation, or some other necessary subset of conditions, and mistakes that subset for a complete description of the intended global structure.

\textbf{Example.} In vehicle routing, the model may impose exactly one inbound and one outbound arc for every customer. Without subtour elimination, however, these local degree relations do not guarantee one connected tour. Multiple disconnected cycles can still satisfy all local degree conditions.

\textbf{Consequence.} The model keeps the local shadow of the true structure but loses its full global shape. Solutions may look locally legal while remaining globally disconnected, fragmented, or improperly closed.

\subsubsubsection{3.4.3 Treating an Approximate Constraint as an Exact One}
\label{app:subsubsubsec:tre_app_con}

\textbf{Definition.} A heuristic rule, empirical threshold, geometric proxy, or engineering approximation is used in place of the exact feasibility condition demanded by the original problem.

\textbf{Typical erroneous form.} Instead of a precise temporal recursion, physical balance law, network-reachability relation, or state-transition equation, the code uses rules such as ``distance below some threshold,'' ``capacity difference not too large,'' or ``empirically connectable.''

\textbf{Example.} In vehicle routing with time windows, the feasibility of visiting customer $j$ after customer $i$ should come from an arrival-time recursion such as
$
T_j \ge T_i + s_i + \tau_{ij} - M(1-x_{ij}).
$
If the code simply allows the arc whenever the geographic distance between the two customers is below some threshold, it replaces an exact temporal condition with a crude proxy.

\textbf{Consequence.} The exact feasible region is replaced by an approximate one. Some genuinely feasible plans are excluded, while infeasible ones may be admitted.

\textbf{Summary.} Erroneous equivalent substitutions are dangerous because the model still seems to contain constraints, and may even look simpler, but the constraints now encode a different feasible region. Typical forms are replacing equalities with inequalities, mistaking local conditions for full structure, and passing off approximate rules as exact ones.

\subsubsection{Logic and Conditional-Chain Errors}
\label{app:subsubsec:log_con_cha_err}

\textbf{Definition.} The model attempts to encode ``if ..., then ...,'' ``if and only if,'' ``only if ..., then allowed,'' or multi-step trigger chains using binary variables, Big-$M$ constraints, activation bounds, or linearization. Yet the encoding itself is wrong: directions are reversed, constants are too loose or too tight, one direction of an equivalence is missing, or a multi-step logical chain breaks partway through.

\textbf{Boundary.} This section concerns how logical implications are encoded mathematically. If the rule itself was misunderstood, the problem belongs to Sec.~\ref{app:subsubsec:con_sem_tra_err}. If the backbone of constraints is missing, the issue belongs to Sec.~\ref{app:subsubsec:mis_con_ske}. If a non-equivalent stand-in replaced the original condition, the issue belongs to Sec.~\ref{app:subsubsec:err_equ_sub}.

\subsubsubsection{3.5.1 Big-$M$ Direction Error}
\label{app:subsubsubsec:big_dir_err}

\textbf{Definition.} A Big-$M$ implication is written with the trigger direction reversed, or with the logical control variable tied to the wrong side of the relaxed constraint. The encoded logic is therefore the opposite of what was intended.

\textbf{Typical erroneous form.} To express ``when $y=1$, enforce $a^\top x\le b$; when $y=0$, do not enforce it,'' one typically writes
$
 a^\top x \le b + M(1-y).
$
If the code writes
$
 a^\top x \le b + My,
$
then the constraint is loose when $y=1$ and tight when $y=0$, which is exactly the opposite logic.

\textbf{Example.} Suppose $y_t\in\{0,1\}$ indicates whether overtime mode is turned on in period $t$, and only then may working hours exceed the regular limit $H$. A natural formulation is $h_t\le H+My_t$. If the code uses $h_t\le H+M(1-y_t)$, the model allows extra hours when overtime is off and tightens the constraint when overtime is on.

\textbf{Consequence.} A variable intended to activate a condition ends up deactivating it. The model then rewards the wrong status: extra resources may appear when a mode is not selected, and selected options may become more constrained rather than less.

\subsubsubsection{3.5.2 Big-$M$ Too Loose or Too Tight}
\label{app:subsubsubsec:big_too_loo}

\textbf{Definition.} The logical form of the Big-$M$ constraint may be correct, but the chosen value of $M$ is much too large or much too small. The constraint then becomes either numerically ineffective or overly restrictive.

\textbf{Typical erroneous form.} A logical implication of the form $a^\top x\le b+M(1-y)$ uses an $M$ chosen without reference to actual time bounds, capacity bounds, variable bounds, or data magnitudes.

\textbf{Example.} In sequencing, if $s_j$ is the start time of task $j$, $C_i$ is the completion time of task $i$, and $y_{ij}=1$ means that $j$ follows $i$, a non-overlap relation may be written as
$
 s_j \ge C_i - M(1-y_{ij}).
$
If the planning horizon is about $100$, then $M$ should be derived at that scale. Choosing $M=10^6$ makes the LP relaxation nearly useless; choosing $M=10$ may forbid legitimate schedules.

\textbf{Consequence.} An oversized $M$ hurts computational strength and solve time, while an undersized $M$ incorrectly shrinks the feasible region and may even make the model infeasible.

\subsubsubsection{3.5.3 Incomplete Logical Linearization}
\label{app:subsubsubsec:inc_log_lin}

\textbf{Definition.} A linear encoding for an ``if and only if,'' max/min trigger, logical-and, logical-or, or activated product is only partially written. One direction or one bound of the equivalence is missing, so the logical variable expresses only a necessary condition rather than the full intended relation.

\textbf{Typical erroneous form.} A common one-sided activation is $x\le Mz$, which guarantees $z=0\Rightarrow x=0$. If the intended rule also requires a minimum operating scale whenever $z=1$, then a lower bound such as $x\ge Lz$ is also needed. Similar incompleteness occurs when absolute-value or max-equality relations are represented only by one side of their epigraph.

\textbf{Example.} In warehouse location, a rule like ``if a warehouse is opened, it must carry at least some minimum throughput'' cannot be captured by the upper activation bound $\sum_j x_{ij}\le M_i y_i$ alone. Without the lower link, the model allows a warehouse to be ``open'' yet carry zero flow.

\textbf{Consequence.} The logical variables look meaningful by name but lack a complete numerical semantics. Facilities may be activated without use, modes may be selected without consequences, and connection indicators may equal one without triggering the full downstream structure.

\subsubsubsection{3.5.4 Broken Conditional Chains}
\label{app:subsubsubsec:bro_con_cha}

\textbf{Definition.} A logical variable is introduced for one local condition or connection, but the implied condition is not propagated through the rest of the chain of related constraints. The logic works locally yet does not support the full intended structure.

\textbf{Typical erroneous form.} A connection, activation, or mode variable controls one or two local conditions, but is not passed forward into all related time, space, resource, qualification, base, or state constraints.

\textbf{Example.} In crew scheduling, a variable may indicate that flight $g$ follows flight $f$ for crew $c$. The model may use it to prevent temporal overlap, yet fail to use it to enforce airport continuity, base consistency, duty-length limits, and rest-window rules. The variable then means only ``temporally connectable,'' not ``legally and operationally connected.''

\textbf{Consequence.} A condition variable exists but does not carry through the full causal chain. The model may allow connections that are feasible in time but impossible in space or regulation.

\textbf{Summary.} Logic and conditional-chain errors occur when the model realizes that implication logic must be encoded, but encodes it badly. Big-$M$ direction mistakes, bad $M$ magnitudes, incomplete linearizations, and broken trigger chains are the most common forms.

\subsubsection{Aggregation and Index-Coding Errors}
\label{app:subsubsec:agg_ind_cod_err}

\textbf{Definition.} The business rule behind a constraint may have been recognized correctly, but when the rule is written as sums, grouped constraints, per-object expansions, or cross-level aggregations, the aggregation level, summation range, or subscript binding is wrong. The constraint may look familiar while applying to the wrong objects or the wrong statistical level.

\textbf{Boundary.} The issue here is how the constraint is written over objects and indices. If the semantic rule itself is wrong, the issue belongs to Sec.~\ref{app:subsubsec:con_sem_tra_err}. If the core structure is missing, it belongs to Sec.~\ref{app:subsubsec:mis_con_ske}. If the problem is Big-$M$ direction or logical propagation, it belongs to Sec.~\ref{app:subsubsec:log_con_cha_err}.

\subsubsubsection{3.6.1 Wrong Aggregation Level}
\label{app:subsubsubsec:wro_agg_lev}

\textbf{Definition.} A constraint that should hold separately for each customer, machine, period, order, or other fine-grained object is aggregated too early and enforced only in total, so local feasibility is lost.

\textbf{Typical erroneous form.} A correct form such as
$
\sum_i x_{ij}=d_j,\quad \forall j
$
is replaced by
$
\sum_j\sum_i x_{ij}=\sum_j d_j.
$
The latter enforces only total balance, not per-customer satisfaction.

\textbf{Example.} In warehouse distribution, each customer's demand must be satisfied individually. If the code enforces only equality between total shipped volume and total demand, some customers may be overserved while others receive nothing.

\textbf{Consequence.} Local feasibility is falsely replaced by aggregate feasibility. The overall accounting may balance while the plan fails at the level where execution actually matters.

\subsubsubsection{3.6.2 Wrong Summation Range}
\label{app:subsubsubsec:wro_sum_ran}

\textbf{Definition.} The set being summed over is wrong. Some objects that should enter the count are missing, or objects that should not enter are included, changing the effective candidate domain of the constraint.

\textbf{Typical erroneous form.} A relation that should sum over a legal candidate set $I(j)$ is written over the full set $I$ or over an incorrect subset $I'(j)$.

\textbf{Example.} In task assignment, if task $j$ may be assigned only to qualified employees in $E(j)$, the correct coverage rule is
$
\sum_{i\in E(j)} x_{ij}=1.
$
If the code sums over all employees $E$, then unqualified employees are allowed; if it sums over an undersized subset, legal assignees are incorrectly excluded.

\textbf{Consequence.} The constraint no longer acts on the correct candidate set. Illegal objects may be admitted and legal ones may be removed, even though the symbolic form still resembles the right formula.

\subsubsubsection{3.6.3 Wrong Subscripts When Calling a Constraint}
\label{app:subsubsubsec:wro_sub_whe}

\textbf{Definition.} The variable itself has been defined correctly, but in a specific constraint expression its subscripts are swapped or rebound so that the constraint acts on the wrong object relation.

\textbf{Typical erroneous form.} A constraint should use $x_{ij}$ but uses $x_{ji}$, or should use $x_{ikt}$ but uses $x_{itk}$. When the subscripts belong to different sets, this changes the semantic relation rather than just the notation.

\textbf{Example.} In a multi-warehouse, multi-customer assignment problem, $x_{ij}$ may denote shipment from warehouse $i$ to customer $j$. The demand constraint should be $\sum_i x_{ij}=d_j$. If the code writes $\sum_i x_{ji}=d_j$, it effectively turns customers into supply nodes and warehouses into demand nodes.

\textbf{Consequence.} The model constrains the wrong relation. Supply and demand, predecessor and successor, origin and destination, or resource and task roles may all be silently interchanged.

\subsubsubsection{3.6.4 Repeated or Missing Aggregation}
\label{app:subsubsubsec:rep_mis_agg}

\textbf{Definition.} A dimension is summed twice so that the same load or demand is counted repeatedly, or a required level of aggregation is omitted so that the statistic is incomplete.

\textbf{Typical erroneous form.} Repeated aggregation occurs when a variable that does not depend on time is summed over time anyway; missing aggregation occurs when a load that should be summed over both requests and periods is summed over only one of those dimensions.

\textbf{Example.} In call-center scheduling, if $x_{itk}$ indicates whether employee $i$ serves request $k$ in period $t$ and each request lasts $h_k$, then daily workload should be constrained by
$
\sum_t\sum_k h_k x_{itk}\le H_i.
$
If the code omits the sum over periods, the bound applies only to each period separately; if it adds an extra summation over a dimension that the variable does not actually carry, it counts the same work multiple times.

\textbf{Consequence.} The measurement scale of the constraint is distorted. Double-counting makes the resource look tighter than it is, and missing aggregation makes the constraint too loose.

\textbf{Summary.} Aggregation and index-coding errors arise when correct conceptual rules are written at the wrong statistical level, over the wrong candidate set, or with the wrong subscript arrangement. The expressions may look right on the surface while acting on the wrong objects and totals.

\subsubsection{Constraint Boundary and Direction Errors}
\label{app:subsubsec:con_bou_dir_err}

\textbf{Definition.} The basic semantics and objects of the constraint may be right, yet the inequality direction, the placement of upper and lower bounds, the point at which a boundary takes effect, or the threshold parameter itself is wrong. The constraint still exists, but it cuts the feasible region in the wrong direction or at the wrong place.

\textbf{Boundary.} This section is about how much is restricted, in which direction, and at which boundary location or time point. Missing default rules belong to Sec.~\ref{app:subsubsec:mis_imp_con}, missing skeleton constraints belong to Sec.~\ref{app:subsubsec:mis_con_ske}, and wrong summation domains belong to Sec.~\ref{app:subsubsec:agg_ind_cod_err}.

\subsubsubsection{3.7.1 Wrong Inequality Direction}
\label{app:subsubsubsec:wro_ine_dir}

\textbf{Definition.} A quantity that should be bounded above, bounded below, guaranteed, or suppressed is constrained with the inequality sign reversed.

\textbf{Typical erroneous form.} A relation intended as $a^\top x\le b$ is written as $a^\top x\ge b$, or a coverage requirement $\sum_i x_{ij}\ge d_j$ is written as $\sum_i x_{ij}\le d_j$.

\textbf{Example.} In production scheduling, if line output in period $t$ must not exceed capacity $C_t$, the correct form is $q_t\le C_t$. Writing $q_t\ge C_t$ turns a maximum-capacity restriction into a minimum-production requirement.

\textbf{Consequence.} A constraint intended to prevent overload, lateness, or undercoverage becomes one that enforces exactly those things.

\subsubsubsection{3.7.2 Swapped Upper and Lower Bounds}
\label{app:subsubsubsec:swa_upp_low}

\textbf{Definition.} The upper and lower bounds of an interval are exchanged, or a bound parameter is placed on the wrong side of the inequality, so the feasible interval shifts, shrinks incorrectly, or becomes empty.

\textbf{Typical erroneous form.} A box constraint $L\le x\le U$ is written as $U\le x\le L$, or the parameters intended for one side of the interval are attached to the other.

\textbf{Example.} If reactor load must stay within a stable band $R^{\min}\le r_t\le R^{\max}$, writing $R^{\max}\le r_t\le R^{\min}$ or splitting it into $r_t\ge R^{\max}$ and $r_t\le R^{\min}$ reverses the allowed interval entirely.

\textbf{Consequence.} Feasible operating regions may disappear, or the variable may be pushed to the wrong side of the domain altogether.

\subsubsubsection{3.7.3 Misplaced Boundary Conditions}
\label{app:subsubsubsec:mis_bou_con}

\textbf{Definition.} Initial conditions, terminal conditions, start-end conditions, or layer-specific boundary requirements are written into the model but attached to the wrong period, node, layer, or object. The boundary exists in form but not at the right location.

\textbf{Typical erroneous form.} An initial inventory condition $I_0=I_0^{\text{given}}$ is written as $I_1=I_0^{\text{given}}$, a terminal condition is imposed on an intermediate period, or a depot-departure condition is applied at every time layer rather than only at the start layer.

\textbf{Example.} In multi-period inventory planning, if $I_0$ is known, the correct relation is $I_0=I_0^{\text{given}}$. Writing $I_1=I_0^{\text{given}}$ shifts the whole time axis by one period. Similar mistakes occur in time-space networks when start or end constraints are enforced on all nodes instead of only the boundary nodes.

\textbf{Consequence.} The feasible region is shifted or misanchored. Time windows, initial-final logic, and start-end behavior may all move to the wrong places.

\subsubsubsection{3.7.4 Misuse of Threshold or Boundary Parameters}
\label{app:subsubsubsec:mis_thr_bou}

\textbf{Definition.} The mathematical form of the constraint is correct, but the threshold, capacity, minimum scale, safety limit, risk bound, or qualification parameter used inside it belongs to the wrong object or the wrong business meaning.

\textbf{Typical erroneous form.} The model should use object-specific parameters such as $C_i$, $L_i$, or $\theta_i$, but instead uses another object's parameter, a global placeholder, or a parameter of the wrong semantic type.

\textbf{Example.} In fleet scheduling, vehicle $k$ should satisfy a load constraint using its own capacity $Q_k$. If the code uses one common $Q$ for all vehicles or mistakenly applies vehicle $k'$'s capacity to vehicle $k$, then light vehicles may be allowed to overload while heavy vehicles become unnecessarily constrained.

\textbf{Consequence.} The shell of the constraint looks correct, but the cut is made at the wrong place because the wrong bound value is used. These mistakes are especially hidden because the formula itself still looks reasonable.

\textbf{Summary.} Constraint boundary and direction errors are not about absent constraints; they are about constraints cutting in the wrong direction or at the wrong place. Common forms include reversed inequalities, swapped interval endpoints, misplaced boundary conditions, and wrong threshold parameters.

\subsubsection{Constraint Strength Imbalance}
\label{app:subsubsec:con_str_imb}

\textbf{Definition.} A constraint may be semantically correct and applied to the right objects, yet it is either much too weak or much too strong, or its slack, tolerance, or penalty parameters are set on the wrong scale. The issue is not whether the constraint exists, but whether it cuts the feasible region with the correct strength.

\textbf{Boundary.} If the rule was mistranslated, the issue belongs to Sec.~\ref{app:subsubsec:con_sem_tra_err}. If structural constraints are missing, the issue belongs to Secs.~\ref{app:subsubsec:mis_con_ske}--\ref{app:subsubsec:mis_imp_con}. If the problem is Big-$M$ logic or equivalent replacement, it belongs to Secs.~\ref{app:subsubsec:err_equ_sub}--\ref{app:subsubsec:log_con_cha_err}.

\subsubsubsection{3.8.1 Constraint Too Weak}
\label{app:subsubsubsec:con_too_wea}

\textbf{Definition.} The constraint keeps only a weak shadow of the original restriction, so many plans that should be excluded remain feasible.

\textbf{Typical erroneous form.} The code keeps only one side of a necessary relation, uses an excessively large tolerance, constrains only totals rather than local structure, or retains a merely relevant-looking but ineffective condition. For example, an exact assignment requirement may be weakened from equality to an upper bound.

\textbf{Example.} If each customer must be assigned to exactly one service center, then
$
\sum_i x_{ij}=1,\quad \forall j
$
is the natural form. Replacing it with
$
\sum_i x_{ij}\le 1
$
allows some customers to remain unassigned.

\textbf{Consequence.} The model admits pseudo-feasible plans that look cheaper or better only because they do less work, serve fewer demands, or assume away responsibilities that should have been mandatory.

\subsubsubsection{3.8.2 Constraint Too Strong}
\label{app:subsubsubsec:con_too_str}

\textbf{Definition.} The constraint is stricter than the original problem intended and removes solutions that should remain feasible, sometimes making the model infeasible.

\textbf{Typical erroneous form.} ``At least'' or ``at most'' is written as ``exactly,'' a soft tolerance is turned into an equality, or a local requirement is incorrectly expanded to every object or every period.

\textbf{Example.} In a multi-warehouse supply problem, a customer's demand may legally be split among several warehouses. If the code adds a uniqueness restriction forcing each customer to be served by exactly one warehouse, it tightens the original business rule unnecessarily and may eliminate valid plans.

\textbf{Consequence.} The model no longer solves the original task but a stricter variant. Costs rise, utilization falls, the number of feasible plans shrinks, and the model may even be declared infeasible although the real problem is feasible.

\subsubsubsection{3.8.3 Slack or Tolerance Parameters on the Wrong Scale}
\label{app:subsubsubsec:sla_tol_par}

\textbf{Definition.} Tolerance parameters, allowable violation levels, soft-constraint slack bounds, or similar control parameters exist in the model, but their magnitudes are badly misaligned with the real business tolerance.

\textbf{Typical erroneous form.} A parameter like $\epsilon$ is set so large that the constraint becomes almost meaningless, or so small that even acceptable variation is treated as a violation. Similar issues arise with slack caps and non-Big-$M$ softening parameters.

\textbf{Example.} In manufacturing scheduling, a lateness variable $l_j\ge 0$ may satisfy $l_j\le \Delta_j$, where $\Delta_j$ is the allowed delivery delay. If the real tolerance is one day but the code sets $\Delta_j=30$ for every order, then the lateness control is nearly void. If it sets $\Delta_j=0$ despite an allowed buffer, then any small delay becomes impermissible.

\textbf{Consequence.} The model either exploits an oversized tolerance window or becomes unnecessarily rigid because the allowed slack is too small. The implemented strength no longer matches the real operational tolerance.

\textbf{Summary.} Constraint strength imbalance concerns whether the constraint cuts with the right force. Constraints that are too weak let loopholes remain; constraints that are too strong remove valid solutions; and badly scaled tolerance parameters distort enforcement even when the formal structure looks fine.

\subsubsection{Scheduling- and Activity-Structure-Specific Errors}
\label{app:subsubsec:sch_act_str_spe_err}

\textbf{Definition.} In scheduling, project networks, crew-connection problems, multimode operations, or optional-activity settings, the model may have the right basic variables and some local constraints, yet fail to preserve the distinctive topology, mode-switching semantics, and activity-activation logic of those problem classes.

\textbf{Boundary.} This section collects the remaining structural errors/hallucinations that are particularly characteristic of time-ordered and activity-based models. General errors/hallucinations in equivalent replacement, logic, aggregation, or boundary direction still belong to earlier sections; Sec.~\ref{app:subsubsec:sch_act_str_spe_err} adds the especially common distortions that arise in precedence networks and mode-choice structures.

\subsubsubsection{3.9.1 Wrong Encoding of Precedence Relations}
\label{app:subsubsubsec:wro_enc_pre}

\textbf{Definition.} A precedence relation among tasks, activities, flights, jobs, or events is recognized, but its start times, finish times, lags, trigger conditions, predecessor sets, or successor sets are encoded incorrectly. The resulting precedence constraint no longer represents the intended topology.

\textbf{Typical erroneous form.} Common forms include replacing ``start after finish'' with ``start after start,'' omitting processing or service duration, writing predecessor sets as successor sets, or using $s_j\ge s_i+p_j$ where $s_j\ge s_i+p_i$ is required.

\textbf{Example.} In project scheduling, if activity $j$ may start only after activity $i$ finishes, with start time $s_i$ and duration $p_i$, then the standard relation is
$
 s_j \ge s_i + p_i.
$
If the code writes $s_j\ge s_i+p_j$, it uses the successor duration instead of the predecessor duration, which can either delay too much or allow too-early starts depending on the durations.

\textbf{Consequence.} The topology of the schedule is corrupted. Tasks may start before their predecessors finish, be delayed for the wrong reason, or distort the critical path and all downstream timing logic.

\subsubsubsection{3.9.2 Wrong Structure for Mode Selection and Optional Activities}
\label{app:subsubsubsec:wro_str_mod}

\textbf{Definition.} When the problem contains multimode execution, optional activities, alternative processes, contingency actions, or intensity levels, the model fails to represent correctly that exactly one mode should be chosen, that choosing a mode should activate the corresponding duration/resource/cost data, or that an optional activity must obey a full structural logic once selected.

\textbf{Typical erroneous form.} A multimode task should satisfy
$
\sum_m y_{im}=1,
$
but the code writes $\le 1$ or $\ge 1$; or optional activities are given variables with no links to duration, resource use, or precedence; or multiple mutually exclusive modes are allowed simultaneously.

\textbf{Example.} Suppose task $i$ may be processed in three modes $m\in\{1,2,3\}$, each with its own duration $p_{im}$ and resource consumption $r_{im}$. If $y_{im}\in\{0,1\}$ indicates the selected mode, then one needs at least
$
\sum_m y_{im}=1
$
and coupling relations that make the realized duration and resource consumption depend on the chosen mode. If the code defines the mode variables but does not switch duration, cost, or resource use according to them, the choice is merely nominal.

\textbf{Consequence.} Mode or optional-activity variables become labels without consequences. The model appears to support multiple modes, intensities, or optional actions, yet the actual schedule, resources, and costs do not change when the mode changes.

\textbf{Summary.} Scheduling- and activity-structure-specific errors usually appear as misencoded precedence relations or mode choices whose downstream consequences are never activated. They often become visible only when one checks the entire activity network or full execution chain rather than isolated local constraints.

\subsection{Implementation Hallucinations}
\label{app:subsec:imp_hal}

\textbf{Definition.} This family concerns failures that arise after the symbolic model is already correct or nearly correct, but the solver-facing program does not faithfully realize that model. The key distinction is that the mathematical object exists, yet the realized code changes it through API misuse, dropped loops, wrong objective sense, incompatible solver settings, or incorrect post-solve extraction. In our current detector, this family is instantiated by the implementation branch that combines code parsing with execution-side inspection, so the appendix definitions below are aligned with the subtype inventory actually used by the system.

\textbf{Boundary.} Implementation hallucinations are not ordinary symbolic modeling errors. If the objective, variable, or constraint is already wrong in $M$ and the code simply mirrors that wrong symbolic model, the root cause belongs to $\mathcal{T}^{O}$, $\mathcal{T}^{V}$, or $\mathcal{T}^{C}$ rather than $\mathcal{T}^{I}$. Conversely, if the symbolic model is faithful but the code changes solver sense, variable domains, index expansion, or reported outputs, the error belongs here. The implementation family is therefore about \emph{math-to-code fidelity}, not about reclassifying upstream formulation mistakes.

\textbf{Practical interpretation.} In the implementation branch, we use two complementary views. A parser extracts solver-side variables, objective sense, instantiated constraints, loops, and index materialization into a code graph $G_S$. An execution-oriented checker then inspects the realized program behavior, including solver traces, created variable families, instantiated constraints, and post-solve readout. This division mirrors how implementation errors appear in practice: some are visible statically in the source, while others only become obvious when the realized code path is compared against the symbolic graph $G_M$.

\textbf{Why this family is operationally important.} In OR pipelines, a large share of deployment failures arise not because the analyst wrote the wrong mathematical idea, but because a seemingly faithful algebraic program instantiates, solves, or reports a different model than the one described on paper. These failures are particularly dangerous because standard sanity checks may still pass: the code can compile, the solver can return an ``optimal'' status, and the reported objective can look numerically plausible. What breaks is the fidelity relation between the symbolic graph and the executable artifact.

\subsubsection{Symbolic-Code Mismatch}
\label{app:subsubsec:imp_sym_code}

\textbf{Definition.} The symbolic model and the solver program encode different mathematical objects. Typical manifestations include wrong objective sense in code, omitted constraint materialization, missing variable registration, and stale coefficient mappings.

\textbf{Boundary.} The issue here is not whether the symbolic model itself is right, but whether code faithfully instantiates it. If the mathematical formulation already omits a required constraint, that is a symbolic constraint hallucination; if the formulation contains the constraint but code never adds it to the solver, it is an implementation hallucination.

\subsubsubsection{4.1.1 Wrong Objective Sense in Code}
\label{app:subsubsubsec:imp_wro_obj_sense}

\textbf{Definition.} The symbolic model specifies one optimization sense, but the solver program materializes the opposite sense or an equivalent sign-flipped objective without a faithful correction.

\textbf{Typical erroneous form.} The symbolic formulation declares
$
\min c^\top x,
$
but the code calls a maximize API, negates coefficients without a matching correction, or reports the solved objective with the wrong sign. A common pattern is a solver call such as \texttt{setObjective(expr, GRB.MAXIMIZE)} when the symbolic graph requires minimization.

\textbf{Example.} Suppose a facility-allocation model minimizes transportation cost
$
\min \sum_{i,j} c_{ij} x_{ij}.
$
If the solver code instead sets the same linear expression under a maximize sense, the executable artifact will prefer expensive routes rather than cheap ones. The symbolic model remains correct, but the realized program optimizes the opposite business criterion.

\textbf{Consequence.} This error flips optimality at the executable level. The program may compile, solve, and even return a numerically plausible value, yet every downstream decision is chosen according to the wrong direction of preference.

\subsubsubsection{4.1.2 Omitted Constraint Materialization}
\label{app:subsubsubsec:imp_omi_con_mat}

\textbf{Definition.} A constraint exists in the symbolic formulation, but the solver code never instantiates it in the realized model.

\textbf{Typical erroneous form.} The symbolic model contains a family such as
$
\sum_j x_{ij} \le C_i,\quad \forall i,
$
yet the corresponding loop or \texttt{addConstrs}/\texttt{ConstraintList} call is absent from code. The code may still build variables and an objective, so the omission is easy to miss.

\textbf{Example.} In multi-period inventory planning, the symbolic model may include inventory recursion
$
I_t = I_{t-1} + q_t - d_t,\quad \forall t.
$
If the solver code creates $I_t$ and $q_t$ but never materializes the recursion constraints, inventory levels across periods become disconnected and can effectively appear from nowhere.

\textbf{Consequence.} The executable feasible region becomes strictly larger than the symbolic feasible region. The solver may exploit the missing family to produce unrealistically cheap or infeasible plans while still appearing to solve the intended model.

\subsubsubsection{4.1.3 Missing Variable Registration}
\label{app:subsubsubsec:imp_mis_var_reg}

\textbf{Definition.} A symbolic decision variable is absent from the solver model because the code fails to create or register it.

\textbf{Typical erroneous form.} The symbolic formulation defines a variable family, but the code never creates it, creates only part of it, or stores it under a container that is never passed to the solver model.

\textbf{Example.} In capacitated facility location, binary opening variables $y_i$ and shipment variables $x_{ij}$ are both needed. If code creates only $x_{ij}$ and forgets to register $y_i$, then any opening-cost term and any linking constraint of the form
$
\sum_j x_{ij} \le U_i y_i
$
cannot be faithfully realized, even if the symbolic model contains them.

\textbf{Consequence.} A missing variable does not merely delete one line of code; it breaks every relation that depends on that variable. The realized model becomes another optimization problem with altered structure and altered feasible decisions.

\subsubsubsection{4.1.4 Stale or Divergent Coefficient Mapping}
\label{app:subsubsubsec:imp_sta_div_coeff}

\textbf{Definition.} The solver code reuses outdated, misaligned, or incorrectly mapped coefficients relative to the symbolic model.

\textbf{Typical erroneous form.} A parameter table is read from the wrong column, from a previous scenario, or with indices shifted relative to the symbolic meaning. The code still constructs a valid linear expression, but with coefficients attached to the wrong objects.

\textbf{Example.} In a transportation model, $c_{ij}$ should denote the shipping cost from warehouse $i$ to customer $j$. If the code accidentally reuses a stale matrix from another scenario or swaps row and column meanings, then
$
\sum_{i,j} c_{ij} x_{ij}
$
is still syntactically valid, but the executable artifact optimizes a cost surface that no longer corresponds to the intended instance.

\textbf{Consequence.} This error is especially dangerous because the code usually compiles and solves normally. The solver receives a coherent model, but it is a model of another parameterization rather than of the symbolic formulation that was supposedly implemented.

\textbf{Summary.} Symbolic-code mismatch errors occur when the executable artifact no longer matches the mathematical object that was meant to be implemented. They are often invisible to purely symbolic review because the symbolic model itself can still be perfectly correct.

\subsubsection{Variable and Domain API Mismatch}
\label{app:subsubsec:imp_var_api}

\textbf{Definition.} The solver API assigns the wrong domain, bounds, or admissible index domain to variables relative to the symbolic model. Representative subtypes are wrong API variable type, wrong bounds in code, and wrong index-domain materialization.

\textbf{Boundary.} This category is the code-level analogue of variable-domain fidelity. It should only be used when the symbolic variable declaration is correct but the API realization changes it, for example by materializing a binary variable as continuous or by expanding variables over an overly broad Cartesian product.

\subsubsubsection{4.2.1 Wrong API Variable Type}
\label{app:subsubsubsec:imp_wro_api_var_type}

\textbf{Definition.} The solver API creates a variable with the wrong type, such as continuous instead of binary or integer.

\textbf{Typical erroneous form.} A symbolic binary or integer variable is created in code with a continuous domain, or a continuous variable is incorrectly forced into an integer API type. In practice this often appears through calls such as \texttt{addVar(vtype=CONTINUOUS)} for a symbolic binary, or by using a generic nonnegative variable constructor where an integral container was required.

\textbf{Example.} Suppose a shipment-count variable $n_{ij}$ represents the number of truck trips on arc $(i,j)$ and should therefore be integral. If the solver code creates it with a continuous type, the executable model may use $2.4$ trips. The symbolic model remains discrete, but the realized API object does not enforce that discreteness.

\textbf{Consequence.} The code changes the feasible set directly. This can produce lower costs, smoother flows, or artificial feasibility that exist only because the solver API no longer respects the symbolic variable type.

\subsubsubsection{4.2.2 Wrong Bounds in Code}
\label{app:subsubsubsec:imp_wro_bounds_code}

\textbf{Definition.} The solver program assigns incorrect lower or upper bounds relative to the symbolic specification.

\textbf{Typical erroneous form.} A lower bound is omitted, an upper bound is relaxed, or bounds are swapped or sign-flipped during code generation.

\textbf{Example.} If production variable $x_t$ should satisfy
$
0 \le x_t \le U_t,
$
but the code creates $x_t$ with no upper bound or with a negative lower bound, the solver can exploit production levels that the symbolic model never intended to admit.

\textbf{Consequence.} Even when all coefficients and constraints look consistent, the executable domain is wrong. This can silently alter both feasibility and optimality and is often hard to detect from objective values alone.

\subsubsubsection{4.2.3 Wrong Index Domain Materialization}
\label{app:subsubsubsec:imp_wro_index_domain}

\textbf{Definition.} The code creates variables on the wrong admissible object set or index domain even though the symbolic model specifies the correct one.

\textbf{Typical erroneous form.} The symbolic model restricts variables to legal arcs, compatible pairs, or candidate facilities, but the code creates them over a larger or different set.

\textbf{Example.} In a network flow model, variables $x_{ij}$ may be defined only for $(i,j)\in A$, where $A$ is the admissible arc set. If code instead creates $x_{ij}$ for all node pairs, the executable model gains artificial shortcut arcs that were never part of the symbolic graph.

\textbf{Consequence.} The code introduces additional decision channels or removes legal ones. The resulting solver model may look reasonable but is structurally defined on the wrong object universe.

\textbf{Summary.} Variable and domain API mismatch errors are implementation-level distortions of the decision space itself. They do not merely affect reporting; they change which values and which indices the solver is actually allowed to optimize over.

\subsubsection{Index and Set Materialization Errors}
\label{app:subsubsec:imp_idx_set}

\textbf{Definition.} The symbolic indices are correct, but loops, comprehensions, or filters in code fail to instantiate them correctly. Typical subtypes are dropped loop or index expansion, partial set expansion, and filtered index loss.

\textbf{Boundary.} These are not symbolic aggregation mistakes. The mathematical indices may be fully correct in $M$, yet the program only instantiates a subset of them, drops one nested loop, or filters legal objects away during code generation.

\subsubsubsection{4.3.1 Dropped Loop or Index Expansion}
\label{app:subsubsubsec:imp_drop_loop}

\textbf{Definition.} A loop, nested index, or repeated expansion required by the symbolic model is omitted when code instantiates the solver object.

\textbf{Typical erroneous form.} A family that should be generated over $(i,t)$, $(i,j,k)$, or every period in a horizon is instantiated only over one dimension because one loop is missing. Typical code patterns include using a one-level comprehension where a nested comprehension is required, placing a constraint constructor outside the inner loop, or calling a bulk-constructor such as \texttt{addConstrs} over an incomplete iterator.

\textbf{Example.} Suppose capacity constraints should hold for every facility $i$ and period $t$:
$
\sum_j x_{ijt} \le C_{it},\quad \forall i,t.
$
If code loops over $i$ but forgets the loop over $t$, then only one period or one default period is materialized in the solver model.

\textbf{Consequence.} The executable model under-enforces an entire slice of the symbolic family. This is a common way for time-coupled or scenario-coupled models to look valid while still missing most of their actual structure.

\subsubsubsection{4.3.2 Partial Set Expansion}
\label{app:subsubsubsec:imp_partial_set}

\textbf{Definition.} The code materializes only part of a required set or index family, leaving a structurally incomplete realized model.

\textbf{Typical erroneous form.} The code expands a family over only some plants, some customers, or some periods because of a mistaken slice, early break, or partial iteration over a container.

\textbf{Example.} In workforce planning, assignment variables may need to exist for every worker-role pair in a qualified set. If code iterates only over the first department or over roles appearing in a partial dictionary, the instantiated model contains only a fragment of the symbolic index family.

\textbf{Consequence.} The executable model is structurally incomplete. Unlike a single dropped coefficient, this error removes an entire region of the symbolic search space.

\subsubsubsection{4.3.3 Filtered Index Loss}
\label{app:subsubsubsec:imp_filtered_index_loss}

\textbf{Definition.} A filtering step in code removes legal symbolic indices, arcs, items, or periods that should remain in the executable model.

\textbf{Typical erroneous form.} A list comprehension or conditional such as \texttt{if capacity > 0} or \texttt{if i != j} removes indices that are legal in the symbolic model but happen to look suspicious to the programmer or to the model itself.

\textbf{Example.} In a transportation network, zero-cost arcs or self-transfer arcs may still be legal and meaningful. If the code filters them out because they look degenerate, then the executable graph no longer matches the symbolic admissible set.

\textbf{Consequence.} The code silently deletes legal structure. This can make the solver miss feasible plans, report infeasibility, or optimize over an artificially restricted network or time horizon.

\textbf{Summary.} Index and set materialization errors arise when a correct symbolic family is instantiated only partially or selectively in code. They are especially common in multi-index OR models because a small loop mistake can erase a large structural block of the executable artifact.

\subsubsection{Solver and Formalization Mismatch}
\label{app:subsubsec:imp_solver_form}

\textbf{Definition.} The code uses an implementation device that is incompatible with the symbolic formulation, such as missing a required linearization, choosing an incompatible solver, or using incorrect numerical options such as a bad Big-$M$ realization.

\textbf{Boundary.} This category should be used when the symbolic form is acceptable in principle but the implementation layer fails to choose a compatible computational realization. It is therefore different from symbolic formalization errors, where the mathematical object itself is written incorrectly before code generation.

\subsubsubsection{4.4.1 Missing Linearization in Code}
\label{app:subsubsubsec:imp_missing_linearization}

\textbf{Definition.} The symbolic formulation requires a code-level linearization or auxiliary construction, but the implementation omits it.

\textbf{Typical erroneous form.} The symbolic model assumes auxiliary variables and linking constraints for fixed charges, binary-continuous products, absolute values, or max operators, but the code writes only the base variables and forgets the linearization device.

\textbf{Example.} If a facility-opening model requires
$
\sum_j x_{ij} \le U_i y_i
$
to linearize the activation of shipments by binary open variable $y_i$, then omitting that linking device in code allows shipments to occur regardless of whether the facility is opened.

\textbf{Consequence.} The executable model is no longer an equivalent realization of the symbolic one. It solves a relaxed or simply different problem because the intended logical or nonlinear structure was never encoded.

\subsubsubsection{4.4.2 Incompatible Solver Selection}
\label{app:subsubsubsec:imp_incompatible_solver}

\textbf{Definition.} The program selects a solver or backend that is incompatible with the realized formulation class.

\textbf{Typical erroneous form.} The code dispatches a nonlinear, mixed-integer nonlinear, semidefinite, or otherwise specialized model to a backend that cannot represent or solve that class faithfully.

\textbf{Example.} A symbolic formulation may legitimately contain nonlinear risk terms, bilinear pooling relations, or mixed-integer quadratic terms. If the code still routes the model to a pure LP backend without reformulation, then the executable pipeline is incompatible with the formulation it is supposed to realize. In practice this is the difference between sending an MILP to a compatible MIP solver and accidentally flattening a nonlinear formulation into an LP-only interface that cannot preserve the intended semantics.

\textbf{Consequence.} The code may fail outright, quietly downgrade the model, or solve a different relaxation. The user may observe solver failure or misleading feasibility messages even though the symbolic model itself was well-formed.

\subsubsubsection{4.4.3 Incorrect Big-$M$ or Numerical Option in Code}
\label{app:subsubsubsec:imp_incorrect_bigm}

\textbf{Definition.} The code uses an incorrect Big-$M$ value or an incompatible numerical option that changes executable behavior relative to the symbolic model.

\textbf{Typical erroneous form.} A Big-$M$ constant is made too small, too large, or disconnected from the symbolic bound logic; alternatively, a numerical option is chosen that changes intended enforcement behavior.

\textbf{Example.} In a precedence-activation model, constraint
$
s_j \ge s_i + p_i - M(1-y_{ij})
$
relies on a valid $M$. If code hard-codes an $M$ that is too small, legal schedules are cut off; if it is too large, the relaxation becomes numerically weak and may admit unstable behavior.

\textbf{Consequence.} The executable model no longer behaves like the symbolic one, even though both still contain a formally similar constraint family. Numerical choices can therefore create implementation hallucinations without changing the visible symbolic syntax.

\textbf{Summary.} Solver and formalization mismatch errors occur when the symbolic model is mathematically meaningful but the computational realization chosen in code is not the one that preserves that meaning. They are often the bridge between a correct formulation and a misleading executable artifact.

\subsubsection{Post-Solve Extraction and Reporting Divergence}
\label{app:subsubsec:imp_postsolve}

\textbf{Definition.} The code solves one model but reads out, reports, or post-processes another object. Typical subtypes include wrong solution variable readout, misreported objective value, and stale result-object or post-processing errors.

\textbf{Boundary.} These errors occur after optimization has already been run. The solver may have received the correct model, but the reported decision or objective is no longer the one associated with the solved artifact. This makes the downstream user observe a hallucinated executable result even though the optimization stage itself may have been correct.

\subsubsubsection{4.5.1 Wrong Solution Variable Readout}
\label{app:subsubsubsec:imp_wrong_readout}

\textbf{Definition.} The code reads, exports, or displays the wrong variable family after solving, so the reported decision object differs from the one optimized.

\textbf{Typical erroneous form.} The solver optimizes one family of decision variables, but the reporting logic extracts auxiliary variables, reduced costs, slack variables, or a different indexed container and presents them as the final decision.

\textbf{Example.} In a routing model, the true decision may be binary arc variables $x_{ij}$. If the report instead extracts node-balance slack or route-position helper variables and presents them as the chosen arcs, the user sees a solution object that is not the one the symbolic model intended to expose.

\textbf{Consequence.} The optimization stage may be correct, but the artifact presented to the user is hallucinated at the reporting layer. This can trigger wrong operational decisions even when the solve itself succeeded.

\subsubsubsection{4.5.2 Misreported Objective Value}
\label{app:subsubsubsec:imp_misreported_obj}

\textbf{Definition.} The solver may optimize the intended model, but the program reports an objective value that is stale, sign-flipped, or taken from the wrong object.

\textbf{Typical erroneous form.} The code prints a presolve value, a negated objective used for an internal transformation, or an objective stored on an outdated model object rather than the final solved value.

\textbf{Example.} A maximization problem may be implemented internally through coefficient negation and later require a sign correction before reporting. If the correction is skipped, the displayed objective becomes the negative of the true solved objective even though the solver optimized the intended model.

\textbf{Consequence.} Downstream evaluation becomes corrupted. Objective agreement checks may fail spuriously, or worse, a wrong reported objective may be mistaken for evidence that the solver implemented the wrong model.

\subsubsubsection{4.5.3 Stale Result Object or Wrong Post-Processing}
\label{app:subsubsubsec:imp_stale_result}
\textbf{Definition.} Post-solve code uses an outdated model object, stale solution container, or incorrect downstream transformation, causing the emitted result to diverge from the solved artifact.

\textbf{Typical erroneous form.} The code solves one model instance but reads from another, reuses cached values from a previous run, or applies an invalid downstream transformation such as inappropriate rounding or regrouping.

\textbf{Example.} In rolling-horizon planning, code may rebuild and solve the model repeatedly. If the report still reads variable values from the previous model object or from a stale result cache, the displayed plan no longer corresponds to the solve that just finished.

\textbf{Consequence.} This is the implementation analogue of a stale belief state. The optimizer may have solved the correct problem, but the emitted artifact belongs to another run, another model object, or another post-processing pipeline.

\textbf{Summary.} Post-solve extraction and reporting divergence errors occur after mathematical optimization has already happened. They are easy to overlook because the solver status looks normal, yet the artifact finally shown to the user is no longer the artifact that was actually optimized.

\textbf{Overall summary.} Implementation hallucinations are the code-level counterpart of symbolic hallucinations. They arise when a mathematically meaningful formulation is mistranscribed, mistrouted through a solver interface, incompletely instantiated, or incorrectly reported. This is why the implementation branch of our detector does not merely lint code: it compares the realized executable artifact against the symbolic model and checks whether the optimization program that actually ran is the same one that the symbolic graph intended. Put differently, symbolic correctness alone is not enough; the executable artifact must preserve objective sense, variable domains, instantiated index sets, solver-compatible reformulations, and post-solve semantics all the way to the final reported decision.

\section{Limitations and Broader Impacts}
\label{app:limitations_impacts}

\textbf{Limitations.} This work has three main limitations. First, the benchmark is bounded by the current seed universe and by the cost of OR-expert validation. Although our evaluation covers 484 clean reference artifacts, 1266 controlled injected artifacts, and 6292 natural LLM-generated artifacts, the cases are drawn from existing optimization-modeling benchmark families. They may not fully represent more specialized industrial settings, such as large-scale stochastic programs, nonlinear nonconvex models, dynamic control problems, or domain-specific formulations with rich physical constraints. The results should therefore be read as evidence for taxonomy-grounded auditing on the evaluated distribution, not as a guarantee of generalization to all optimization-modeling tasks.

Second, the taxonomy and labels necessarily rely on OR-expert judgment. This is appropriate because structural hallucinations cannot be labeled reliably by objective-value agreement alone, but nearby error types can still have ambiguous boundaries. The controlled injected benchmark intentionally uses one constructed hallucination per case, which makes localization metrics interpretable but does not cover all interacting multi-error patterns. The benchmark also tracks all 83 specific types, while only 58 currently admit at least one defensible construction from the clean seed pool. In the natural benchmark, we annotate artifact-level and major-category labels rather than exhaustive subtype labels, because natural artifacts often contain overlapping errors and full subtype annotation at this scale would be substantially more expensive and less reliable.

Third, OptArgus detects and localizes hallucinations but does not automatically repair them. Its reports provide structured evidence for human modelers, LLM-based repair modules, or agentic modeling workflows, but we do not evaluate whether those findings can be converted into correct repaired formulations or solver implementations. Future work should extend the benchmark to harder industrial instances, develop stronger subtype-level natural annotations, and study closed-loop detection--repair--verification workflows.

\textbf{Broader impacts.} The intended impact of this work is to make LLM-assisted optimization modeling safer to inspect before use in decision-support settings such as logistics, energy systems, transportation, finance, and healthcare. By auditing the structural alignment among the problem description, symbolic formulation, and solver implementation, OptArgus can help identify errors that would be missed by objective-value checks alone.

The main risk is over-reliance. False negatives may allow incorrect optimization models to enter downstream workflows, while false positives may waste expert effort or reduce trust in valid models. OptArgus should therefore be used as an auditing aid, not as a standalone correctness certificate. Deployments involving private industrial data should follow appropriate privacy, security, and access-control procedures, and future systems should report inference and development compute transparently.

\section{Compute Resources}
\label{app:compute_resources}

Our experiments do not involve training large-scale models from scratch or fine-tuning pretrained models.
The compute cost comes from two different stages: constructing the natural benchmark and evaluating the detectors.
The clean benchmark ($484$ cases) and controlled injected benchmark ($1266$ cases) are dominated by manual OR-expert curation, validation, and error injection rather than model training.
The natural benchmark requires model-output generation on the same $484$ seeds across thirteen models.
For the general-purpose models, this generation was performed through commercial or OpenAI-compatible API endpoints where applicable.
For the fine-tuned models for optimization modeling---\texttt{LLMOPT-Qwen2.5-14B}, \texttt{OptMATH-Qwen2.5-7B}, \texttt{ORLM-LLaMA-3-8B}, and \texttt{SIRL-Qwen2.5-7B}---we used local GPU deployments to produce $1936=4\times484$ artifacts from this fine-tuned class.
The generated solver programs were then screened and, when executable, run in local CPU-side solver environments, including Gurobi, PuLP/CBC, or SciPy MILP backends depending on the generated code.
The final natural-benchmark labels were assigned manually by OR experts.

Detector evaluation is mostly LLM-inference and CPU orchestration.
For clean, injected, and natural evaluation, both the Single-Agent Detector and OptArgus read the benchmark problem statement, symbolic model, and solver code as text, call the configured LLM API endpoint, and write structured reports.
Solver/code execution is used for benchmark construction, post-hoc screening, and injected-case QA; it is not the dominant operation inside the OptArgus evaluation loop.
All reported detector evaluations use the default LLM temperature $0$, so the main results are deterministic evaluations over fixed benchmark artifacts rather than repeated stochastic sampling runs.
The final OptArgus natural-benchmark run was launched as thirteen model waves with ten parallel shards per wave and covered all $6292$ artifacts; the recorded wall-clock span from the first wave launch to the final merged metrics was about $8$ hours.
The final clean and injected OptArgus runs used the same API-based detector pipeline on $484$ and $1266$ artifacts, respectively, with smaller wall-clock cost because of their smaller scale.

All primary local orchestration, solver execution, and GPU-backed local model-output generation were conducted on commercial cloud or workstation nodes with Intel(R) Xeon(R) Platinum 8350C-class CPUs, 32 CPU cores, 1024 GB RAM, NVMe SSD storage, and access to up to 8 NVIDIA A100 SXM4 80GB GPUs.
GPU resources were used for local fine-tuned model-output generation and development runs, not for training or fine-tuning the detectors.
For API-served models, provider-side compute is not directly observable; the reproducible local cost is therefore best described by the artifact counts, parallel shard configuration, local CPU/GPU resources, and recorded wall-clock time above.

\end{document}